\documentclass[twoside,11pt]{article}

\usepackage{blindtext}

\usepackage[abbrvbib, preprint]{jmlr2e}

\usepackage{jmlr2e}
\usepackage{graphicx}
\usepackage{booktabs}
\usepackage{pdfpages}
\usepackage{xcolor}
\usepackage{wrapfig}
\usepackage{subcaption}
\usepackage{amsmath}
\usepackage{cleveref}
\usepackage{tabularx}
\usepackage{siunitx}
\usepackage{subcaption}

\usepackage{lastpage}
\jmlrheading{25}{2026}{1-\pageref{LastPage}}{03/26; Revised ?/2?}{?/2?}{21-0000}{Prasse et al.}

\ShortHeadings{From Codebooks to VLMs: Climate Change Visuals}{Prasse et al.}
\firstpageno{1}

\begin{document}
\title{From Codebooks to VLMs: Evaluating Automated Visual Discourse Analysis for Climate Change on Social Media}

\author{\name Katharina Prasse \email katharina.prasse@uni-mannheim.de \\
        \name Steffen Jung \email steffen.jung@uni-mannheim.de \\
       \addr
       Chair of Machine Learning, University of Mannheim\\
       Mannheim, BW 68131, Germany
       \AND
       \name Isaac Bravo \email isaac.bravo@tum.de \\
       \name Stefanie Walter \email stefanie.walter@tum.de \\
       \addr Department of Governance, Technical University Munich\\
       München, BA 80333, Germany
        \AND
       \name Patrick Knab \email patrick.knab@tu-clausthal.de \\
       \name Christian Bartelt \email bartelt@tu-clausthal.de \\
       \addr Clausthal University of Technology\\
       Clausthal-Zellerfeld, NI 38678, Germany 
        \AND
       \name Margret Keuper \email keuper@uni-mannheim.de \\
       \addr 
       Chair of Machine Learning, University of Mannheim\\
       Mannheim, BW 68131, Germany \& \\
       Max Planck Institute for Informatics, Saarland Informatics Campus \\
       Saarbruecken, SL 66123, Germany }

\editor{My editor}

\maketitle

\begin{abstract}
Social media platforms have become primary arenas for climate communication, generating millions of images and posts that - if systematically analysed - can reveal which communication strategies mobilise public concern and which fall flat.
We aim to facilitate such research by analysing how computer vision methods can be used for social media discourse analysis.
This analysis includes application-based taxonomy design, model selection, prompt engineering, and validation. 
We benchmark six promptable vision-language models and 15 zero-shot CLIP-like models on two datasets from X (formerly Twitter) — a 1,038-image expert-annotated set and a larger corpus of over 1.2 million images, with 50,000 labels manually validated — spanning five annotation dimensions: animal content, climate change consequences, climate action, image setting, and image type.
Among the models benchmarked, Gemini-3.1-flash-lite outperforms all others across all super-categories and both datasets, while the gap to open-weight models of moderate size remains relatively small. 
Beyond instance-level metrics, we advocate for distributional evaluation: VLM predictions can reliably recover population level trends even when per-image accuracy is moderate, making them a viable starting point for discourse analysis at scale.
We find that chain-of-thought reasoning reduces rather than improves performance, and that annotation dimension specific prompt design improves performance.
We release tweet IDs and labels along with our code at ~\url{https://github.com/KathPra/Codebooks2VLMs.git}.
\end{abstract} 


\newpage
\section{Introduction}
\label{sec:intro}

Climate change is widely recognised as the defining crisis of the twenty-first century, with the Intergovernmental Panel on Climate Change warning that limiting global warming to 1.5\textdegree C demands rapid, far-reaching transformations across all sectors of society~\citep{ipcc2023}. 
Achieving such transformations depends not only on scientific evidence, accurate change measurement, and policy design but also on public understanding, engagement, and political will - factors that are increasingly shaped by social media~\citep{williams2015, weller2025using}. 
Platforms such as X (formerly Twitter), Instagram, and Facebook have become primary arenas for climate communication, collectively hosting billions of posts that show how individuals perceive climate risks, attribute responsibility, and evaluate proposed solutions~\citep{pearce2019}. 
Within this discourse, visual content plays a disproportionately powerful role: images attract more engagement than text-only posts~\citep{Casas2019}, evoke stronger emotional responses~\citep{chapman2016, grabe2009image}, and enhance long-term recall of information~\citep{paivio1968pictures}.
Beyond cognition and affect, visuals actively shape attitudes and opinions through framing effects~\citep{powell2015, todorov2009, rodriguez2011levels}, while also raising awareness and calling for action \citep{isaaclit}.
In the context of climate change, these properties make images uniquely influential in shaping public understanding of risk, responsibility, and urgency~\citep{obrien2015}.

Despite the demonstrated power of imagery in climate communication, the vast majority of computational research on online climate discourse has focused on textual data, e.g.~\citep{falkenberg2022growing,vanVliet2021, olteanu2015comparing, effrosynidis2022climate,samantray2019credibility, littman2019climate}, analysing sentiment in tweets, topic-modelling news articles, or tracking the spread of misinformation through linguistic cues~\citep{cody2015, chen2021}. 
This text-centric bias leaves a critical blind spot: the millions of photographs, infographics, satellite composites, memes, and data visualisations shared daily on social media remain largely unexamined at scale. 
Within recent years, the field of visual climate change analysis has started to grow, e.g~\citep{ bravo2025global, bravo2025viral, o2020more,hayes2021greta,hayes2025visual, blewett2025beyond, born2019bearing, obrien2015, zhou-etal-2025-media}.
Manual content-analysis methods, while offering rich qualitative insight, are inherently connected to high costs and studies as thus often focusing on small, non-representative samples, e.g.~\citep{rose2016, schafer2023news}. 
The vast volume of climate-related visual content - exceeding 500,000 images shared annually on X alone -  makes comprehensive manual analysis extremely costly. 
While approaches such as sampling, crowdsourcing, or hybrid human–machine pipelines can partially address this challenge, recent advances in computer vision models offer a scalable pathway for systematically analysing this data. 
This is essential to provide researchers and policymakers with systematic evidence on which visual content dominates public discourse, how it varies across regions and platforms, and whether specific visual strategies effectively mobilize public concern or, conversely, foster disengagement and fatigue~\citep{wang2018, born2019}.
Understanding the visual dimension of climate discourse is essential for designing effective science communication strategies; evidence suggests that the visual themes of climate change (e.g., polar bears versus human impacts, distant futures versus local consequences) significantly influence audience perception of urgency, agency, and self-efficacy~\citep{obrien2015, corner2015}.

Recent advances in deep learning have made such large-scale visual analysis technically feasible. 
Convolutional neural networks (CNNs) and vision-transformer (ViT) architectures have achieved human-level performance on general image classification benchmarks~\citep{he2016, dosovitskiy2021}, while image-text aligned model such as CLIP enables the zero-shot classification of data of varying content~\citep{radford2021}. 
More recently, vision-language models [VLMs] such as ChatGPT~\citep{chatgpt}, Gemini~\citep{gemini3}, and Qwen~\citep{qwen3technicalreport} have demonstrated remarkable capacities for jointly reasoning over visual and textual content, opening new possibilities for zero-shot image classification, captioning, and contextual interpretation without task-specific fine-tuning~\citep{chatgpt, gemini3}.
These capabilities have already been successfully applied in clustering, where VLM's image descriptions have shown to detect pattern in climate change communication~\citep{arminio2025leveraging}.
A comparative analysis of existing methods for classifying the full typological spectrum of climate change imagery on social media - spanning scientific visualisations, protest photography, nature and wildlife imagery, memes, infographics, and personal narratives - is our contribution.
This includes model selection, prompting, and evaluation protocol, while discussing the impact of application-specific taxonomies.

\subsection{Problem Description}
\label{sec:problem}

The emergence of powerful vision-language models has raised an appealing prospect for social science researchers: VLMs can generate taxonomy-specific image labels used for content analysis. 
However, the reliability of such an approach is far from established. 
It is therefore necessary to rigorously evaluate the conditions under which automated methods produce sufficiently accurate annotations - and to have metrics which fit the application.
To address this, we organise our investigation around four research questions that collectively examine the full pipeline from model selection and prompting through taxonomy design, validation, and failure analysis. 
For the evaluation, we have created a manually validated dataset of 50k climate change images from X that spans five annotation dimensions: animal content, climate change consequences, climate action, image setting, and image type.

\paragraph{RQ1: Can VLMs generate annotations for visual climate discourse analysis?}
We investigate whether current VLMs can reproduce human-quality annotations when provided with a social science codebook. 
Beyond individual image-level accuracy, we examine whether the predicted label distributions align with the ground truth at the corpus level to establish whether the labels can be used for trend analysis.
 
\paragraph{RQ2: How does taxonomy design influence classification performance?}
Social science codebooks are theoretically grounded and differ fundamentally from standard computer vision benchmarks, as they may be culturally contingent or hierarchically structured.
Within a single annotation scheme, a researcher might distinguish between broad classes such as ``land mammals'' alongside narrow ones like ``polar bears''~\citep{bravo2025viral}, or between ``forest/trees'' and the more specific ``rain forest''~\citep{mcgarry2025fire}. Such hierarchical entanglement between categories can introduce systematic confusion for classifiers, which may struggle to distinguish semantically overlapping classes.
We investigate this effect by analysing confusion patterns across categories with varying degrees of semantic proximity.
Additionally, we compare confusion patterns between popular images in \textit{ClimateCT} and the larger, noisier \textit{ClimateTV} to assess whether data sampling influences the observed entanglement.
 
\paragraph{RQ3: How relevant is the choice of CV model?} 
We benchmark six promptable VLMs and 15 zero-shot CLIP-like models across all five annotation dimensions on both \textit{ClimateCT} and \textit{ClimateTV}. 
For each dimension, we determine which model configuration best approximates human-quality labels. 
We also assess the relationship between model size and performance, and whether the performance gap between curated (\textit{ClimateCT}) and heterogeneous (\textit{ClimateTV}) data is consistent across architectures.
 
\paragraph{RQ4: What effect does prompting have on performance?}
We systematically ablate prompt components across all five super-categories, varying (i)~the length of category descriptions (labels only vs.\ extended explanations with examples), (ii)~the inclusion of super-category definitions, (iii)~the explicit triggering of chain-of-thought reasoning, and (iv)~spatial localisation instructions directing the model where to look in the image (foreground or the image as a whole).
To account for prompt sensitivity, we generate paraphrases of each prompt configuration and report performance with mean and standard deviation across prompt variants.

\medskip
\noindent Taken together, these four research questions provide a comprehensive evaluation framework that moves beyond the simple question of ``how accurate is model X'' to address the methodological decisions - taxonomy design, model selection, prompting strategy, and validation protocol - that social science researchers must navigate when integrating automated image classification into their workflows.

\medskip
\noindent Our main contributions can be summarised as:
\begin{enumerate}
    \item We systematically review visual climate change research and highlight that current studies either rely on manual expert annotations of a small subset or 1-2 method evaluations. 
    Prompt components beyond explanations, such as focus localisation, in-depth task descriptions, and thinking, have not yet been discussed. 
    \item We provide a comparative analysis of VLMs in the context of social science research of climate change communication [RQ1] with a special focus on taxonomy [RQ2]. We investigate model selection [RQ3] and prompting [RQ4] for five dimensions (animals, climate change consequences, climate action, image setting, and image type) on representative data. For evaluation, we have manually validated 250k image labels \footnote{We share image Twitter IDs and labels for transparency and reproducibility.}. We find that Gemini-3.1-flash-lite clearly outperforms all other assessed models, with a gap to the best open-weight model of $0-8\%$.
    \item We identify several strategies for VLM annotation in the context of visual climate change communication: create model prompts per super-category, where simpler tasks require simpler prompts. Moreover, we suggest to use task-specific evaluation metrics, such as Jensen Shannon divergence for trend-analysis.
\end{enumerate}

\section{Literature Review}
\label{sec:relatedwork}

\subsection{Visual Climate Change Communication}
\label{sec:visual_climate_comm}

Early work by~\citet{obrien2015} established that the choice of climate imagery significantly affects public engagement, demonstrating that images depicting everyday human impacts elicit stronger feelings of personal relevance than iconic but distant visuals such as polar bears or melting glaciers. 
This finding has been corroborated and extended by subsequent studies:~\citet{corner2015} distilled seven evidence-based principles for effective visual climate communication, emphasising the importance of showing real people, local impacts, and emotionally authentic narratives.
\citet{born2019bearing} examined the dominance of polar bear imagery in outlets such as National Geographic, arguing that the repeated use of a narrow set of iconic visuals risks rendering climate change as a geographically and temporally remote issue, thereby reducing perceived agency among audiences.
\citet{o2020more} manually annotated 1,278 climate images across multiple platforms, highlighting the formulaic nature of visual coverage and calling for more diverse and inclusive representations. 

More recent work has expanded the scope of visual analysis beyond traditional news media to social media platforms. 
\citet{hayes2021greta} analysed the visual representation of climate protest in UK media and the Getty Images archive, revealing how the framing of activists shapes public perception of the climate movement. 
To this end, the authors manually annotated 1,869 images.
\citet{leon2022social} manually annotated 380 images with one label per image with the goal of predicting post engagement.
\citet{hayes2025visual} and~\citet{blewett2025beyond} have further explored how visual framing varies across digital media ecosystems, while~\citet{mcgarry2025fire} analysed 350 images from X (formerly Twitter) to investigate wildfire communication, which they manually coded with respect to 6 dimensions, i.e. object type, actor, iconography, affect, voice, and information sharing.
\citet{bravo2025viral} have investigated characteristics of images which are part of popular tweets. 
To this end, they have manually annotated 662 images for 13 dimensions (13 labels/image).

Collectively, this body of work demonstrates that visual imagery is not merely an aesthetic concern but a substantive driver of public attitudes, political engagement, and behavioural intentions regarding climate change. 
These studies rely on manual content analysis of relatively small, purposively sampled image collections.

\subsection{Computational Analysis of Climate Discourse}
\label{sec:computational_climate}

Computational approaches to climate change discourse have proliferated in recent years, though they remain overwhelmingly text-centric. 
Sentiment analysis of climate-related tweets has been used to track public mood in response to policy announcements and extreme weather events~\citep{cody2015}, while topic modelling has revealed the thematic structure of online climate conversations~\citep{chen2021}. 
Larger-scale studies have mapped the network structure of climate discourse~\citep{williams2015, falkenberg2022growing, cann2021ideological}, identified polarisation dynamics~\citep{samantray2019credibility}, and constructed comprehensive datasets of climate-related social media posts for downstream analysis~\citep{littman2019climate, effrosynidis2022climate}. 
\citet{vanVliet2021} and~\citet{olteanu2015comparing} have further demonstrated the value of computational methods for understanding how climate narratives are constructed and contested across platforms.

While these approaches have yielded substantial insights, the visual dimension of climate discourse has only recently begun to attract computational attention. 
\citet{yan2025multimodal} analysed WeChat posts with 32,327 images, which were embedded into the CLIP-ViT-B/32~\citep{radford2021} space and were fed as input to BERTopic~\citep{bertopic}.
\citet{zeng2024understanding} analysed 7,564 TikTok videos related to climate change using Google Vision API~\citep{googlecloudvision} for object detection, face emotion recognition, and text extraction.
Subsequently, they assigned one of four labels (person-dominated, nature-dominated, text-dominated, and mixed) to each video.
Additionally, they performed topic modelling using BERTopic~\citep{bertopic} using extracted text and video metadata as input.
\citet{mooseder2023social} applied image classification methods to climate-related social media content, representing one of the first systematic attempts to move beyond manual coding. 
In their work, they embedded more than 2 million X images in the VGG16 embedding space~\citep{vgg16} and applied K-means clustering ($k=5,000$) to detect common visual elements.
In order to validate their results, they inspected 100 randomly selected images per cluster to assign a label to all images in this cluster.
\citet{bravo2025global} extended this line of work by examining the emotional and engagement effects of different visual frames on Twitter/X on a global scale. 
In their work, they annotated over 3 million images using zero-shot CLIP~\citep{radford2021} with 5 custom probability thresholds, which have been manually validated. 
\citet{zhou-etal-2025-media} have analysed 1,184 Reddit memes manually annotated with media frames (real, hoax, cause, impact, action allocation, action propriety, action adequacy, action prospect), stances (convinced, sceptical, neither), humour, personalisation, and responsibility.
They have experimented with VLMs (LLaVA \& Molmo) and LLMs (Mistral \& Qwen) ability to predict their labels in a 4-shot setting, all model choices have 8B parameters.
\citet{qian2024convergence} use pre-trained VGG-hybrid features to cluster images to compare content distribution across platforms.
They focus on five content categories: 1) climate activism, 2) nature landscape/wildlife, 3) technology, 4) data visualization, and 5) infographics.
Moreover,~\citet{zhang2024image} compare image features for their suitability for clustering social media images. 
The authors compare clustering of bag-of-visual-words, DeepCluster using AlexNet and VGG features, and pretrained ResNet and VGG features.
Similarly,~\cite{prasse2023towards} compared VLMs embeddings spaces for clustering climate change images and~\citet{tornberg2025aesthetics} cluster CLIP features of climate change images.

Despite these advances, existing computational studies of climate change images have not systematically evaluated the capabilities of modern vision-language models for this task. 
Our work builds directly on this emerging literature by providing a comprehensive, multi-dimensional evaluation of VLM-based annotation across the full spectrum of climate change imagery.

\subsection{Vision-Language Models for Image Classification}
\label{sec:vlms}

The rapid evolution of computer vision over the past decade has progressively lowered the barrier to large-scale image analysis.
A pivotal development was the introduction of contrastive vision-language pretraining through CLIP~\citep{radford2021}, which enabled zero-shot image classification by learning joint representations of images and natural language descriptions. 
CLIP and its successors showed that models trained on large-scale image-text pairs from the web could generalise to novel visual domains without task-specific fine-tuning, a property of particular value when labelled training data is scarce or expensive to produce - as is typically the case in social science research contexts.

The subsequent emergence of large-scale vision-language models (VLMs) has further expanded the frontier. 
Models such as GPT-5~\citep{chatgpt}, Qwen~\citep{qwen3technicalreport}, and Gemini~\citep{gemini3} integrate visual perception with sophisticated language reasoning, enabling not only classification but also detailed image description, visual question answering, and contextual interpretation. 
These capabilities are especially relevant for social media imagery, where the meaning of a visual is often contingent on overlaid text, cultural context, and platform-specific conventions that purely visual classifiers cannot capture. 
VLMs have been used for 500k news media frame analysis~\citep{arora2025multi} for 15 generic frames appropriate for analysis of US news. 
They use Mistral-7B for text annotations and Pixtral12B for image annotations and compare prompts with short, medium, and long frame descriptions and output formats.
\cite{arminio2025leveraging} use VLMs to map images to text descriptions which are the used for clustering.
The authors compare VGG16 clustering to the clustering of VLMs (GPT-4-turbo~\citep{openai2024gpt4technicalreport} and LLaVA~\citep{llava}) image description's text embeddings.
\cite{piqueras2025imageworthktopics} fit their model vSTM to the images' CLIP~\citep{radford2021} embeddings.

However, VLMs also present well-documented challenges~\citep{basugeodiv, li2025ravenea, das2026more, snaebjarnarson2025taxonomy}. 
Studies have shown that promptable models can be sensitive to prompt formulation, producing inconsistent classifications when the same image is described using different phrasings~\citep{radford2021, lu2024prompts, errica2025did}. 
Hallucination - the generation of confident but factually incorrect descriptions - remains a concern, particularly for domain-specific content where pretraining data may be sparse. 
Furthermore, the performance of VLMs on subjective or culturally contingent classification tasks, such as distinguishing theoretically grounded visual themes of climate change, has not been systematically assessed. 
Our work directly addresses this gap by evaluating multiple VLMs across five fine-grained, theory-grounded annotation dimensions that are central to climate communication research, examining both model selection and prompting strategies.

\subsection{Annotation and Evaluation in Social Science Contexts}
\label{sec:annotation_eval}

The use of automated methods for data annotation in social science raises methodological considerations that differ from standard computer vision benchmarks. 
In manual content analysis, reliability is typically established through inter-coder agreement metrics such as Krippendorff's alpha or Cohen's kappa~\cite{krippendorff2018content}, with codebooks developed iteratively to capture the nuanced, context-dependent categories that social scientists employ~\citep{rose2016}. 
\citet{bravo2025viral} designed their codebook this way and later employed it when using computational methods~\citep{bravo2025global}.


Prior work on using language models for annotation in the social sciences has demonstrated both promise and pitfalls. 
Studies in political science and communication have shown that large language models can approximate crowd-worker annotations for text classification tasks, though performance varies substantially across categories and domains~\citep{pangakis2024,egami2024, gilardi2023}.
\cite{prasse2025spy} employed clustering for image annotation and compared several embedding spaces for their suitability in the application.
The extension of these findings to visual annotation via VLMs is a natural but non-trivial step: images introduce additional sources of ambiguity (e.g., composition, colour, spatial relationships) that text-based annotation frameworks do not address. 
Moreover, the prompting strategy used with VLMs - including the level of detail in category descriptions, the use of examples, and the framing of the task - can significantly influence classification outcomes, yet systematic guidance on prompt design for social science annotation tasks remains limited.
Our evaluation is designed to provide the practical guidance that is currently missing from this literature, offering specific recommendations on model selection, prompt design, and expected performance across the five annotation dimensions.

\section{Data Choice and Experimental Set-up}
We are publishing our paper along with the code and the X id's of the analysed images (incl. labels).
For each image, we share when it was posted and how the label was created (manual vs. automatic).
According to X's terms of service, users retain all rights to their content submitted to X.
By uploading content, users grant X a worldwide, non-exclusive, royalty-free license to share their data. 
Researchers are allowed to access X data through the API and conduct analysis while respecting the users' reasonable expectation of privacy and refraining from investigating sensitive user information.
Given that individuals' perception of climate change may be related to their political affiliations and beliefs, we conduct our analysis exclusively on the images and disregard the authors of the tweets.
Sharing X post ids complies with user post deletions and content removal.

\subsection{Annotation Scheme}
\label{sec:anno}
In the following, we first outline the proposed annotation scheme.
We argue that high-dimensional topics like climate change require high-dimensional annotation schemes, i.e., multiple annotations per image. We have taken a subset of the image labels from~\citep{bravo2025viral}, which are in line with previous works, e.g.~\citep{rebich2015image, mooseder2023social}. 
We have combined the super-categories \textit{human activity} and \textit{adaptation and mitigation} in the super-category \textit{climate action}. 
The annotation statistics confirm, that this combination did not have an effect on the quality of the code book, as annotation quality metrics remained constant (see \Cref{app:ct_stats}).
In this paper, we focus on the following five super-categories: \\

\noindent\textbf{(i) \textit{animals}}: pets, farm animals, polar bear, land mammal, sea mammal, fish, amphibian/reptile, invertebrates, birds, other/mixed animals, no animals 

\noindent\textbf{(ii) \textit{climate action}}: protest, politics, sustainable energy, fossil energy, other climate action, no climate action 

\noindent\textbf{(iii) \textit{consequences}}: biodiversity loss, covid/health, drought, floods, wildfires, other extreme weather, melting ice, sea level rise, rising temperature, human rights, economic consequences, other consequences, no consequences

\noindent\textbf{(iv) \textit{setting}}: residential/commercial, industrial, agricultural/rural, indoor space, arctic/antarctic, ocean/coastal, desert, forest/jungle, other nature, outer space, other setting, no setting

\noindent\textbf{(v) \textit{type}}: photo, photo collage, illustration, meme, data visualisation, screenshot/text, infographic, poster/event invitation, other type
\\

The annotation scheme is neither non-overlapping nor equidistant while containing varying degrees of diversity.
This a defining difference from other computer vision annotation schemes, but highly relevant in the context of social media research.
Hence, we refer to our labels as categories instead of classes, as they do not fit into the common understanding of classes, in line with previous works~\citep{arora2025multi, intentonomy}.
Examples of the nature of annotation categories are:
While \textit{animals} contains various general animal classes, while the \textit{polar bear}, the most popular animal in climate change visuals~\citep{born2019bearing}, is given its dedicated category. 
Moreover, in most categories, we have included an \textit{other} and a \textit{none} category to cover the visual themes that are to date under-explored by the literature.

\subsection{Data Selection}
We select our data sample as X (formerly Twitter) posts that meet the following criteria: (1) search term: contain either the term "climate change", "climatechange", or "\#climatechange", (2) image: at least one visual, and (3) time frame:  was shared in the time period of 2019 - 2022 (Jan 1 - Dec 31).
We scraped the data in Q1/2023 and downloaded data for 4 complete years.
We use the exact terms as used in social science research~\citep{bravo2025viral}.
Our sample has also been used for social science research and is thus an excellent starting point for method inspection.
The tweets and corresponding information have been scraped via the API using the free academic access (while it was still available). 
\\
We created two distinct subsets:

\paragraph{Climate Change Twitter [\textit{ClimateCT}, 1,038 images]} contains popular images with high quality annotation from domain experts (social scientists). To this end, we leverage~\citep{bravo2025viral}’s manual annotation of 662 images within our sample.
These images were part of the top 10 most popular tweets for each month (460 tweets, 1-4 images/tweet). The category labels were confirmed as relevant based on this subset by~\citep{bravo2025viral}.
We manually extended this sample by 376 images to ensure a fixed minimum number of images per label (21 images/label). These additional images were then manually annotated by 2 independent domain experts
(metrics can be found in \cref{app:ct_stats}).  

\paragraph{Climate Twitter Visuals [\textit{ClimateTV} 1,220,784 images]} contains images for all tweets not contained in \textit{ClimateCT}. In this sample, we include only the first image shared alongside the tweet and not all images. This is also a standard practice from social sciences taken from~\citep{bravo2025global}.
In order to clean the data, we used the DataComp pipeline~\citep{datacomp} to deduplicate the images and remove NSFW content.
The final subset of images is labelled automatically. 
This mapping process is two-stage: we first generate image descriptions using Gemini~\citep{team2023gemini} and then employ the Llama~3.0 model~\citep{llama3herd} to map them to our labels. 
This process is done separately for each super-category. 
The prompts can be found in the code base.
In order to assess the labelling quality, we have 3 independent QualityMatch workers manually confirm our labels.
We validate the automatic annotations by comparing the results against manually confirmed samples (Statistics can be found in \cref{app:tv_stats}).
With the support of QualityMatch, we acquire label validations for 50,000 images, i.e. 4\% of the entire dataset.
In order to capture the bandwidth of the dataset, we select 45,000 images based on cluster centroids using KMEANS on the images' embeddings in dinov2-vit-s14 space.
The remaining 5,000 images are sampled randomly from the remaining images.


\subsection{Model Selection and Evaluation}

This study has the goal of advancing social science research for large scale data analysis. 
We evaluate vision-language models [VLMs] capability to capture pattern in image data.
Our evaluation comprises zero-shot models such as CLIP by~\citep{radford2021learning} with various backbones and training datasets as provided in the OpenCLIP library by~\citep{openclip}.
In addition, we employ promptable models with both open-weights, i.e. Qwen-3~\citep{qwen3technicalreport}, Moondream by~\citep{moondream}, and Gemma by~\citep{gemma}, and closed-weights, i.e. ChatGPT-5.4-mini by~\citep{chatgpt} and Gemini-3.1-flash-lite by~\citep{gemini3}.
A complete list of models employed can be found in \cref{app:models}.
For both model types, we ablate the selection of zero-shot labels and prompts, respectively, on \textit{ClimateCT}.
More information regarding input engineering can be found in \cref{subsec:promptabl}.

We ablate and evaluate the models based on mean-class accuracy due to the long-tail distribution of classes, as discussed in \cref{app:dataset}.
In the evaluation, we further report accuracy, precision, recall, and F1 score.
%
We also computed $\chi$-squared distance between the ground truth and predicted label distribution, along with total variation of distribution and Jenson Shannon divergence.
We perform the evaluations exclusively on manually (confirmed) labels.

\subsection{Prompt Ablation}
\label{subsec:promptabl}

For promptable VLMs, we compare (1) the effectiveness of adding explanations besides the category labels, (2) defining the super-category for the current tasks, (3) explicitly triggering thinking, and (4) telling the model where to look (foreground, background, everyone).
We create paraphrases of the prompts using Qwen (Qwen3-VL-8B-Instruct) and use five versions of each prompt for the ablation and three for benchmarking.
The prompt for paraphrasing is ``please paraphrase this prompt without changing the category names and their mapping. This is the prompt: [prompt] ''.
For the paid models, we evaluate on a single prompt for the large \textit{ClimateTV} dataset, for \textit{ClimateCT} we use three prompts.
We can thus report model performance with a mean and standard deviation. 
Within zero-shot models, we compare (1) the inclusion of examples and handling of categories (tokenising/treating as a string), (2) the point of aggregation (before/after inference), and the choice of template (clip standard \textit{"A photo of an x "} vs. ImageNet template w/ 80 versions averaged). 
The best configuration is (1) no examples and category tokenisation, (2) aggregation before inference, and (3) the ImageNet template.
This ablation compares 15 model predictions.

\section{Analysis and Results}
\label{sec:analysis}

In this section, we present the results for each of the four research questions with the goal of refuting or confirming the research questions.
Our analysis has shown that Gemini-3.1-flash-lite has superior performance for this task compared to the other analysed models.
The full benchmarking analysis can be found in \cref{subsec:RQ3}.

\subsection{RQ1: Can VLMs replace manual annotation for visual climate discourse analysis?}
\label{subsec:RQ1}

\begin{wraptable}{r}{0.45\textwidth} 
\centering
\small
\sisetup{
    round-mode = places,
    round-precision = 2,
    table-format = 1.2
}
\begin{tabular}{l S}
\toprule
\textbf{Super-Category} & {\textbf{Macro Acc.}} \\ 
\midrule
Animals          & 0.76 \\
Climate action   & 0.70 \\
Consequences     & 0.67 \\
Setting          & 0.61 \\
Type       & 0.67 \\ 
\bottomrule
\end{tabular}
\caption{Best macro accuracy for \textit{ClimateTV} shows medium to high performance, with \textit{animals} as the best and \textit{setting} as the worst super-category [Gemini-3.1-flash-lite].}
\label{tab:bestacc_vlm}
\end{wraptable}

\Cref{tab:bestacc_vlm} shows that the accuracy of the Gemini-3.1-flash-lite annotation is at a moderate performance level. 
Especially object-oriented super-categories such as \textit{animals} show a high level of mean class-wise accuracies [macro accuracy].
Both climate change related super-categories \textit{consequences} and \textit{climate action} show a medium performance, and \textit{setting} and \textit{type} have the lowest mean category-wise accuracy.
\Cref{app:gemini} shows that precision which are near perfect $\geq .96$ for the \textit{None} class in both \textit{climate action} and \textit{consequences}. 
This indicates that VLM predictions can be reliably used to filter out irrelevant images.
Given that recall is lower ($.57-.31$), this filtering will not be complete, as still not related images remain in the sample.
Contrarily, \textit{rising temperatures}, \textit{sustainable energy}, and \textit{protest} have high recall scores ($.74-.79$), while precision is lower
 ($.14-.60$).
VLM predictions exhibit high sensitivity but lower specificity.
We have used prompt styles as identified in \cref{subsec:RQ4}, where we analyse which level of detail is needed for which super-category.

\begin{table}[ht]
\small
\centering
\caption{Comparison of Statistical Divergence: \textit{ClimateTV} \& \textit{ClimateCT} show the same trends, with higher distributional alignment on the popular images in \textit{ClimateCT}.}
\label{tab:combined_statistical_divergence}
\begin{tabular}{lrrr|rrr}
\toprule
& \multicolumn{3}{c}{\textbf{ClimateTV}} & \multicolumn{3}{c}{\textbf{ClimateCT (Mean $\pm$ Std)}} \\
\cmidrule(lr){2-4} \cmidrule(lr){5-7}
\textbf{Task} & \multicolumn{1}{c}{$\chi^2$} & \textbf{TV} & \textbf{JSD} & \multicolumn{1}{c}{$\chi^2$} & \textbf{TV} & \textbf{JSD} \\
\midrule
Animals        & $1,976.09$    & $.02$ & $.00$ & $11.11 \pm 1.12$   & $.02 \pm .00$ & $.00 \pm .00$ \\
Climate Action & $86,329.38$   & $.40$ & $.15$ & $97.83 \pm 7.20$   & $.11 \pm .01$ & $.01 \pm .00$ \\
Consequences   & $1,114,993.17$ & $.46$ & $.23$ & $1001.60 \pm 195.37$ & $.39 \pm .03$ & $.13 \pm .02$ \\
Setting        & $36,411.37$   & $.44$ & $.23$ & $93.58 \pm 22.85$  & $.11 \pm .01$ & $.02 \pm .01$ \\
Type           & $1,991.97$    & $.09$ & $.01$ & $58.08 \pm 3.46$   & $.05 \pm .00$ & $.01 \pm .00$ \\
\bottomrule
\addlinespace
\multicolumn{7}{l}{\small \textit{Note: TV = Total Variation; JSD = Jensen-Shannon Divergence. Leading zeros removed.}} \\
\end{tabular}
\label{tab:distr}
\end{table}

However, a critical distinction emerges when evaluating the model's ability to capture discourse-level trends rather than individual labels. 
When comparing the predicted label distributions against ground truth, the categories \textit{animals} and \textit{type} show the strongest alignment, with Jensen-Shannon (JS) divergences $\leq .10$, as shown in \Cref{tab:distr}.
These distribution alignment trends remain consistent across datasets. 
Notably, the divergence is generally higher on the larger, more heterogeneous \textit{ClimateTV} dataset than on the smaller, curated \textit{ClimateCT} dataset.
This suggests that while individual images in complex categories like \textit{type} are frequently misclassified, the errors are non-systematic. 
Consequently, the aggregate category distribution can still be reliably extracted, suggesting that VLMs may serve as a viable proxy for population-level discourse analysis even when item-level accuracy is sub-optimal.
For \textit{animals} and \textit{type}, the weighted accuracy is higher than the macro-accuracy, whereas the opposite is true for the other super-categories.
This indicates that if category accuracies vary too much, the distribution may not be represented well using the predictions.

\subsection{RQ2: How does taxonomy design influence classification performance?}
\label{subsec:RQ2}
We see consistent confusions between semantically close categories, i.e. \textit{floods} and \textit{sea level rise} or \textit{biodiversity loss} and \textit{droughts} in \Cref{tab:conf_tv}.
Moreover, we observe that \textit{other consequences} appears frequently in the top 10 most confused categories, as the borders between concrete consequences and general, ``other" consequences appear hard to distinguish between.
In every super-category, the broad ``other" and ``no" categories are part of the top 10 confusions to further confirm the observation.
   
\begin{table}[ht]
\centering
\caption{Top 10 most frequent classification errors on \textit{ClimateTV} [Gemini-3.1.-flash-lite]. Confusion scores indicate the proportion of true classes predicted as the indicated category.}
\label{tab:confusion_analysis}
\small
\begin{tabular}{rllc}
\toprule
\textbf{\#} & \textbf{True Class} & \textbf{Predicted As} & \textbf{Confusion Score} \\
\midrule
1  & Human rights                 & Other consequences           & $.30  $ \\
2  & Sea level rise               & Floods                       & $.24  $ \\
3  & Biodiversity loss            & Drought                      & $.17  $ \\
4  & Other extreme weather events & Floods                       & $.14  $ \\
5  & No consequences              & Other consequences           & $.13  $ \\
6  & Other consequences           & Other extreme weather events & $.11  $ \\
7  & Biodiversity loss            & Wildfires                    & $.09  $ \\
8  & Covid \& general health      & Rising temperature           & $.09  $ \\
9  & Other consequences           & Rising temperature           & $.08  $ \\
10 & Other consequences           & Biodiversity loss            & $.08  $ \\
\bottomrule
\end{tabular}
\label{tab:conf_tv}
\end{table}

\begin{figure}[htb]
    \centering
    \begin{subfigure}{0.45\linewidth}
        \centering
        \includegraphics[width=\linewidth]{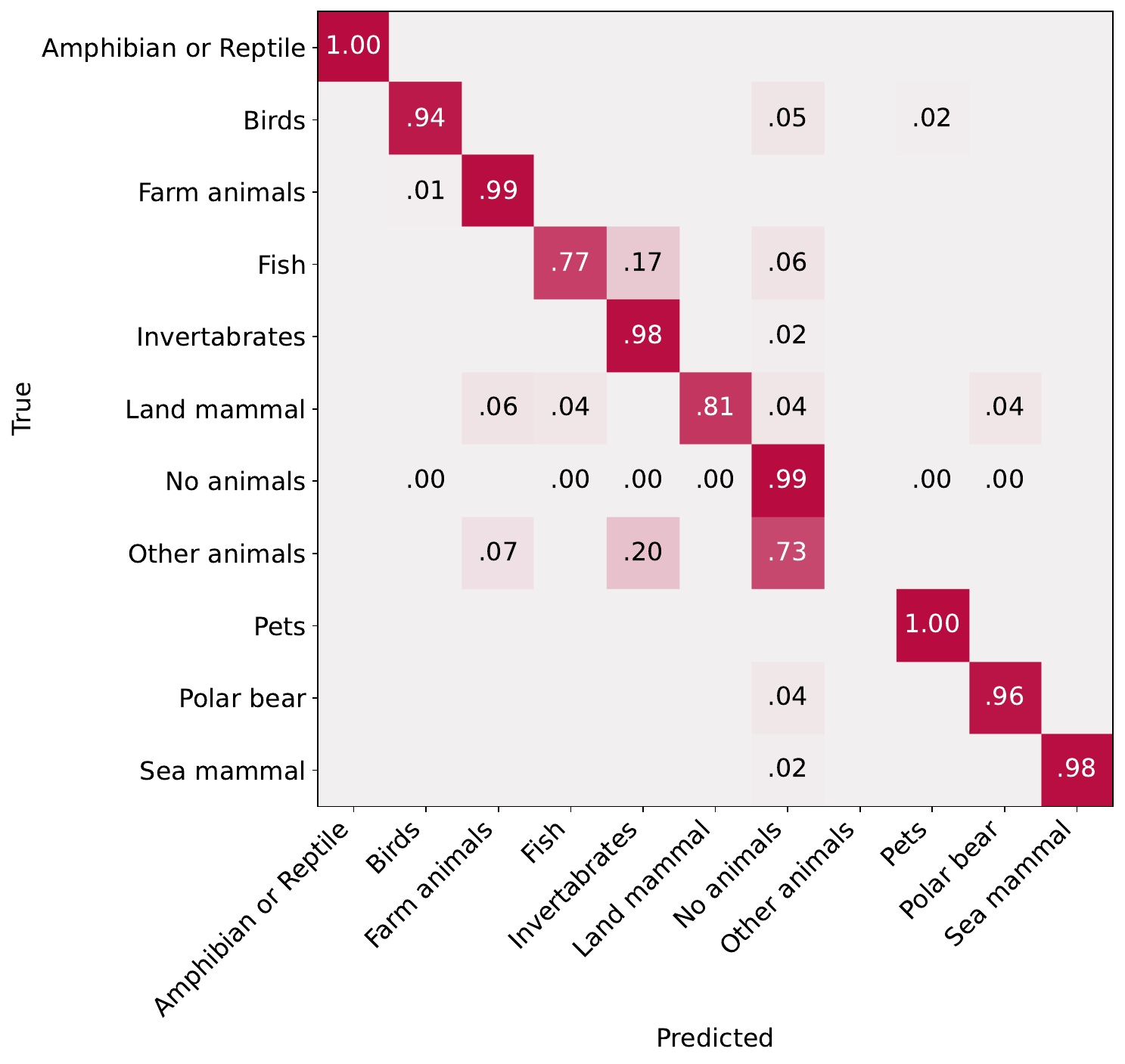}
        \caption{\textit{ClimateCT}}
    \end{subfigure}
    \hfill 
    \begin{subfigure}{0.45\linewidth}
        \centering
        \includegraphics[width=\linewidth]{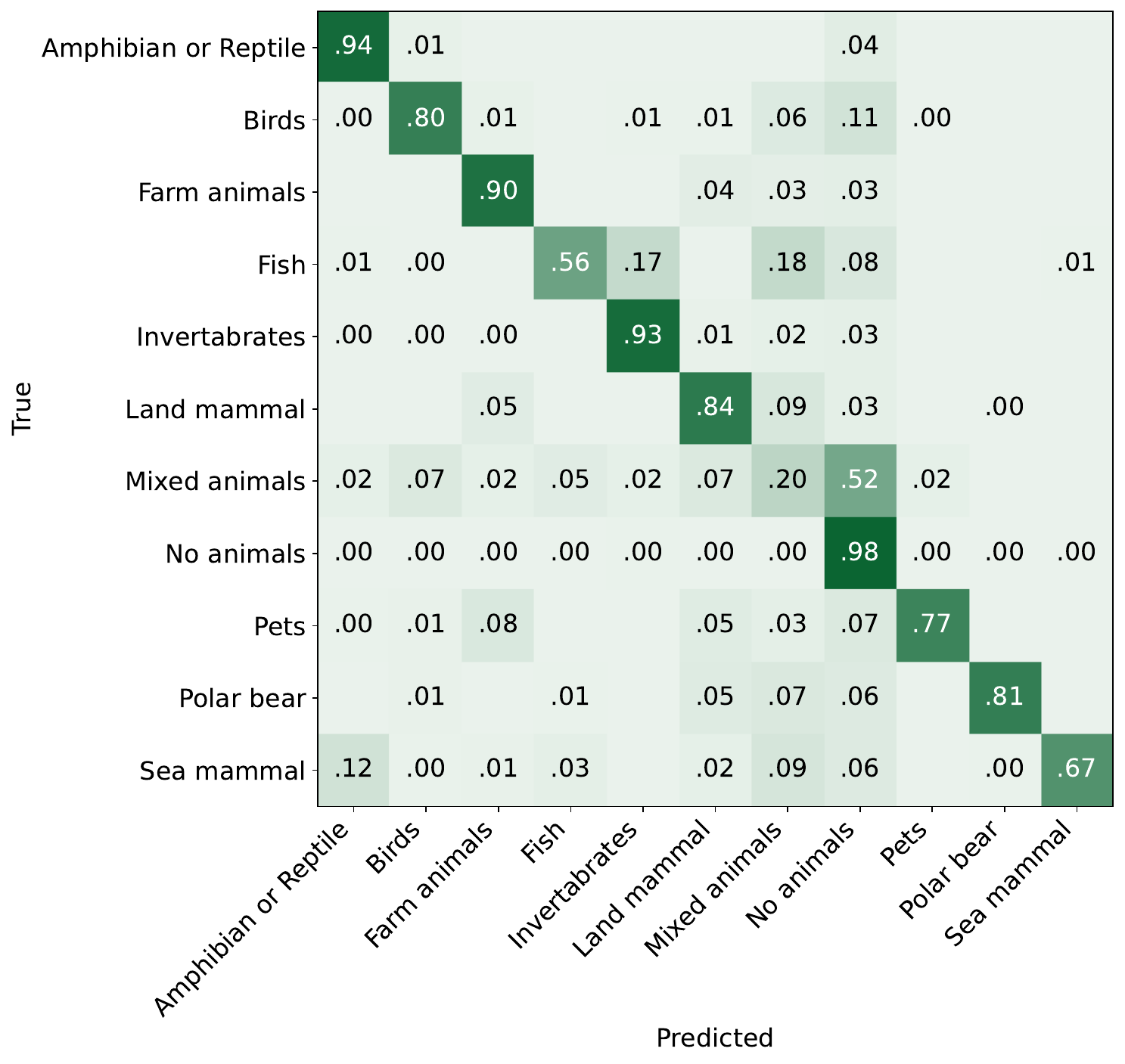}
        \caption{\textit{ClimateTV}}
    \end{subfigure}
    \caption{Confusion matrices for super-category \textit{animals} for both datasets show more confusions in the automatically annotated dataset.}
    \label{fig:confusion_matrix_animals}
\end{figure}

The confusions between overlapping categories of varying specificity exist (comp. \Cref{fig:confusion_matrix_animals}).
We can observe the expected confusion between \textit{polar bear} and \textit{land mammal}, alongside others.
When comparing the confusions, it stands out that they are often non-symmetric: \textit{pets} are confused as \textit{farm animals} but not vice versa.
This can also be observed for \textit{polar bear}, which is confused with \textit{land mammals}, but not vice-versa.
The model appears to be consistent in their predictions for specific categories and appears to overpredict the existence of general categories or more frequently seen categories in the training data.

Overall, it stands out, that on the manually annotated dataset has less confusions overall and also less severe confusions exist.
This makes sense because the automatic labelling and manual confirmation by non-experts holds more degrees of freedom compared to few expert annotators. 
Partly the differences can be explained in terms of dataset size, as \textit{ClimateTV} is 50x larger than \textit{ClimateCT}, however, the different confusion patterns indicate that the annotation pipeline also appears to have an effect.
\medskip
\medskip

\subsection{RQ3: How relevant is the choice of CV model?}
\label{subsec:RQ3}

\begin{wrapfigure}{r}{0.45\textwidth}
    \centering
    \includegraphics[width=\linewidth]{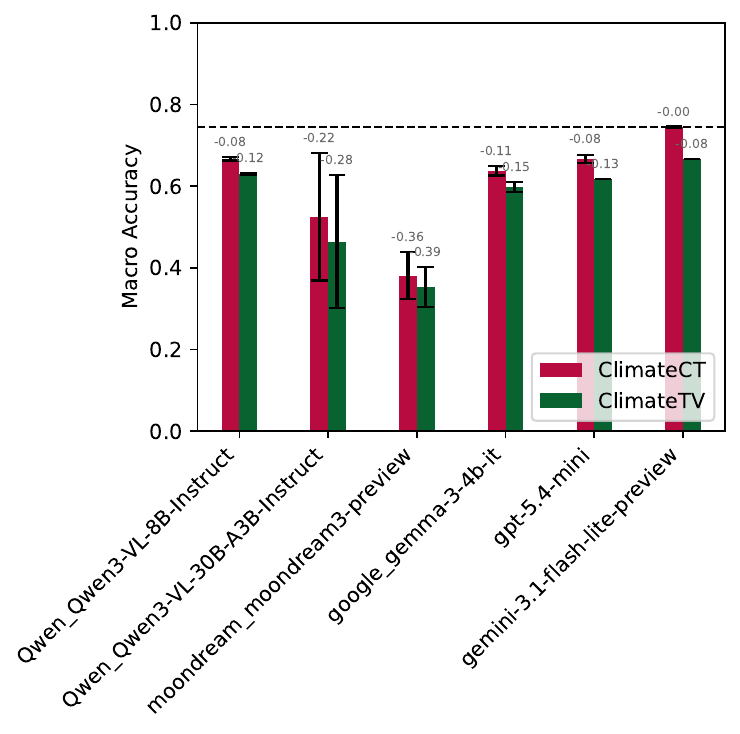}
    \caption{VLM Benchmarking for \textit{consequences} on \textit{ClimateTV} and \textit{ClimateCT}.}
    \label{fig:bench_vlm}
\end{wrapfigure}

We benchmark six promptable VLMs and 15 zero-shot VLMs, such as CLIP, whereof Gemini-3.1-flash-lite is the clearly best model for both datasets and all super-categories.
Between the compared models in \Cref{fig:bench_vlm}, model size does not have a strong impact, as Qwen3 with 8 billion parameters performs on par or better than gpt-5.4 mini.
Moreover, all models performed $0.3-0.8$ better on \textit{ClimateCT} than on \textit{ClimateTV}, in line with our expectation that popular images are easier to classify compared to the random selection of all images shared in the context of climate change (\textit{ClimateTV}).

The performance drop between the 8 billion and the 30 billion Qwen3 model is interesting.
The predictions of Qwen3 (30 billion) also have the largest variance across all analysed models.
This varying performance for different prompts is the reason for the drop in mean macro-accuracy.
One thing that stands out is that model variance is comparable on \textit{ClimateCT} and \textit{ClimateTV}.
Thus, we can reduce the cost of our analysis by only running a single prompt for the paid models on the large \textit{ClimateTV}.
\medskip

\begin{wraptable}{r}{0.45\textwidth} 
\centering
\small
\caption{Best macro accuracy for \textit{ClimateTV} using CLIP-like models shows large discrepancies to promptable VLMs.}
\sisetup{
    round-mode = places,
    round-precision = 2,
    table-format = 1.2
}
\begin{tabular}{l SS}
\toprule
\textbf{Super-Category} & {\textbf{Macro Acc.}} & {$\Delta$} \\ 
\midrule
Animals          & 0.60 & -0.16 \\
Climate action   & 0.41 & -0.29 \\
Consequences     & 0.56 & -0.11 \\
Setting          & 0.46 & -0.15 \\
Type       & 0.44 & -0.20 \\ 
\bottomrule
\end{tabular}
\label{tab:bestacc_clip}
\end{wraptable}

Comparing these results with CLIP-like models, we can observe a large discrepancy, with the best models achieving much lower macroaccuracies, as shown in \Cref{tab:bestacc_clip}.
The analysed CLIP-like models struggle to predict the negative classes correctly, a well known problem for CLIP.
However, when comparing the metrics without considering the \textit{no consequences} category, as shown in \Cref{tab:nocon}, the macro classification accuracy of Gemini-3.1-flash-lite is still superior to MetaCLIP ViT/L's.
While both models have a high precision when predicting \textit{no consequences}, the recall of MetaCLIP is significantly lower compared to Gemini-3.1-flash-lite's.
Given that the data for this application is inherently noisy, a model's ability to remove uninteresting content is highly relevant.

\begin{table}[ht]
\centering
\small
\caption{Comparison of classification accuracy: impact of excluding the majority category (\textit{no consequences}).}
\label{tab:consequences_comparison}
\begin{tabular}{lcccc}
\toprule
& \multicolumn{2}{c}{\textbf{Meta CLIP ViT-L}} & \multicolumn{2}{c}{\textbf{Gemini-3.1-flash-lite}} \\
\cmidrule(lr){2-3} \cmidrule(lr){4-5}
\textbf{Metric} & \textbf{Incl. None} & \textbf{Excl. None} & \textbf{Incl. None} & \textbf{Excl. None} \\
\midrule
Macro Accuracy    & $.55$ & \textbf{$.60$} & $.62$ & \textbf{$.64$} \\
Weighted Accuracy & $.15$ & \textbf{$.66$} & $.35$ & \textbf{$.72$} \\
Precision (None)  & $.99$ & --             & $.99$ & --             \\
Recall (None)     & $.09$ & --             & $.31$ & --             \\
\bottomrule
\end{tabular}
\begin{minipage}{0.8\linewidth}
\vspace{0.5em}
\small \textit{Note: `Excl. None" represents the model's performance when computing the metrics without \textit{no consequences}.}
\end{minipage}
\label{tab:nocon}
\end{table}

\subsection{RQ4: What effect does prompting have?}
\label{subsec:RQ4}

We ablated prompt construction for the five super-categories with respect to the several components: explanations and examples (long), thinking triggering (think), focus localisation to foreground / background (loc), and super-category explanations (super-cat).
The basic prompt consists of a short task description and the category list, including mapping to target answers. 
More details can be found in \cref{subsec:promptabl}.

\begin{figure}[ht]
    \centering
    \includegraphics[width=0.9\linewidth]{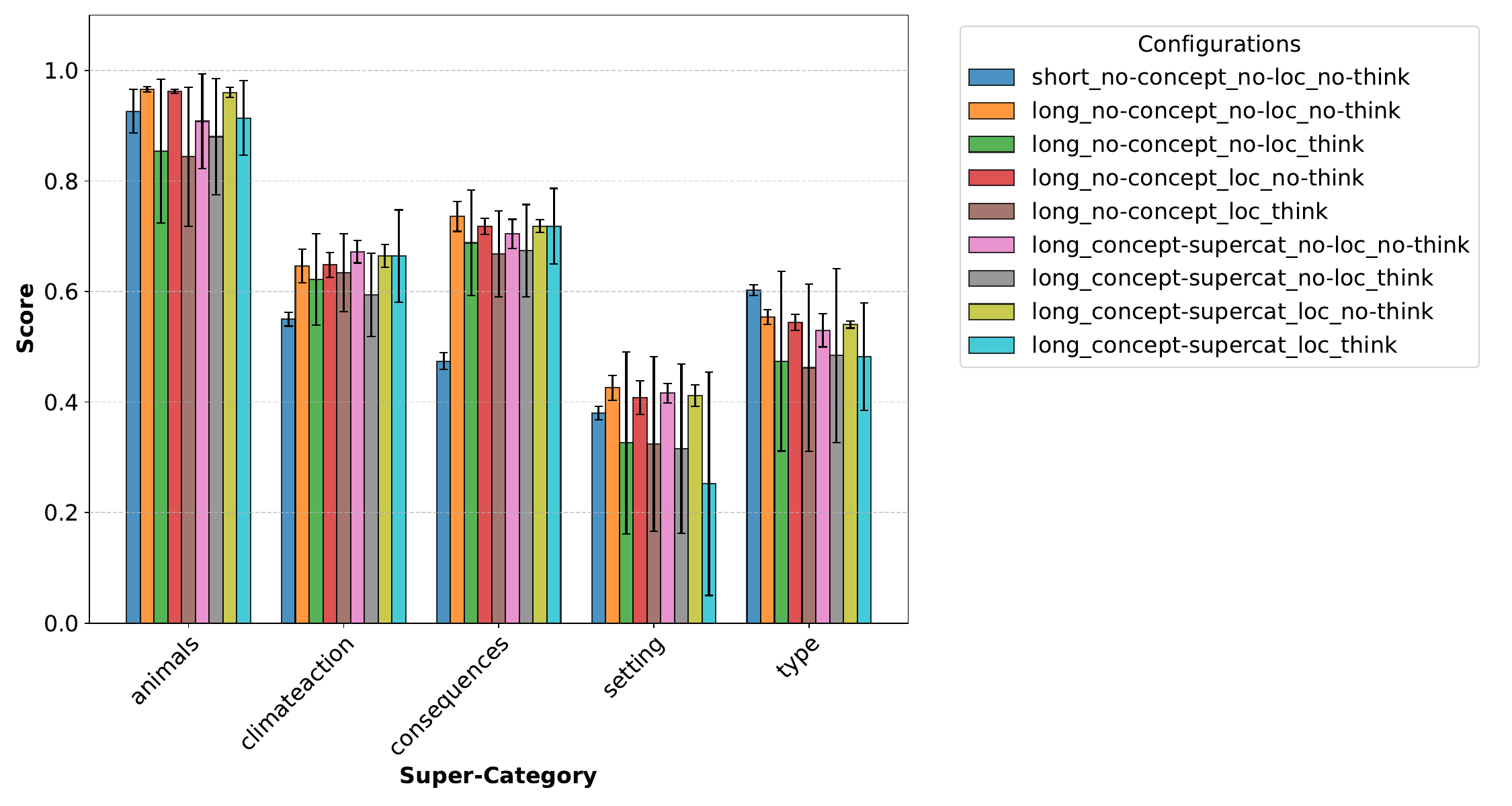}
    \caption{We ablated 9 prompt types for all super-categories and found that short prompts work best for \textit{type}, while all other super-categories benefit from explanations/examples.}
    \label{fig:prompt_ablation}
\end{figure}

We found differences in accuracy for different prompts as \Cref{fig:prompt_ablation} shows.
The prompt component for thinking (e.g. ``think carefully step by step") reduced the performance for most super-categories compared to non-thinking prompts.
An exception for this makes the most involved prompt with all features enabled, i.e., localisation, thinking, conceptualisation of super-category and examples, for the super-categories \textit{climate action} and \textit{consequences}.
However, this prompt configuration has a higher variance compared to the same prompt without thinking. 
We thus removed the thinking component from our prompt for all super-categories.

\textit{Type} has the highest accuracy when using short prompts with only class names and task instructions given.
In contrast to this, all other super-categories benefit clearly from additional information, most strongly \textit{consequences}, where the transition from the short prompt to long variations consistently yields a significant performance boost of $+0.3$.
For most super-categories, adding examples and explanations
is sufficient for high classification accuracy. 
Adding further prompt components does not significantly influence the performance, thus we keep the setting of prompts with explanations for \textit{animals}, \textit{consequences}, and \textit{setting}.
For \textit{climate action}, adding super-category explanations increases the classification accuracy while reducing the variance.
We thus chose this prompt configuration for all further experiments.
Out of the five super-categories, \textit{climate action} is the most abstract, thus the inclusion of super-category level explanations seems appropriate.

For CLIP-like models, the prompt ablations showed that using the \textit{ImageNet} prompt template of 80 variations of ``a photo of [category]" achieves higher performance compared to a single prompt.
All other features did not have a significant impact on prompt performance and were therefore discarded.


\section{Discussion \& Conclusion}
\label{sec:discussion}

This study aimed to evaluate whether vision-language models can serve as reliable tools for automated visual discourse analysis in the context of climate change communication on social media. 
Through a systematic investigation of four research questions - spanning annotation replacement, taxonomy design, model selection, and prompt engineering - we analysed a realistic sample of four years of climate-related X posts across five annotation dimensions. 
Our findings paint a nuanced picture: VLMs are a powerful but imperfect tool, and their effective deployment requires careful attention to methodological choices that the current literature has largely overlooked.

\paragraph{VLMs as annotation tools: promising but not a drop-in replacement.}
Our results for RQ1 (\cref{subsec:RQ1}) demonstrate that VLMs achieve moderate to high macro accuracy across the five super-categories, with \textit{animals} performing best (0.76) and \textit{setting} proving most challenging (0.61). 
This performance is encouraging but falls short of the ``upload a codebook and go'' ideal. 
A critical insight, however, emerges from the distributional analysis: for super-categories such as \textit{animals} and \textit{type}, the Jensen-Shannon divergence between predicted and ground-truth distributions is negligible ($\leq .01$), even when individual image-level errors are frequent. 
This suggests that VLM predictions can reliably recover population-level discourse trends - the primary quantity of interest in most social science applications - even when per-image accuracy is sub-optimal. 
The practical implication is significant: researchers interested in tracking \textit{how} climate change is visually framed over time or across platforms may be able to use VLM annotations directly, provided that the errors are non-systematic. 
For super-categories with higher distributional divergence, such as \textit{consequences} and \textit{climate action}, more caution is warranted, and human validation of a representative subsample remains advisable.

The comparison with alternative automated methods further contextualises these results. 
CLIP-based zero-shot classifiers achieve substantially lower macro accuracies across all super-categories (up to 0.29 points lower), with a well-documented weakness in predicting negative classes. 
Among the models benchmarked, Gemini-3.1-flash-lite emerged as clearly the best-performing model across both datasets and all super-categories.

\paragraph{Taxonomy matters as much as model choice.}
The analysis for RQ2 (\cref{subsec:RQ2}) reveals that taxonomy design is a first-order determinant of classification quality - arguably as influential as the choice of model itself. 
We observe consistent confusions between semantically proximate categories (e.g., \textit{floods} and \textit{sea level rise}, \textit{biodiversity loss} and \textit{drought}), and the broad ``other'' and ``none'' categories appear disproportionately in the top confusion pairs across all super-categories. 
These confusions are often asymmetric: for instance, \textit{pets} are misclassified as \textit{farm animals} but not vice versa, and \textit{polar bear} is confused with \textit{land mammal} but not the reverse. 
This pattern suggests that models tend to overpredict general or more frequently encountered categories at the expense of specific ones - a bias that social science researchers should anticipate when designing codebooks for automated annotation.

The comparison between \textit{ClimateCT} and \textit{ClimateTV} confusion matrices is instructive: the expert-annotated \textit{ClimateCT} exhibits fewer and less severe confusions than the larger, automatically labelled \textit{ClimateTV} - even though we only look at the human confirmed labels. 
This is partly attributable to dataset size, but the differing confusion patterns also indicate that the annotation pipeline itself introduces systematic biases.
For practitioners, this finding underscores the importance of grounding automated annotations in a well-validated expert-coded subset rather than relying solely on pipeline-generated labels.

\paragraph{Model selection: size is not everything.}
The benchmark results for RQ3 (\cref{subsec:RQ3}) challenge the assumption that larger models necessarily produce better classifications. 
Qwen3 with 8 billion parameters performs on par with or better than GPT-5.4 mini across multiple super-categories, while the larger 30-billion-parameter Qwen3 variant shows a performance \textit{drop} alongside the highest prediction variance of all evaluated models. 
This finding has practical implications for cost-sensitive research settings: smaller open-weight models may offer a favourable accuracy-to-cost ratio, particularly when inference must be run at scale over hundreds of thousands of images. 
The consistent performance gap between \textit{ClimateCT} and \textit{ClimateTV} (0.03-0.08 points across models) further confirms that popular, widely-shared images are systematically easier to classify than the long tail of heterogeneous content - a selection bias that researchers should account for when extrapolating from curated evaluation sets to full-corpus deployment.

\paragraph{Prompt engineering: targeted effort pays off.}
The prompt ablation of RQ4 (\cref{subsec:RQ4}) yields actionable and dimension-specific recommendations. 
Short prompts with only category names suffice for \textit{type}, where the categories are visually self-explanatory. 
In contrast, all other super-categories benefit from extended explanations and examples, most dramatically \textit{consequences}, where longer prompts yield a consistent accuracy boost of approximately $+0.3$. 
For \textit{climate action} - the most abstract super-category - adding super-category-level definitions further improves accuracy while reducing variance. 
Counterintuitively, explicitly triggering chain-of-thought reasoning (``think carefully step by step'') \textit{reduces} performance for most super-categories, suggesting that for visual classification tasks with well-defined categories, additional reasoning steps may introduce noise rather than clarity. 
Spatial localisation instructions similarly provide no consistent benefit. 
The low prompt variance across paraphrases on both \textit{ClimateCT} and \textit{ClimateTV} suggests that results are robust to minor prompt reformulations, which is reassuring for reproducibility.

\paragraph{Limitations.}
Several limitations should be acknowledged. 
First, our analysis is restricted to a single platform (X/Twitter) and the generalisability of our findings to other platforms (e.g., Instagram, TikTok) remains to be established. 
Second, our annotation scheme, while grounded in established social science codebooks~\citep{bravo2025viral}, is not exhaustive - dimensions such as affect, stance, or narrative framing are not captured. 
Third, the evaluation of VLMs is inherently a snapshot: model capabilities evolve rapidly, and the specific performance figures reported here will likely be superseded by future model releases. 
However, the \textit{methodological} findings - regarding taxonomy design, distributional vs.\ sample-wise evaluation, and prompt engineering - are likely to remain relevant across model generations. 
Fourth, the DINOv2-based diversity analysis, while informative, is constrained by the expressiveness of the embedding space and may not capture all dimensions of visual variability relevant to social science categories.
Finally, the use of QualityMatch workers for large-scale label validation, while cost-effective, introduces a different source of annotation noise compared to expert coding, as for all such tasks.
This is reflected in the lower acceptance rates for \textit{setting} and \textit{type}.

\paragraph{Implications for social science research.}
Our findings have several implications for the growing community of researchers seeking to integrate computer vision methods into social science workflows. 
First, we recommend that researchers adopt a two-tiered evaluation strategy: sample-wise metrics (accuracy, F1) for understanding per-category reliability, and distributional metrics (Jensen-Shannon divergence, total variation) for assessing whether aggregate trends can be trusted. 
Second, codebook design should be treated as a modelling decision: categories at inconsistent levels of abstraction will systematically degrade classifier performance, and the inclusion of broad ``other'' categories, while analytically convenient, creates predictable confusion hotspots. 
Third, prompt engineering effort should be allocated selectively - simple categories need simple prompts, while abstract or domain-specific categories benefit from detailed explanations, but not from chain-of-thought or localisation instructions. 
Fourth, open-weight models of moderate size represent a viable and cost-effective alternative to closed-weight APIs for large-scale annotation tasks.

\paragraph{Conclusion.}
Visual discourse analysis at scale is no longer a distant prospect - it is technically feasible today. 
However, the transition from manual to automated annotation is not a simple substitution but a methodological shift that requires careful consideration of taxonomy design, model selection, prompt formulation, and validation strategy. 
By providing a systematic evaluation across these dimensions, grounded in a large-scale real-world dataset of climate change imagery, this study offers concrete, actionable guidelines for researchers and practitioners.
We release our image IDs, labels, and code to support transparency and reproducibility, and we hope that this work will help bridge the gap between the computer vision and social science communities in their shared effort to understand how climate change is visually communicated in public digital spaces.


\acks{This work is partially funded by the BMFTR project 16DKWN027b Climate Visions, DFG research unit 5336 Learning2Sense and by the German Federal Ministry for Research, Technology, and Space (BMFTR).
I am grateful to Teja, Max, Shashank, and Simon for always offering the right words at the right time.
and Gemini for editing the figures and tables.
All experiments were run on the computational resources of the
University of Mannheim and the Max Planck Institute for Informatics.
We further acknowledge QualityMatch for supporting the data annotation.}


\newpage

\appendix

\section{Dataset Description}
\label{app:dataset}
To give context to our data sample, we share sample images for several categories, describe category diversity, and label combinations across super-categories.

\subsection{Example Images from \textit{ClimateCT} and \textit{ClimateTV}}
\label{app:climatetv_ex}
\begin{figure}[!ht]
\centering
\captionsetup{type=figure}
    \includegraphics[height=2cm, width=3cm]{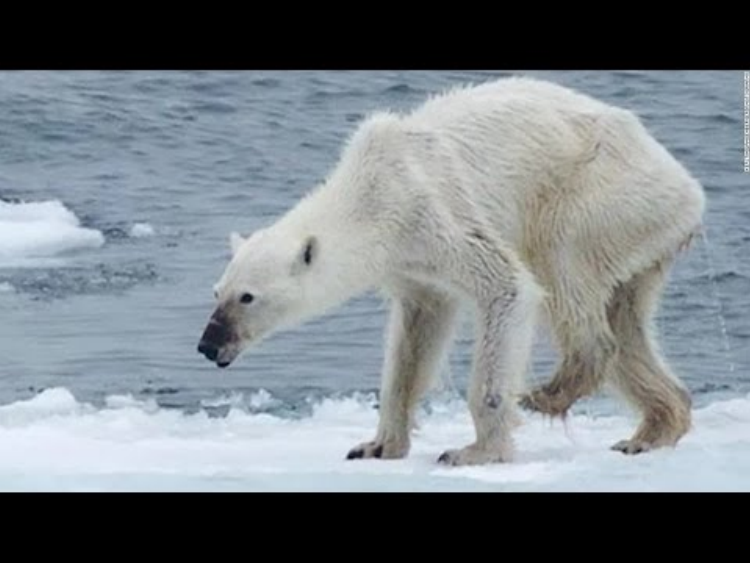}\hfill
    \includegraphics[height=2cm, width=3cm]{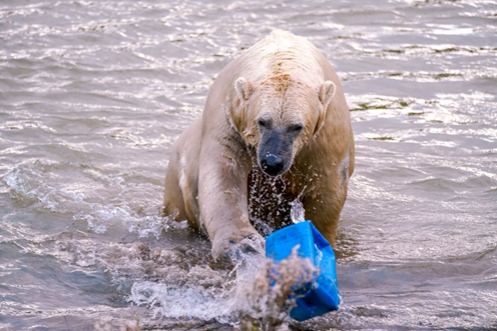}\hfill
    \includegraphics[height=2cm, width=3cm]{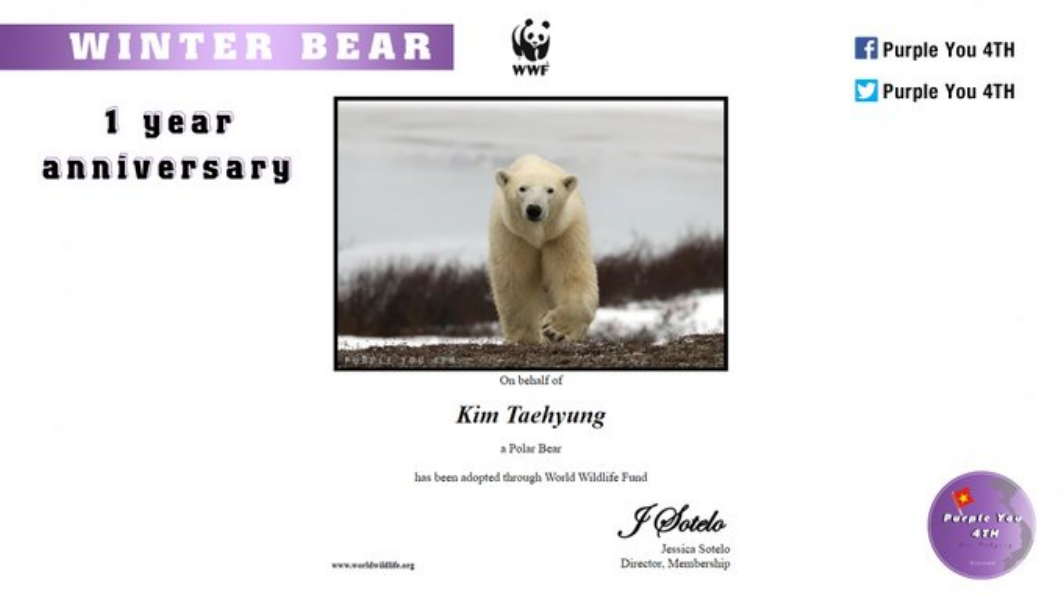}\hfill
    \includegraphics[height=2cm, width=3cm]{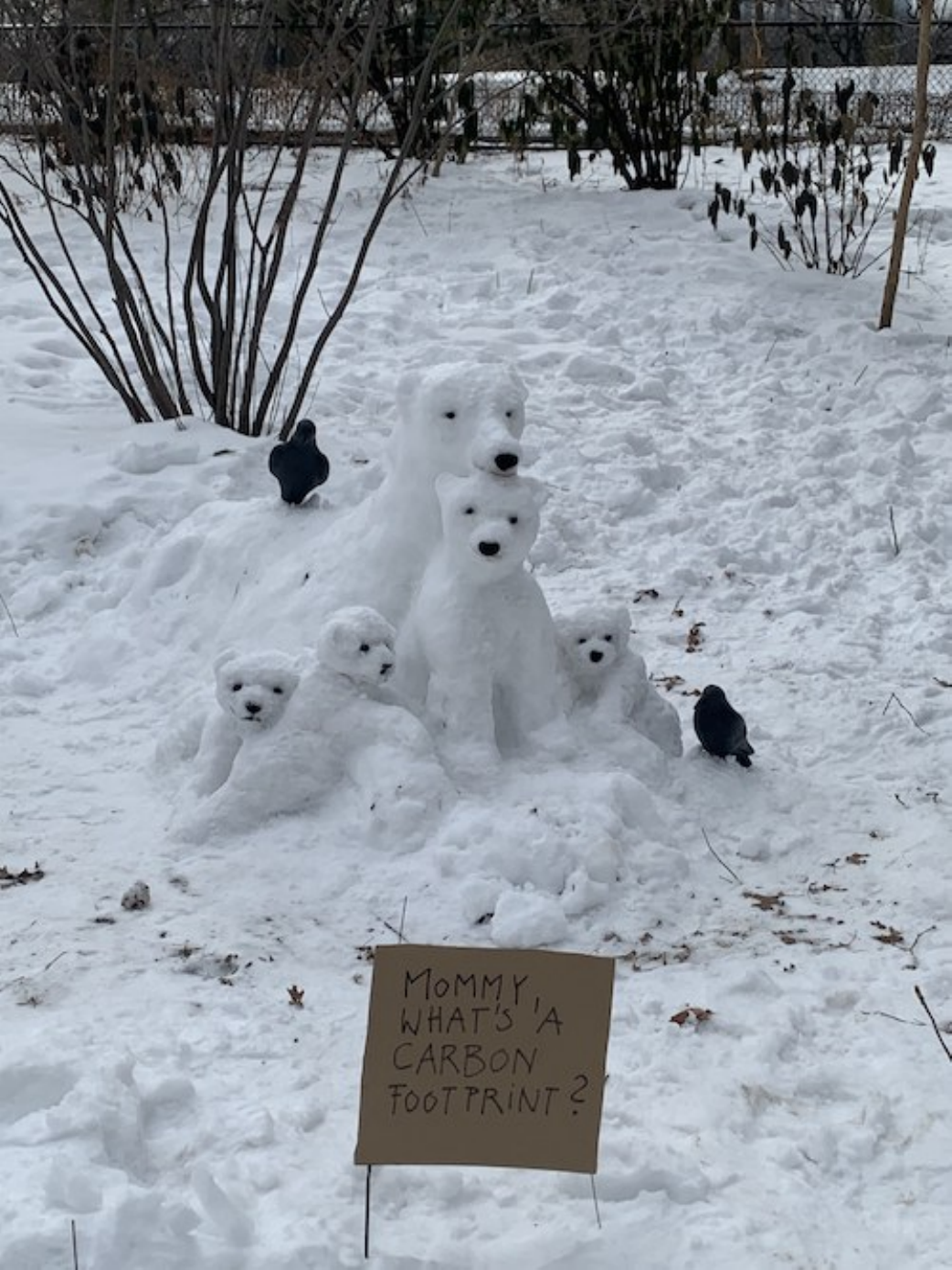}
    \\[\smallskipamount]
    \includegraphics[height=2cm, width=3cm]{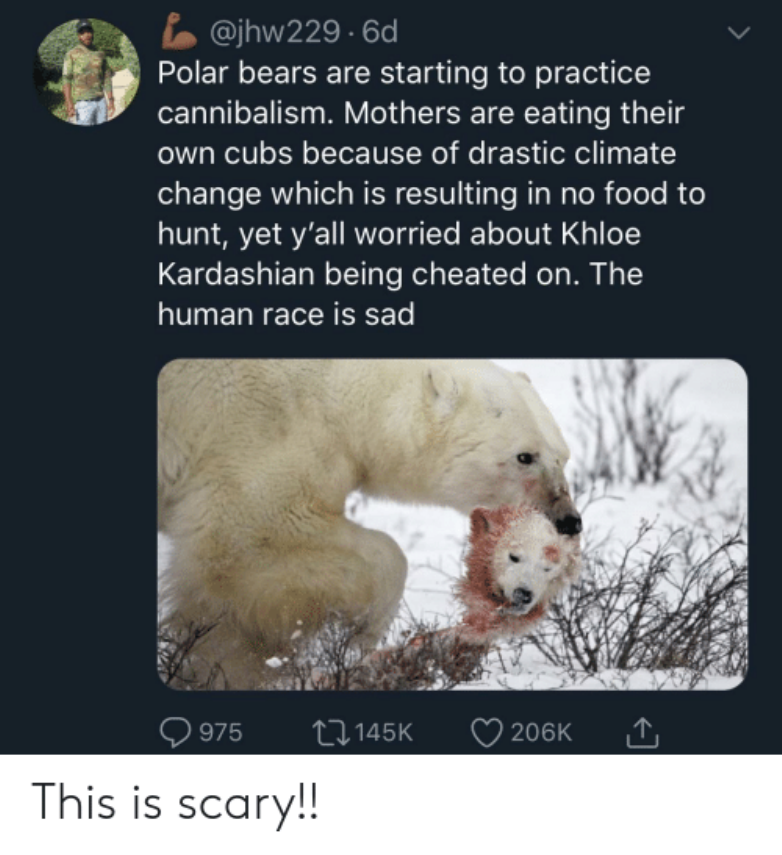}\hfill
    \includegraphics[height=2cm, width=3cm]{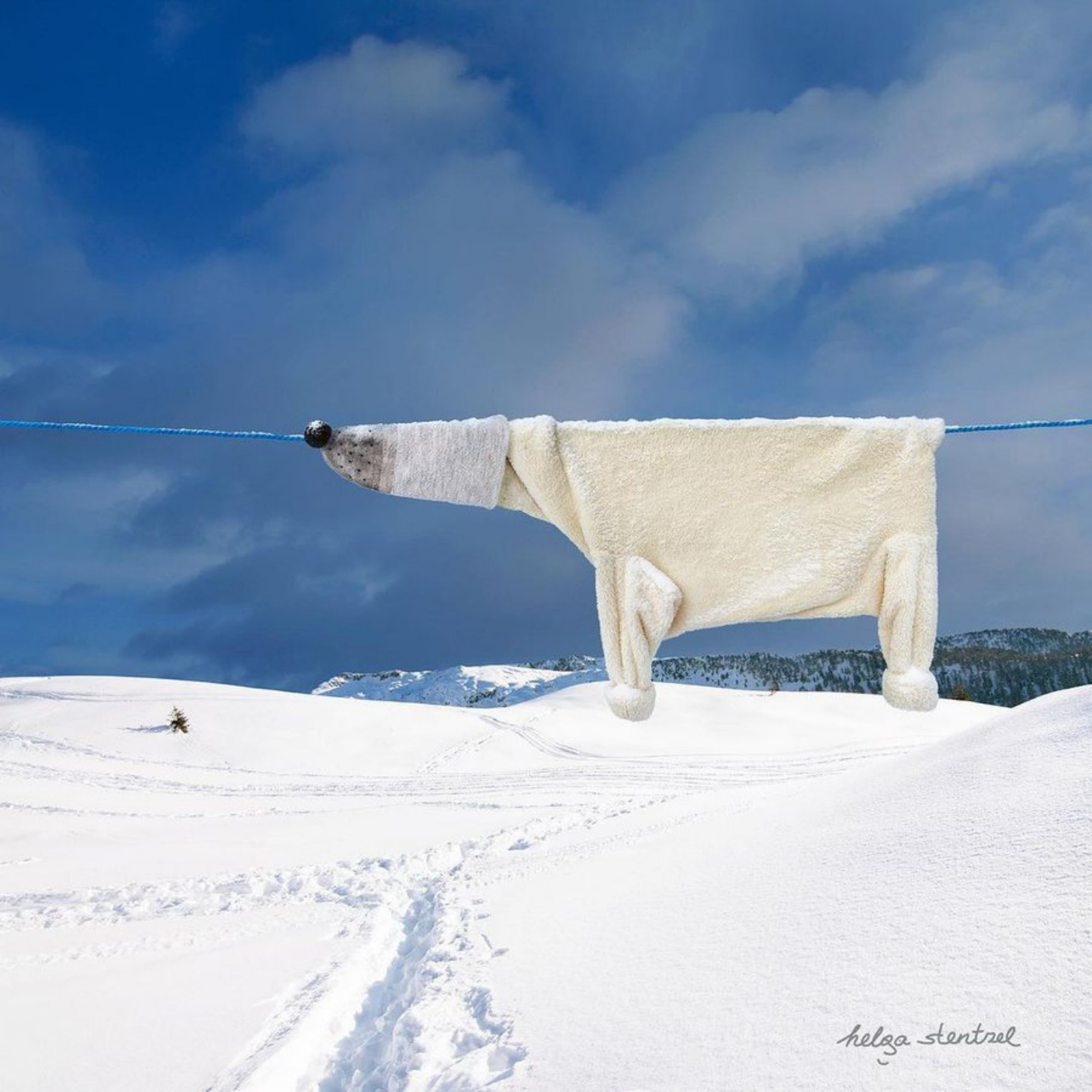}\hfill
    \includegraphics[height=2cm, width=3cm]{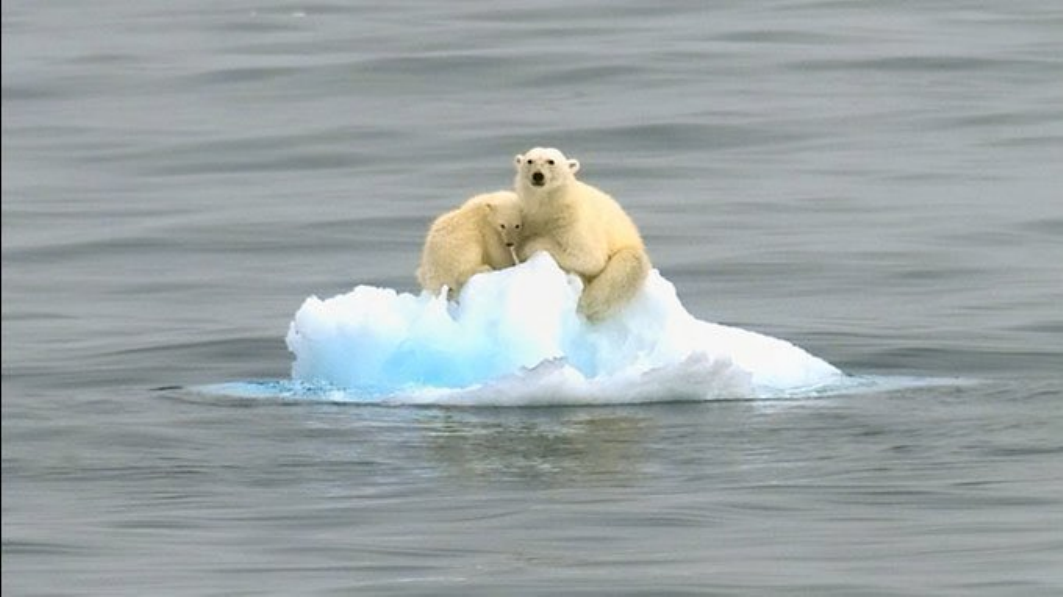}\hfill
    \includegraphics[height=2cm, width=3cm]{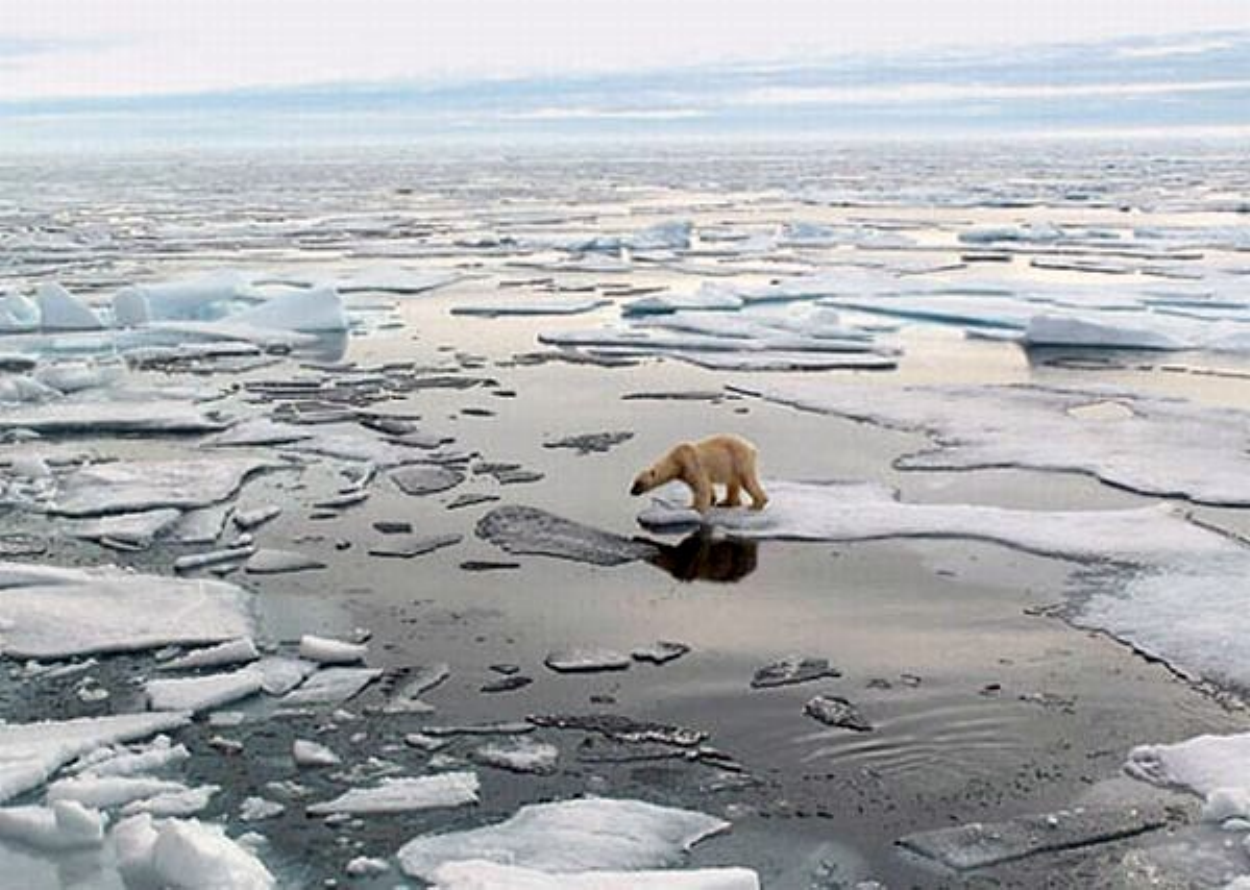}

\captionof{figure}{\textit{ClimateCT} image examples for the super-class \textit{animals}' class \textit{polar bear}.}
\label{fig:an_pol_ct}
\end{figure}
In the following, we provide images from the \textit{ClimateCT} and \textit{ClimateTV} data sets to provide intuition for the nature of the analysed images.
While all are collected using the same set of keywords, the larger \textit{ClimateTV} data set naturally contains a more diverse set of images compared to the smaller \textit{ClimateCT}, which is sampled from the most popular tweets and extended with manually selected, prevalent examples of underrepresented categories. 

\begin{figure}[!ht]
\centering
\captionsetup{type=figure}
    \includegraphics[height=2cm, width=3cm]{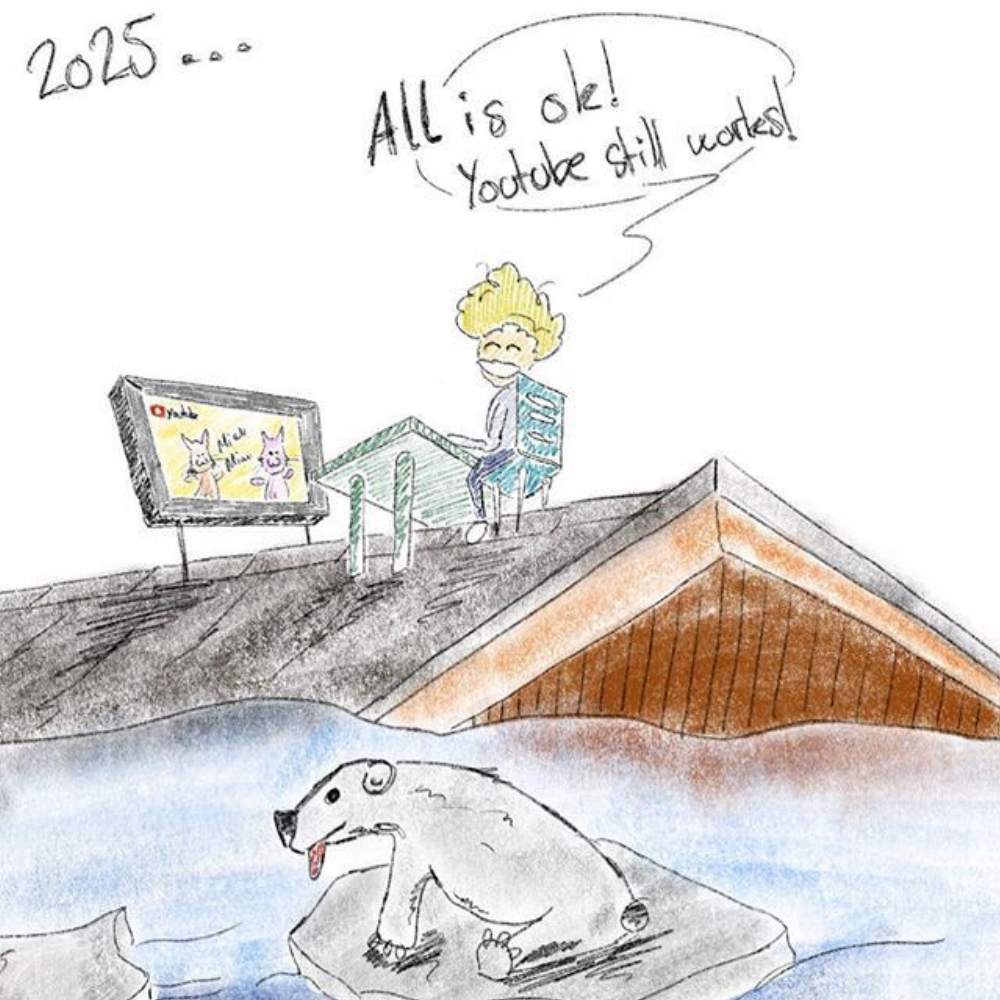}\hfill
    \includegraphics[height=2cm, width=3cm]{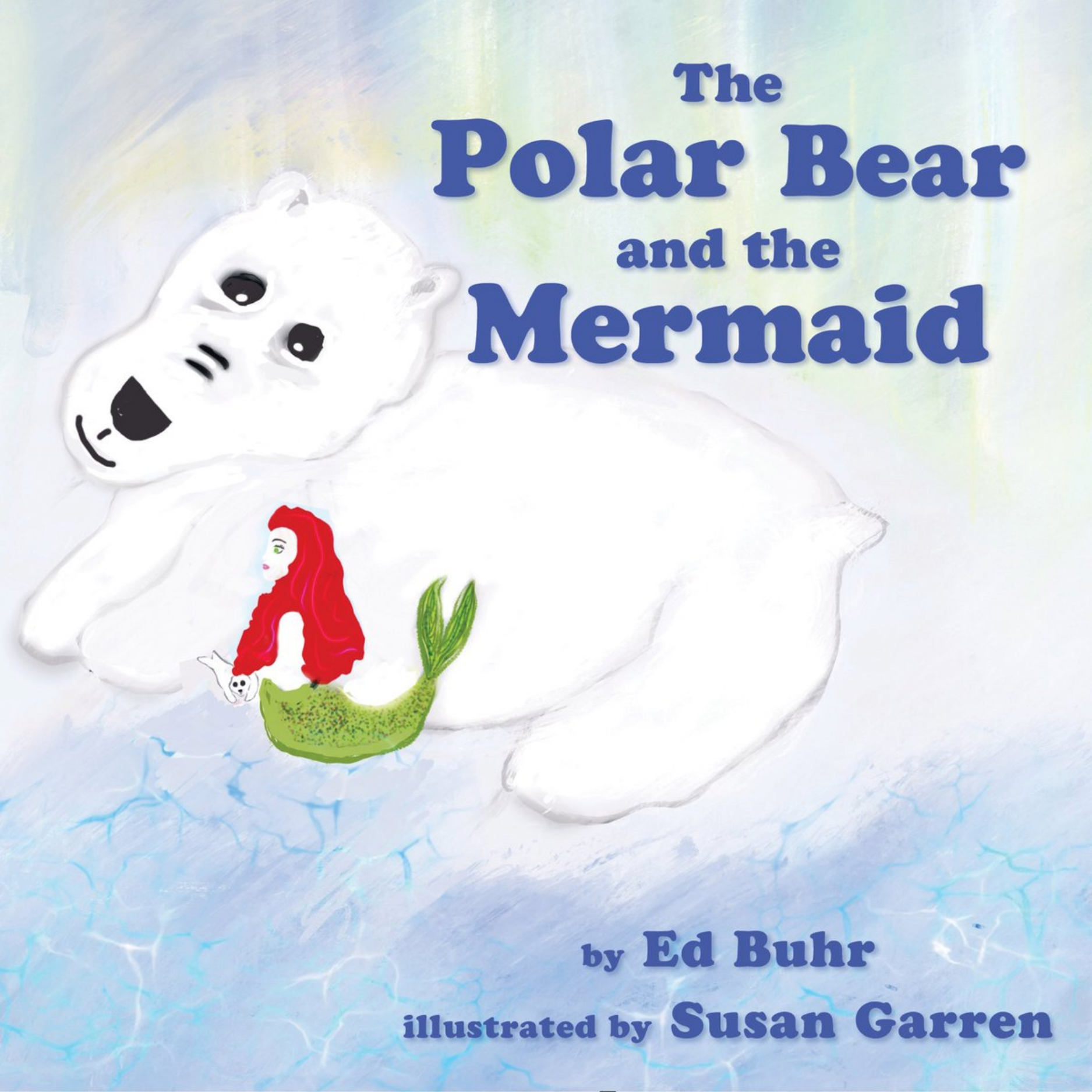}\hfill
    \includegraphics[height=2cm, width=3cm]{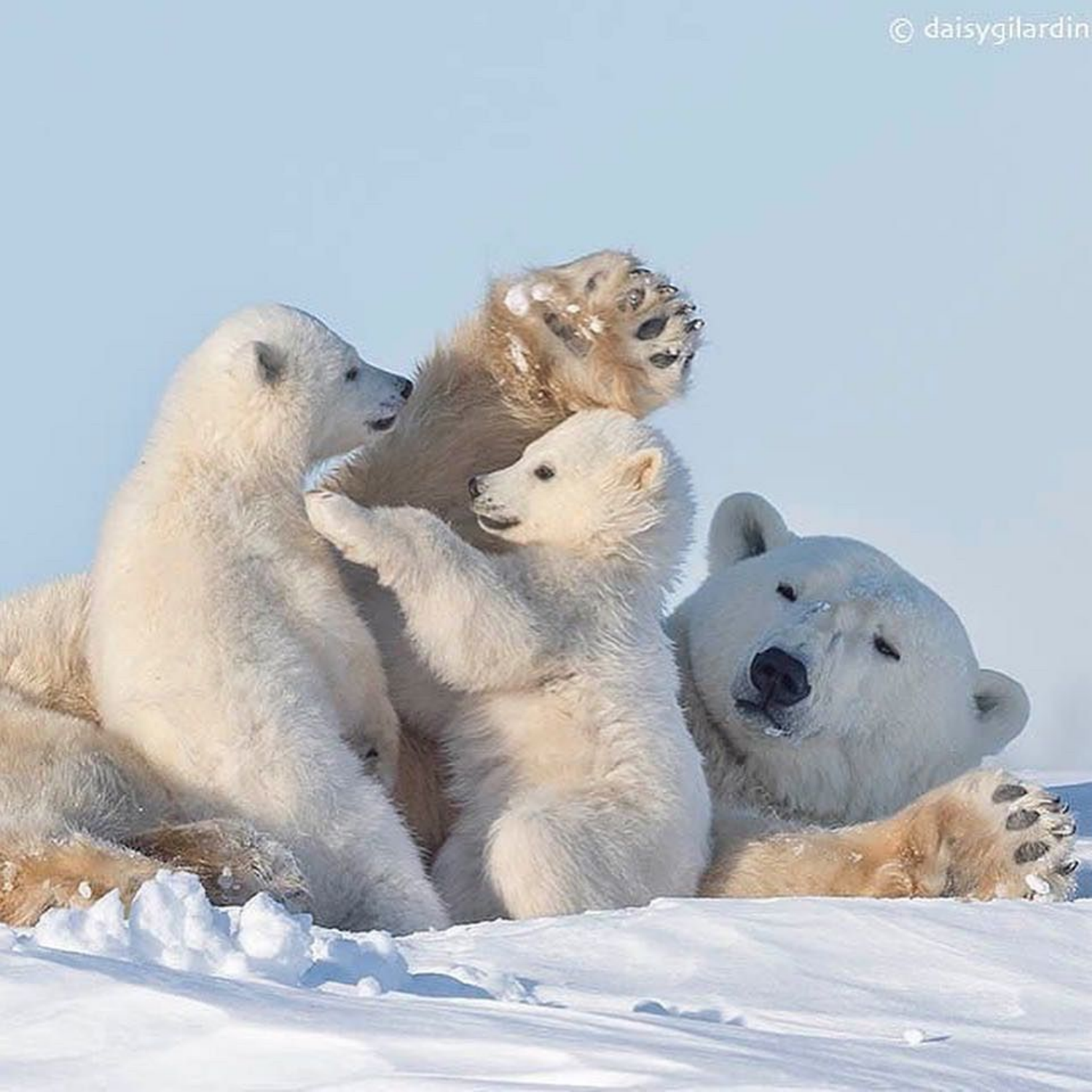}\hfill
    \includegraphics[height=2cm, width=3cm]{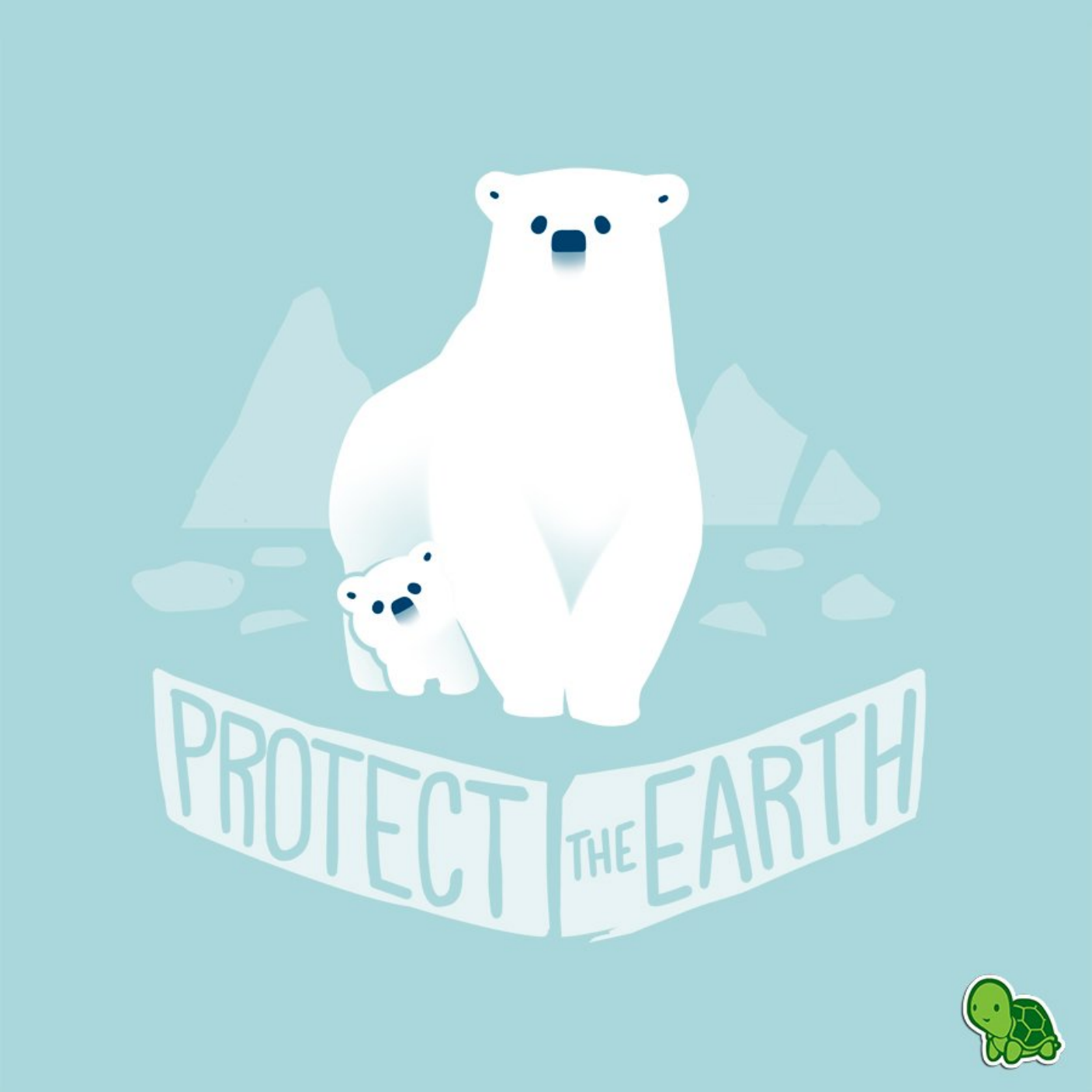}
    \\[\smallskipamount]
    \includegraphics[height=2cm, width=3cm]{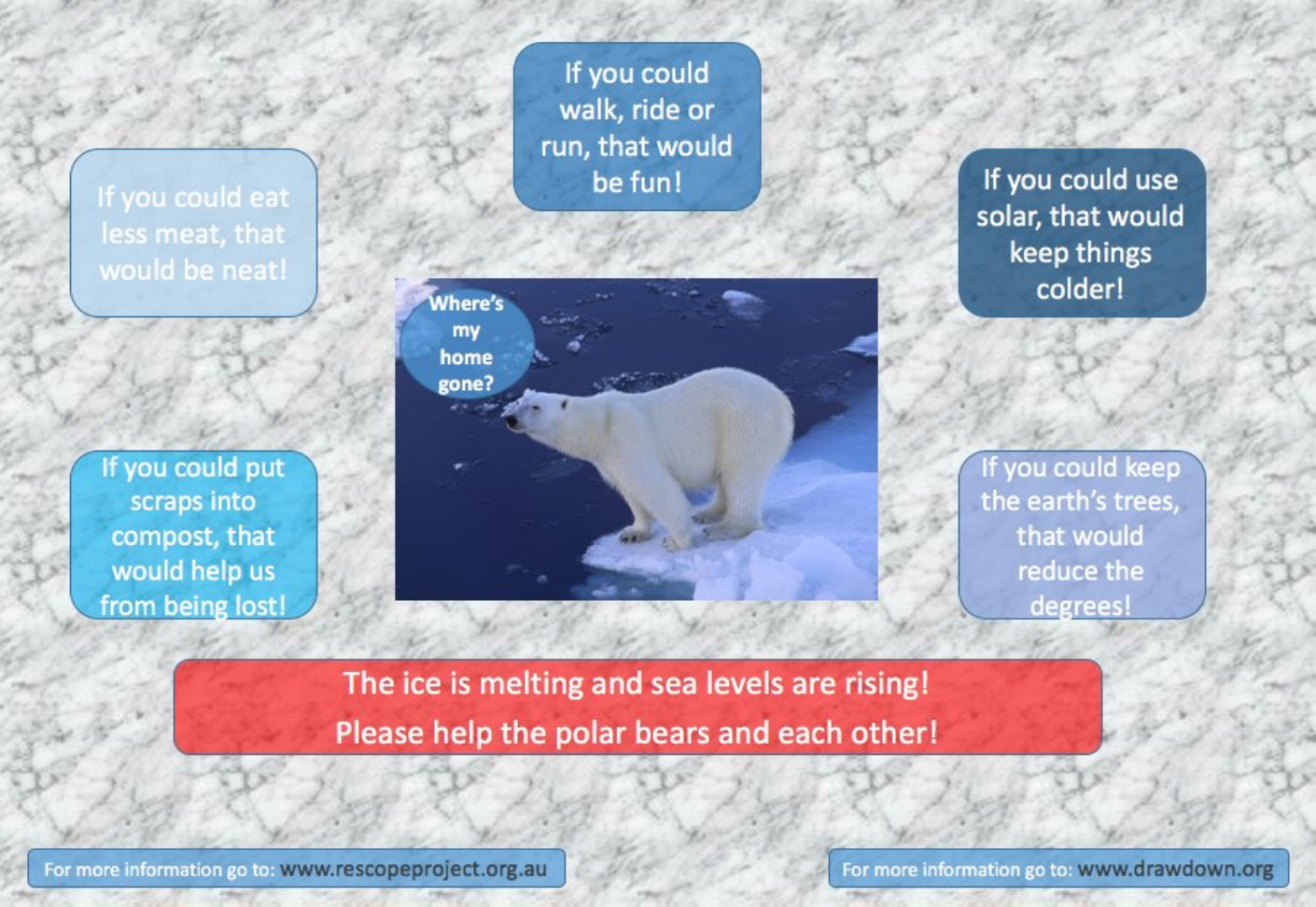}\hfill
    \includegraphics[height=2cm, width=3cm]{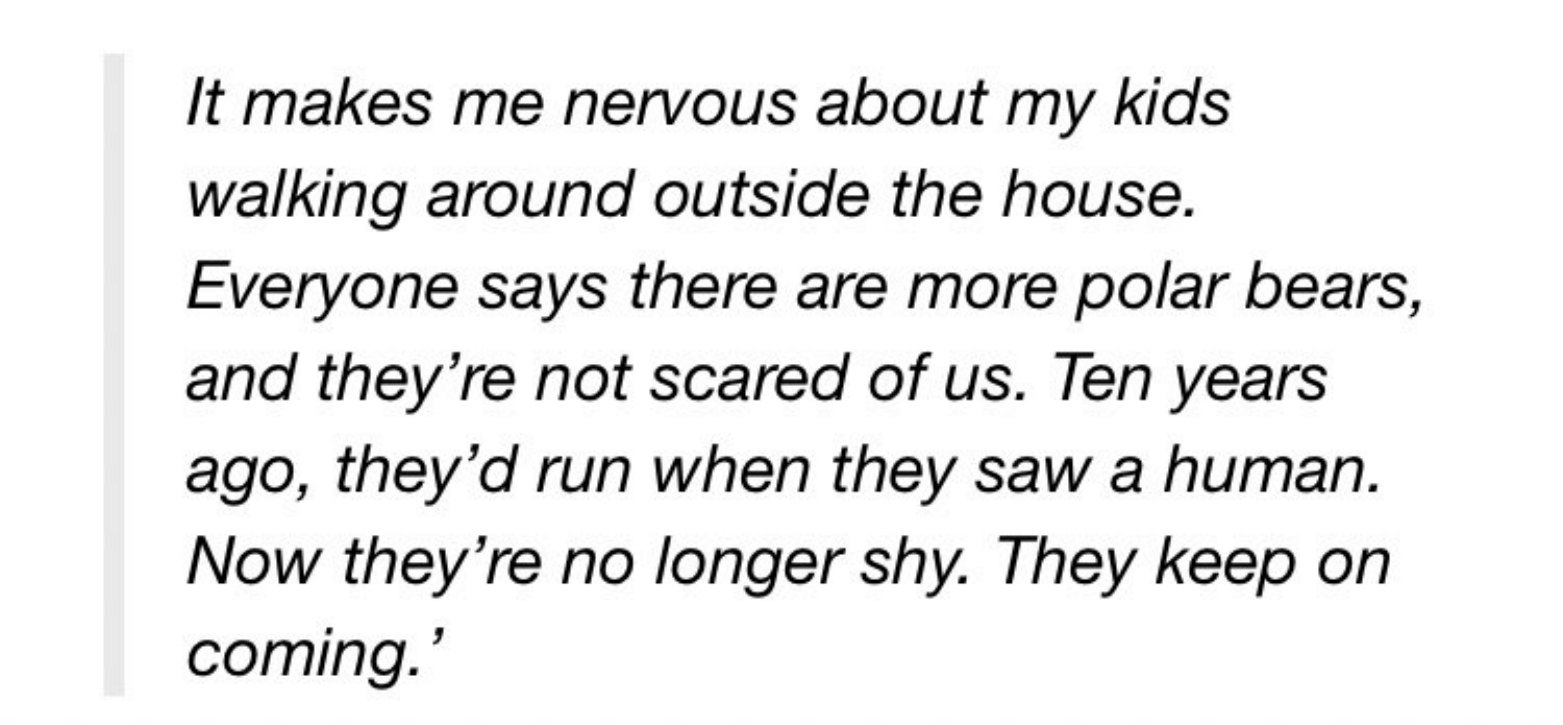}\hfill
    \includegraphics[height=2cm, width=3cm]{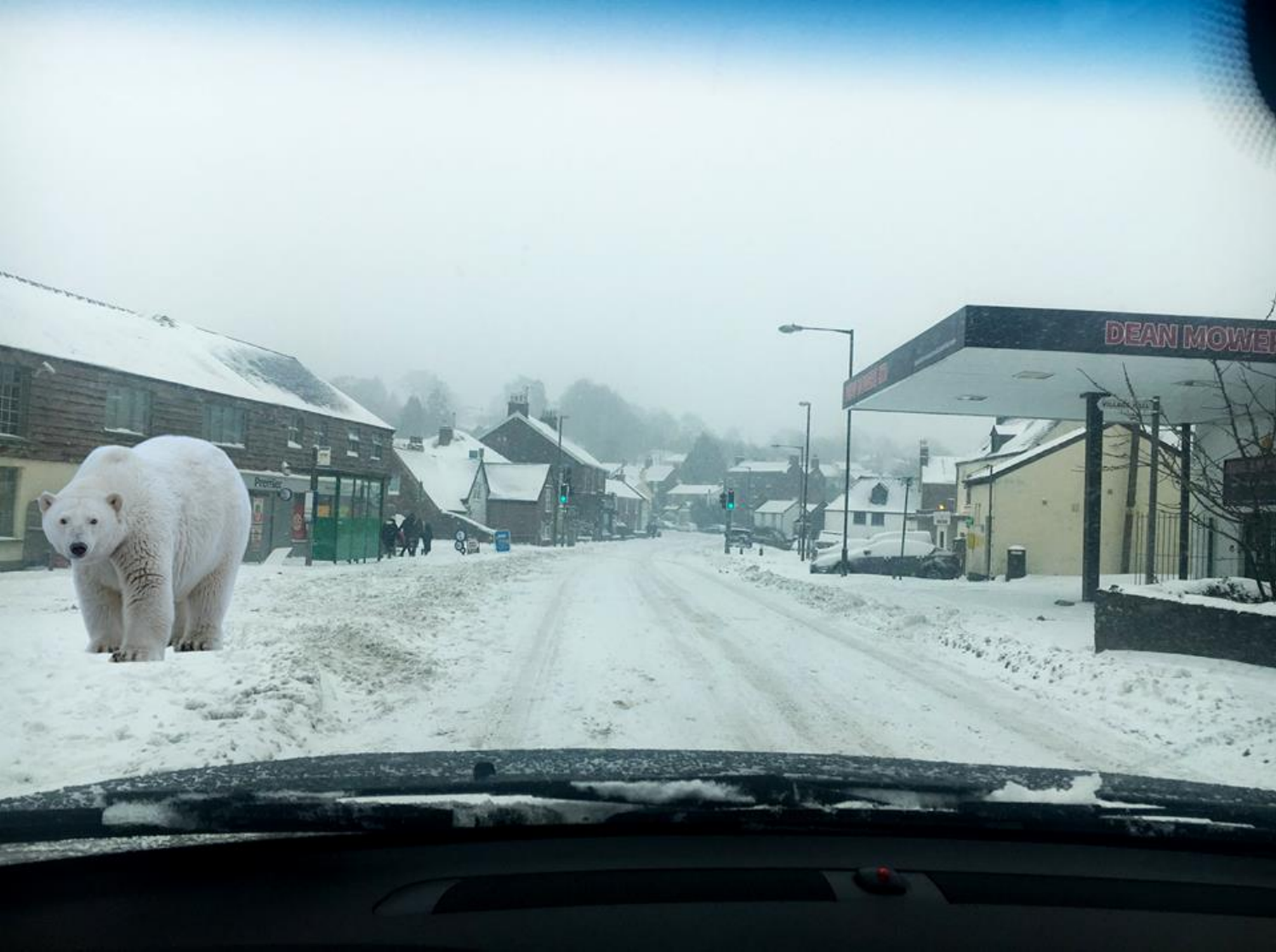}\hfill
    \includegraphics[height=2cm, width=3cm]{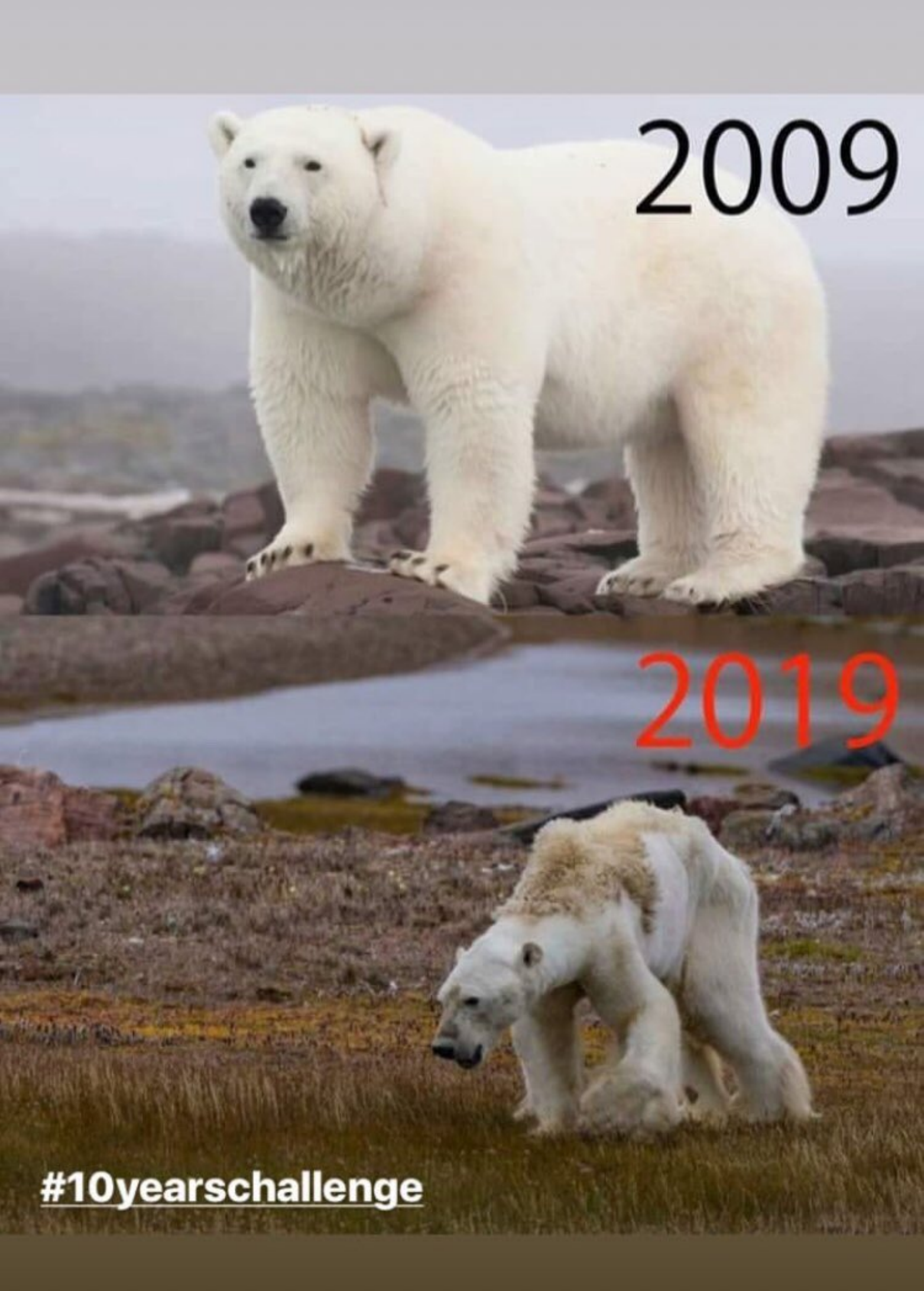}

\captionof{figure}{\textit{ClimateTV} image examples for the super-class \textit{animals}' class \textit{polar bear}.}
\label{fig:an_pol_TV}
\end{figure}

Our most specific class, \textit{polar bear} visualizes this well:
\autoref{fig:an_pol_ct} and \autoref{fig:an_pol_TV} show images of climate change's most famous animal.
Both data sets include real polar bears in various health states besides more abstract representations of this class.
\textit{ClimateCT} contains, among others,  a polar bear costume and a polar bear formed out of snow (\autoref{fig:an_pol_ct}, while \textit{ClimateTV} contains~e.g. illustrations and textual references in a screenshot (\autoref{fig:an_pol_TV}). 
These samples shows that multi-dimensional category design is beneficial: For example, polar bear images may be of various \textit{types}, within different \textit{settings}, containing several climate change \textit{consequences}, and possibly include climate change-related actions (\textit{climate action}).

\begin{figure*}[!ht]
\centering
    \includegraphics[height=2cm, width=3cm]{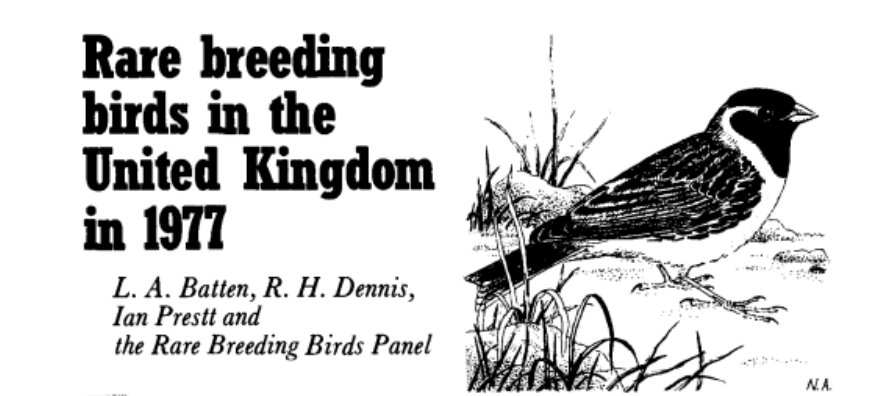}\hfill
    \includegraphics[height=2cm, width=3cm]{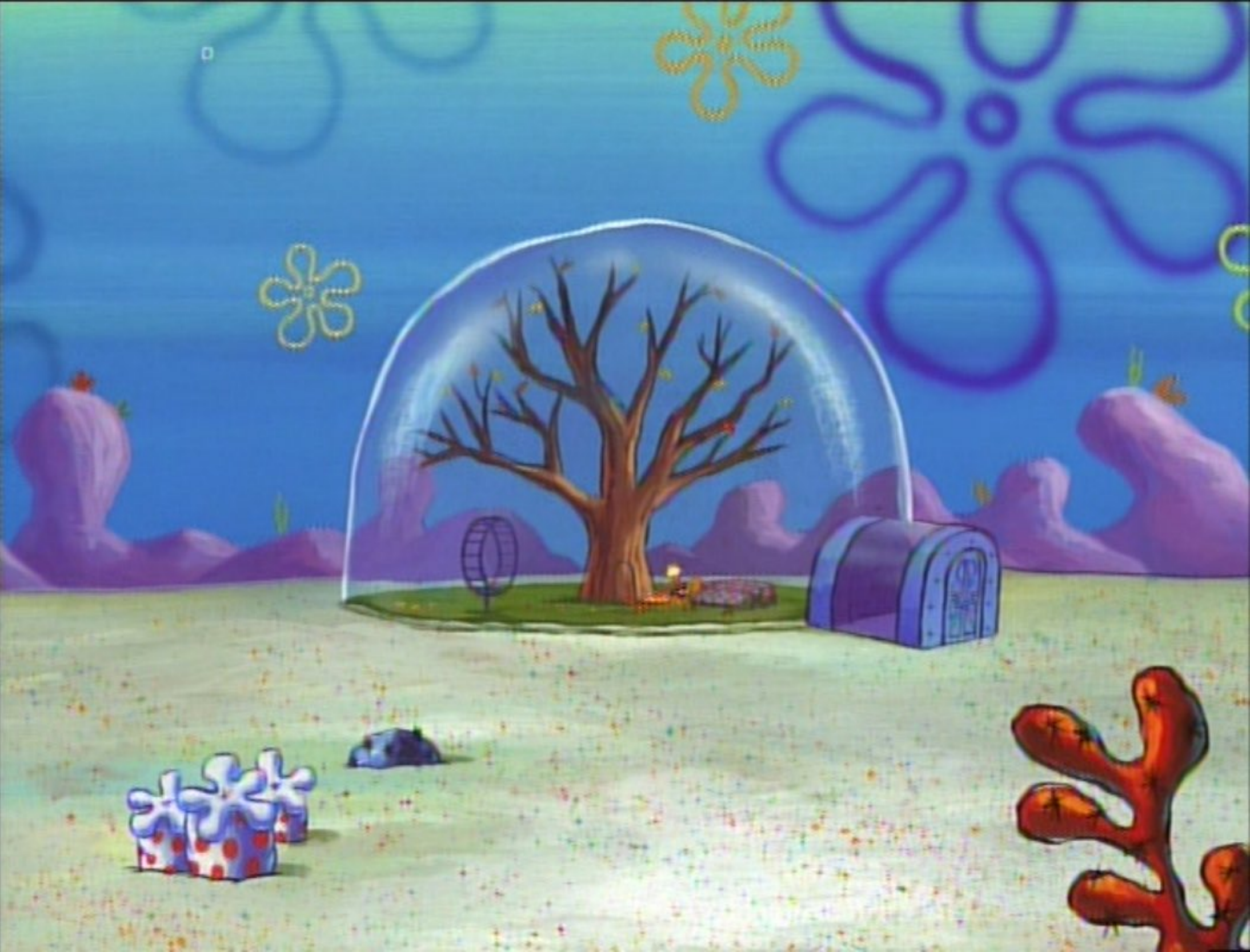}\hfill
    \includegraphics[height=2cm, width=3cm]{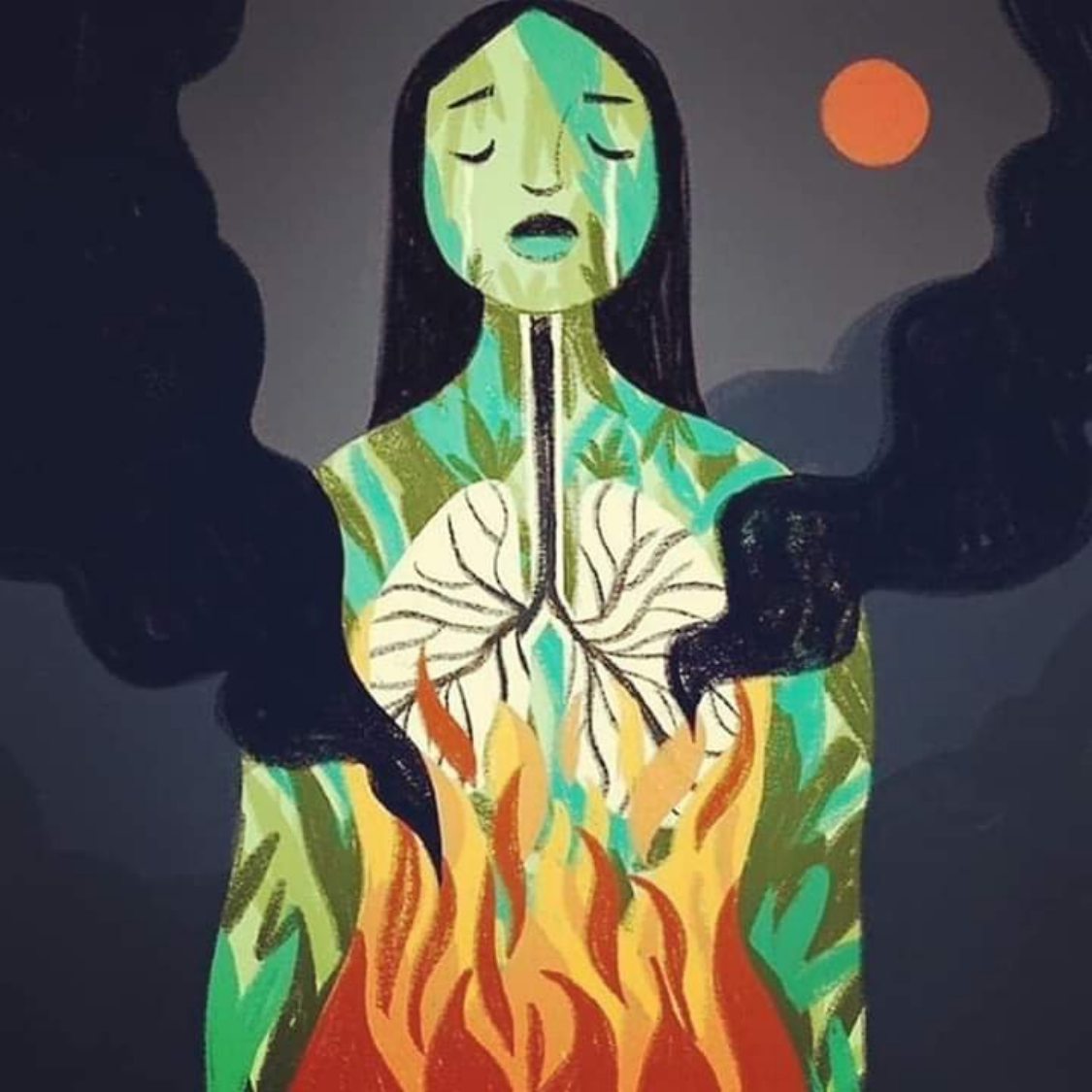}\hfill
    \includegraphics[height=2cm, width=3cm]
    {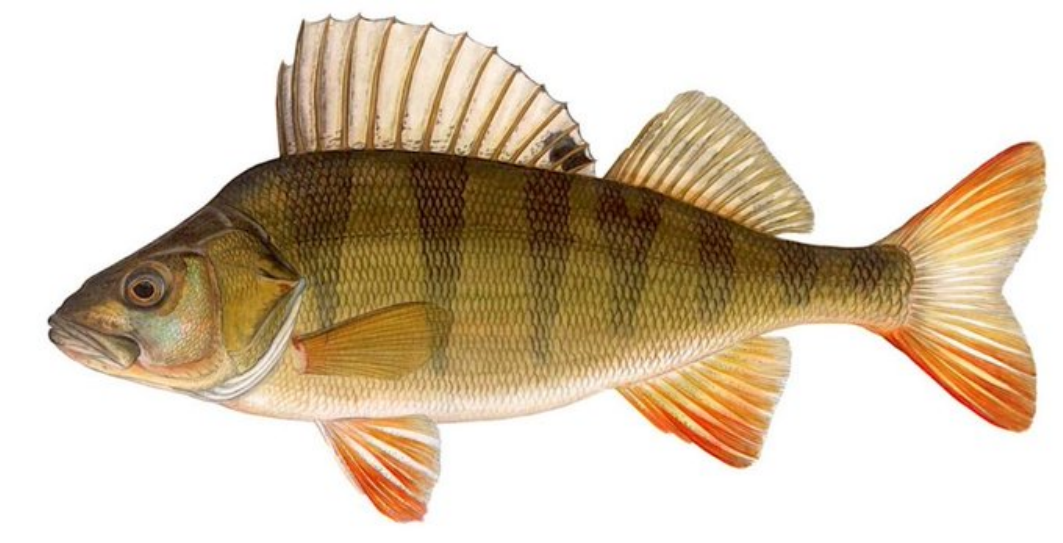}    
    \\[\smallskipamount]
    \includegraphics[height=2cm, width=3cm]{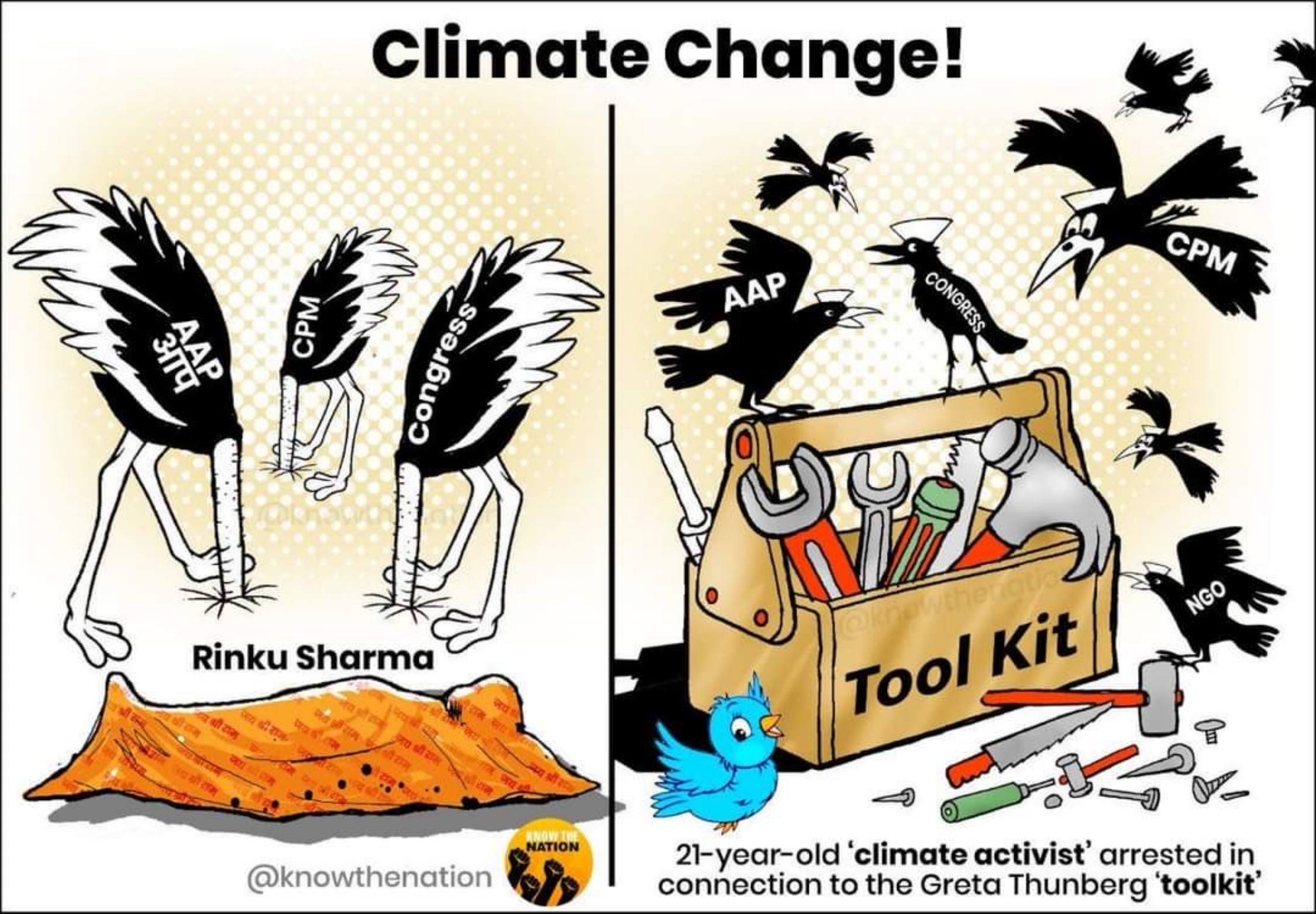}\hfill
    \includegraphics[height=2cm, width=3cm]{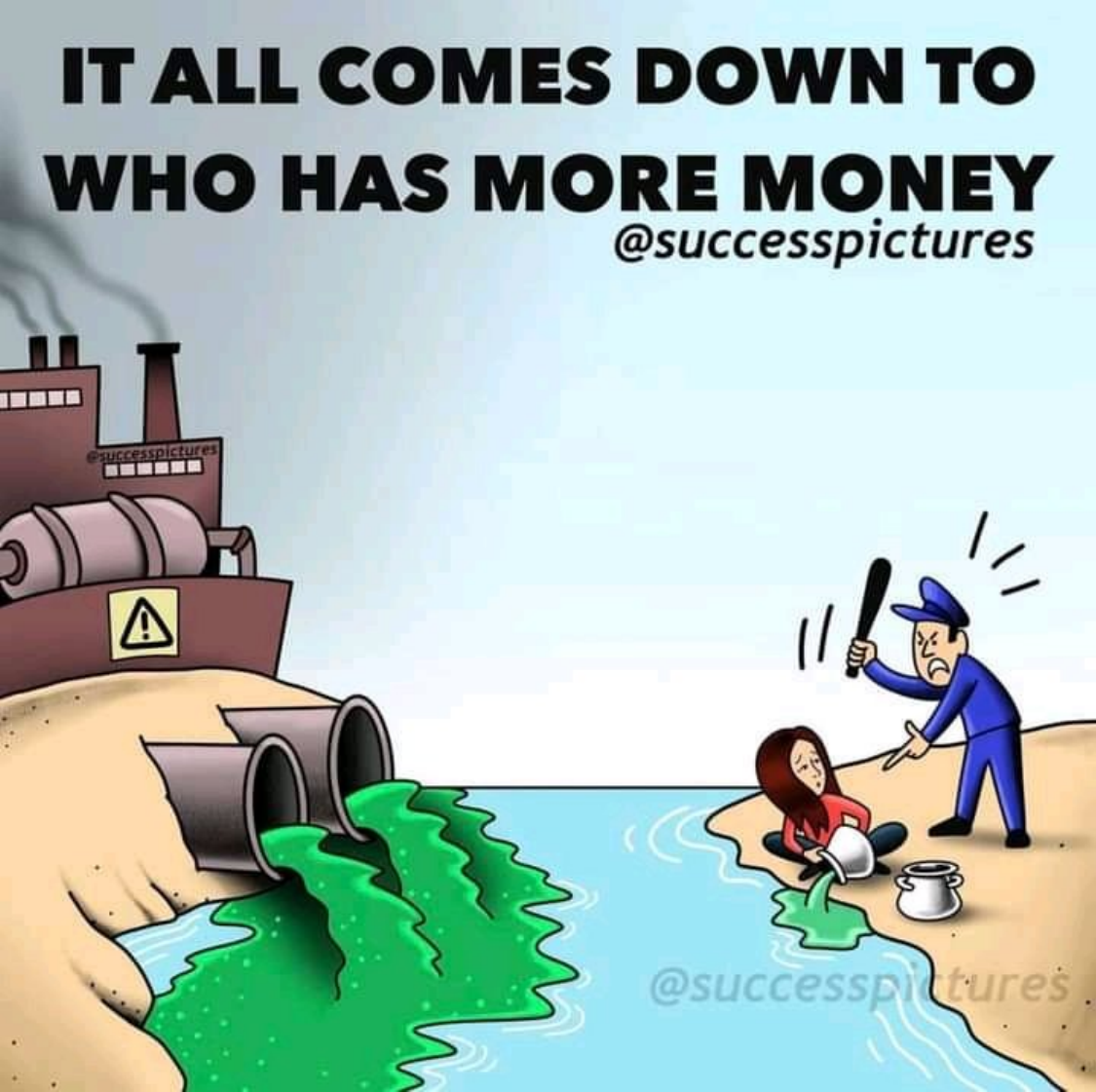}\hfill
    \includegraphics[height=2cm, width=3cm]{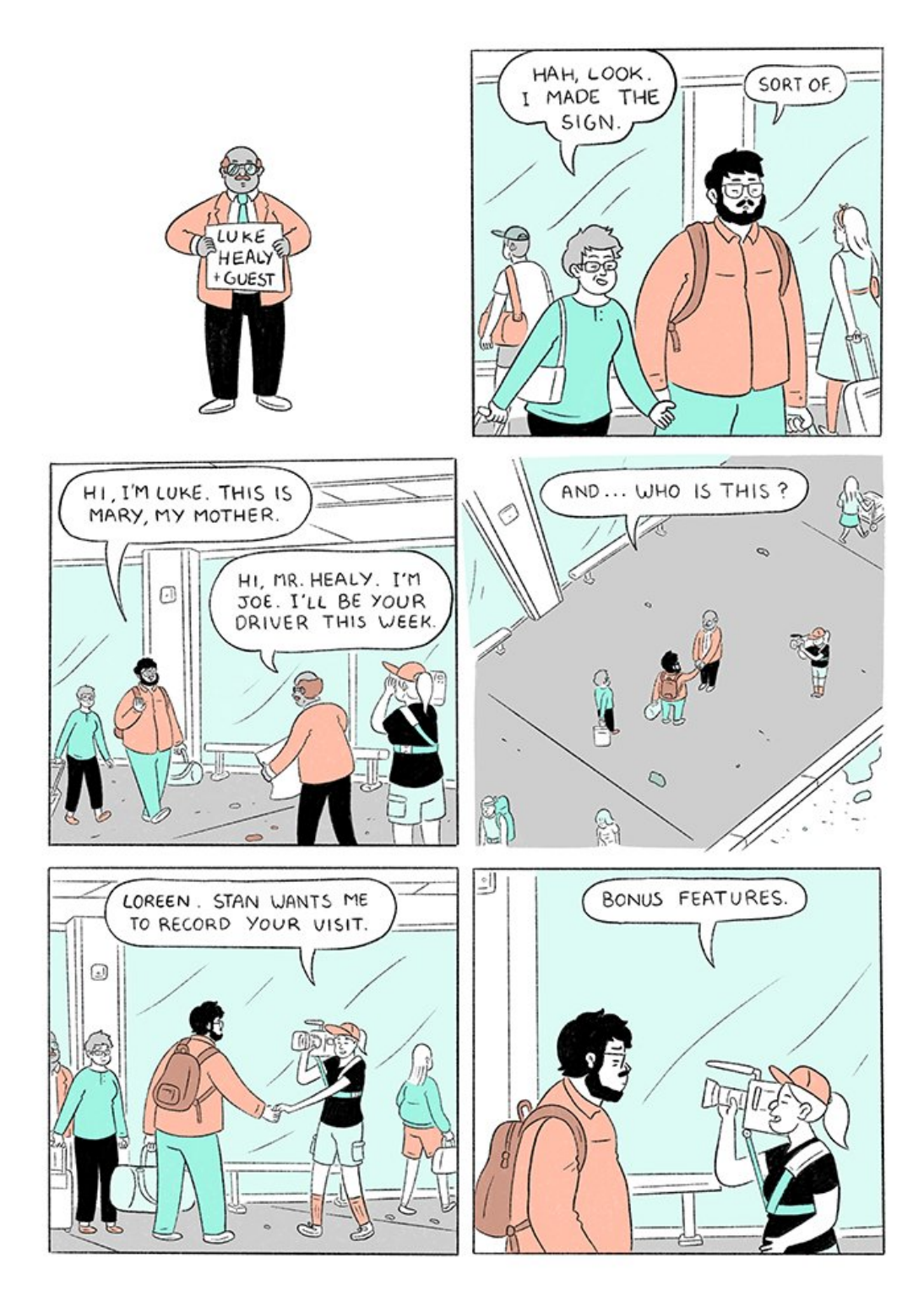}
    \includegraphics[height=2cm, width=3cm]{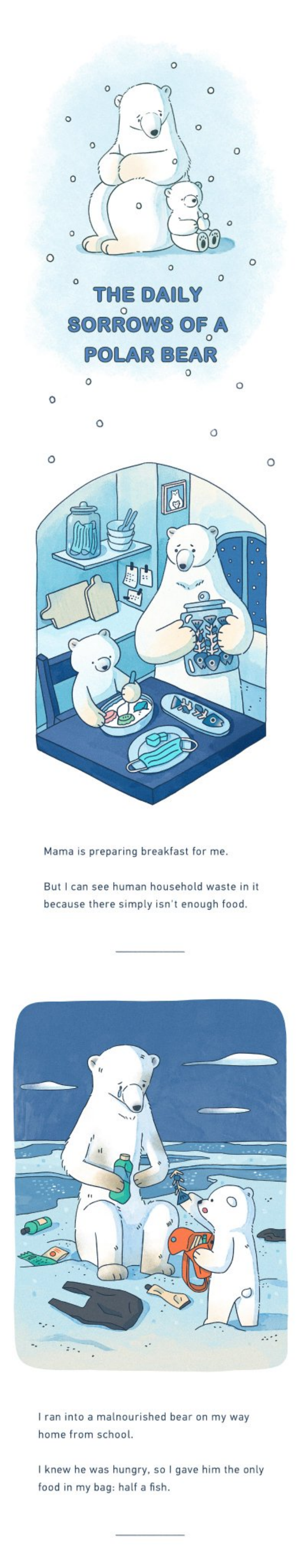}\hfill
\caption{\textit{ClimateCT} image examples for the super-category \textit{type}'s class \textit{illustration}. }
\label{fig:type_ill_ct}
\vspace{1em}
\end{figure*}
\begin{figure*}[!ht]
\centering
    \includegraphics[height=2cm, width=3cm]{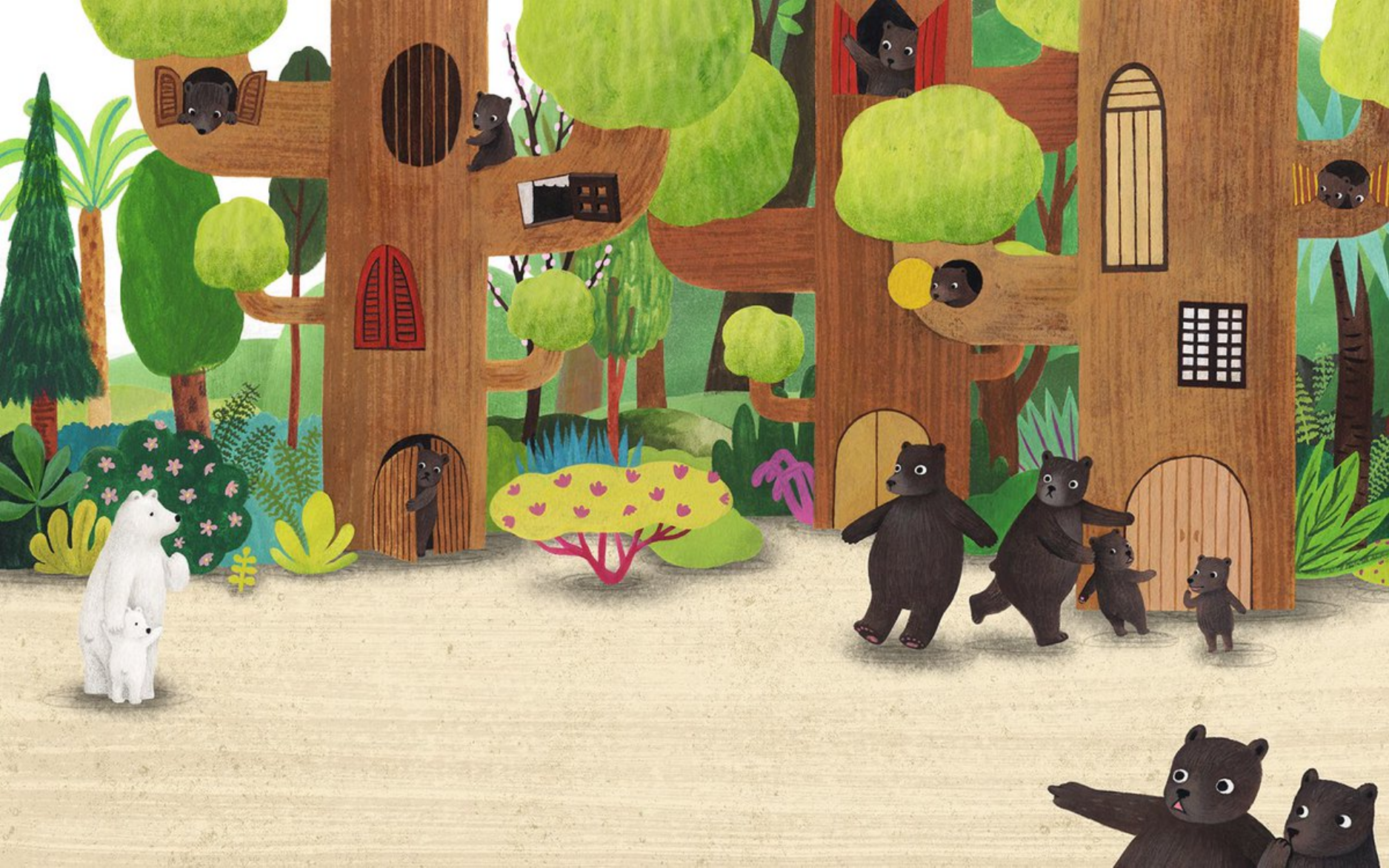}\hfill
    \includegraphics[height=2cm, width=3cm]{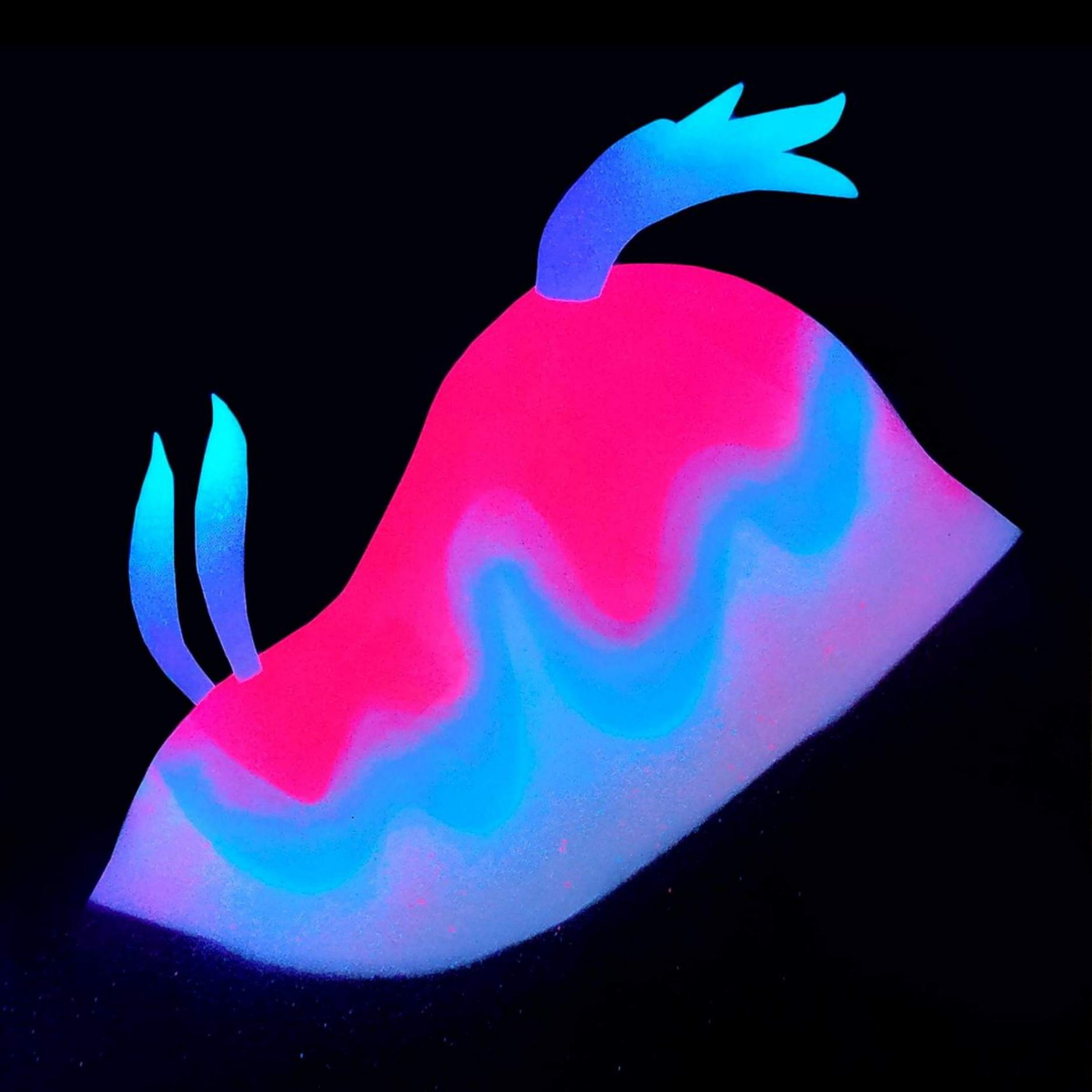}\hfill
    \includegraphics[height=2cm, width=3cm]{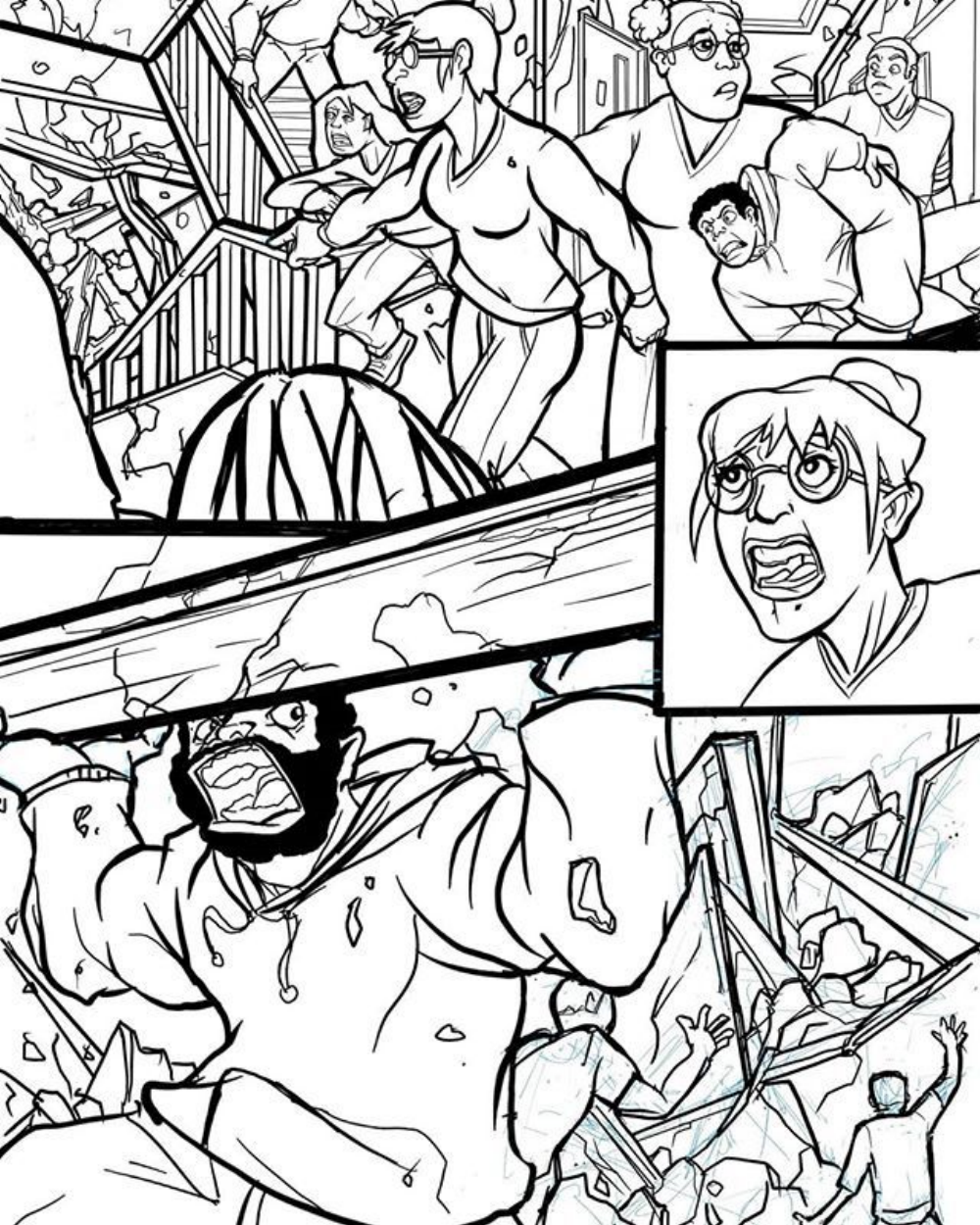}\hfill
    \includegraphics[height=2cm, width=3cm]{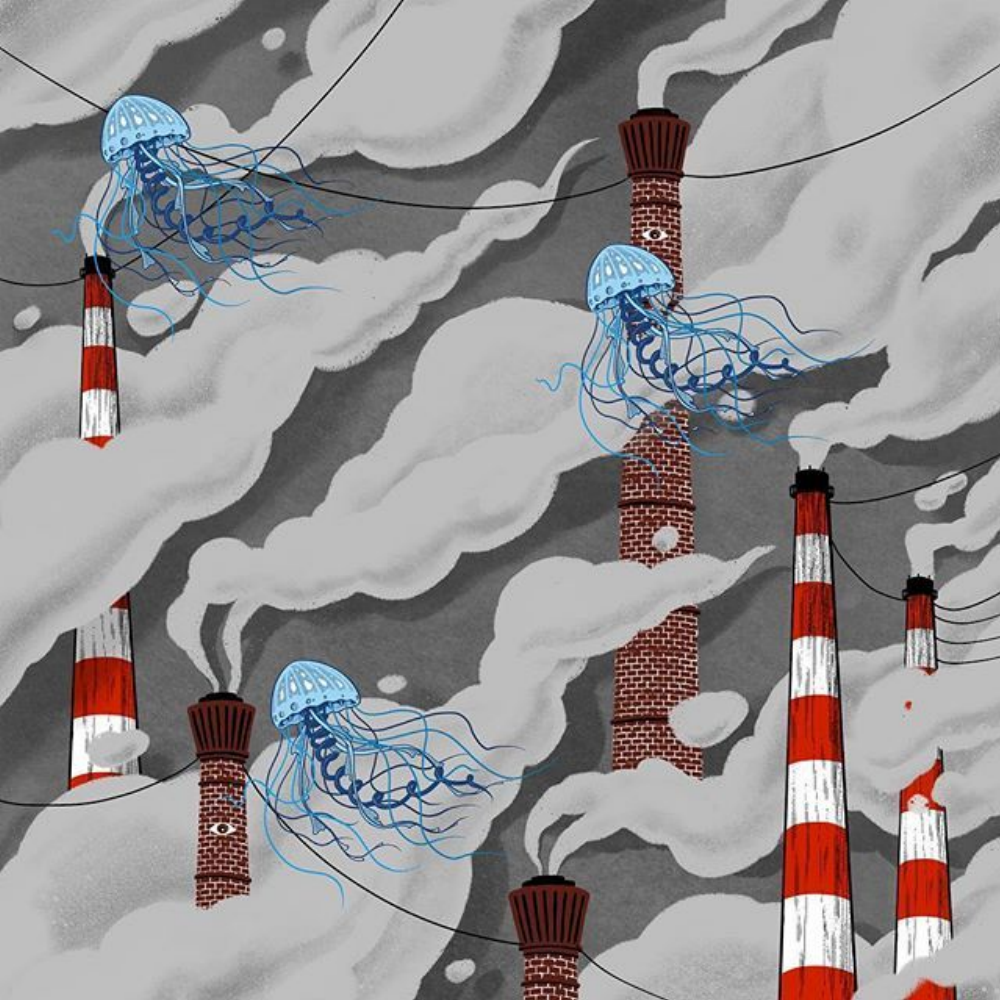}
    \\[\smallskipamount]
    \includegraphics[height=2cm, width=3cm]{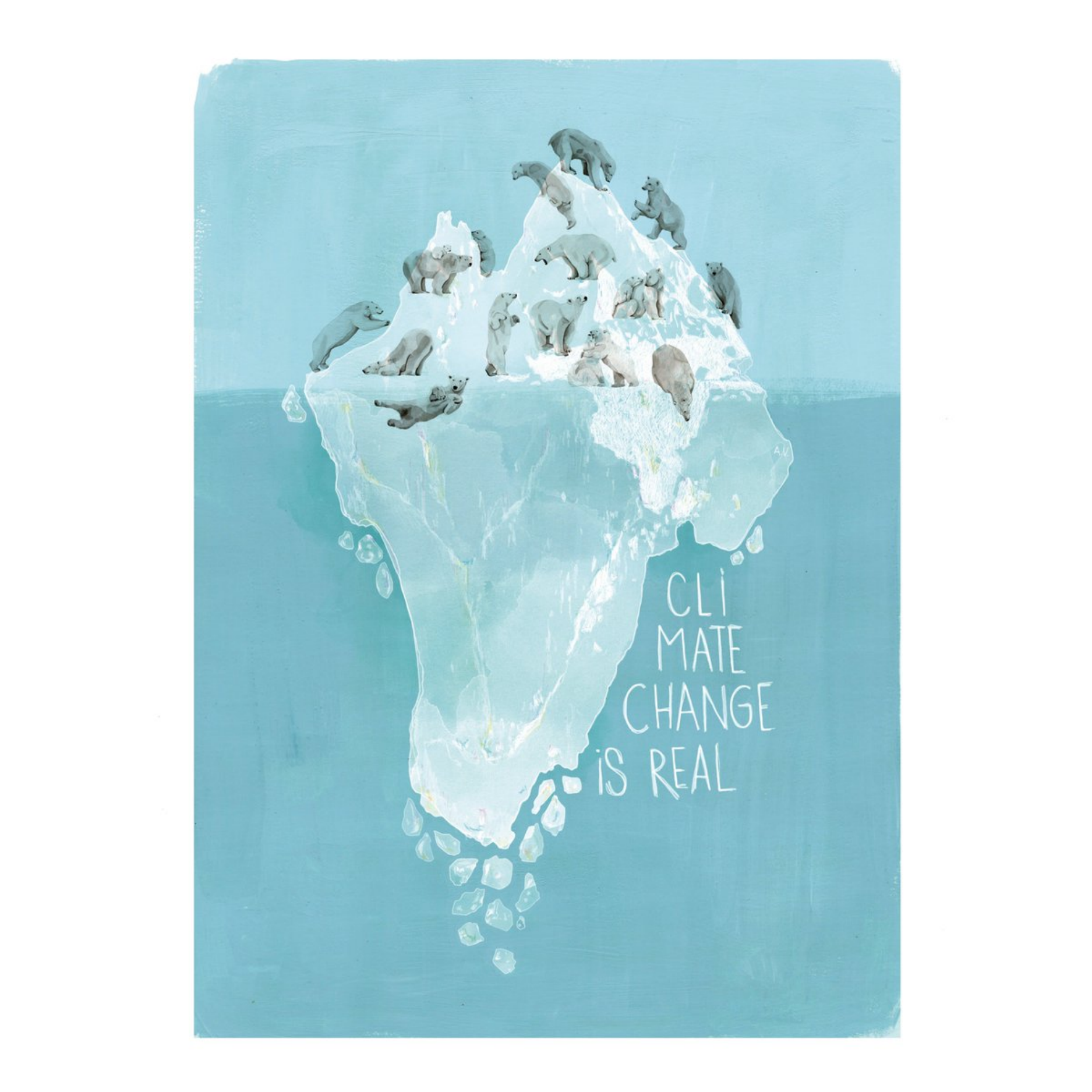}\hfill
    \includegraphics[height=2cm, width=3cm]{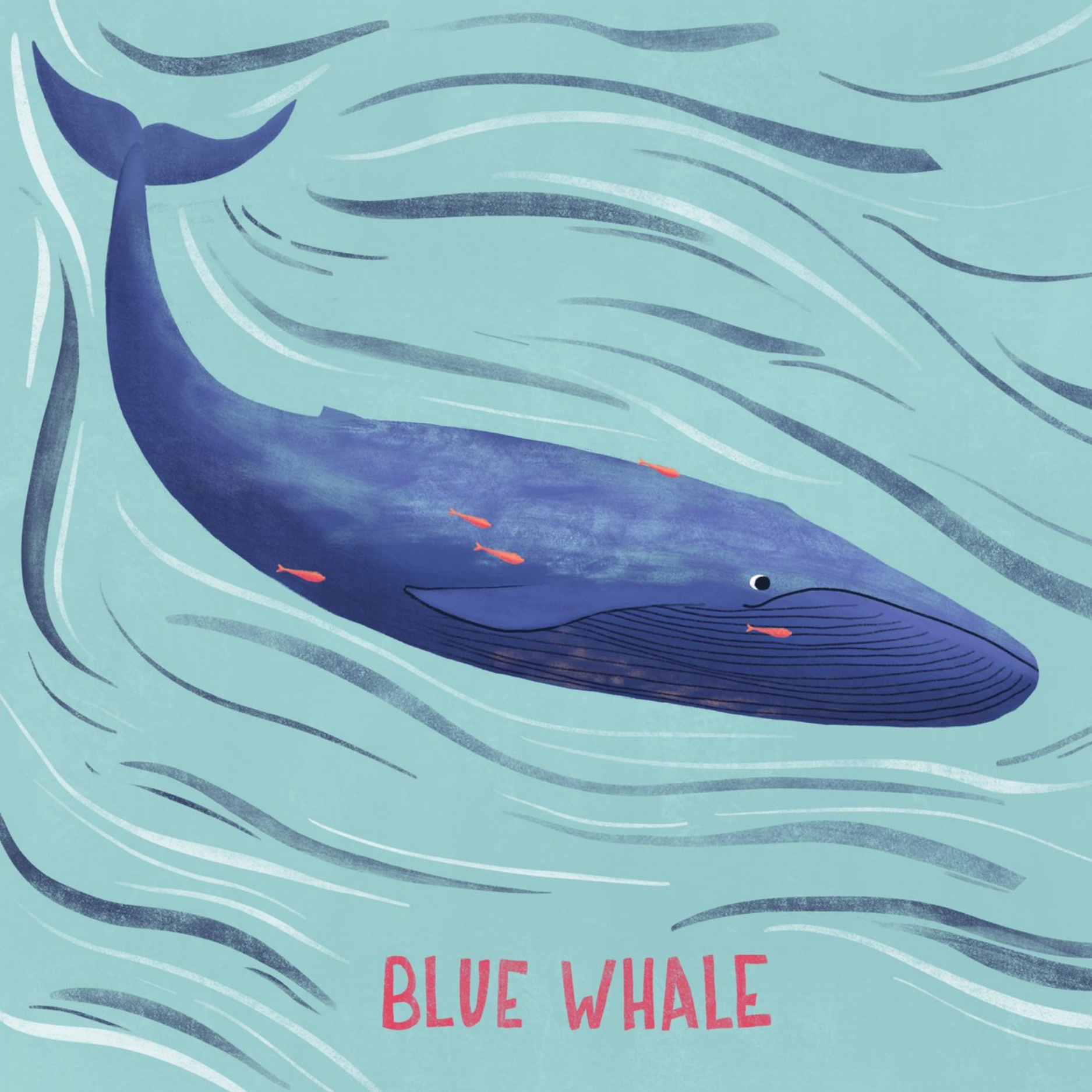}\hfill
    \includegraphics[height=2cm, width=3cm]{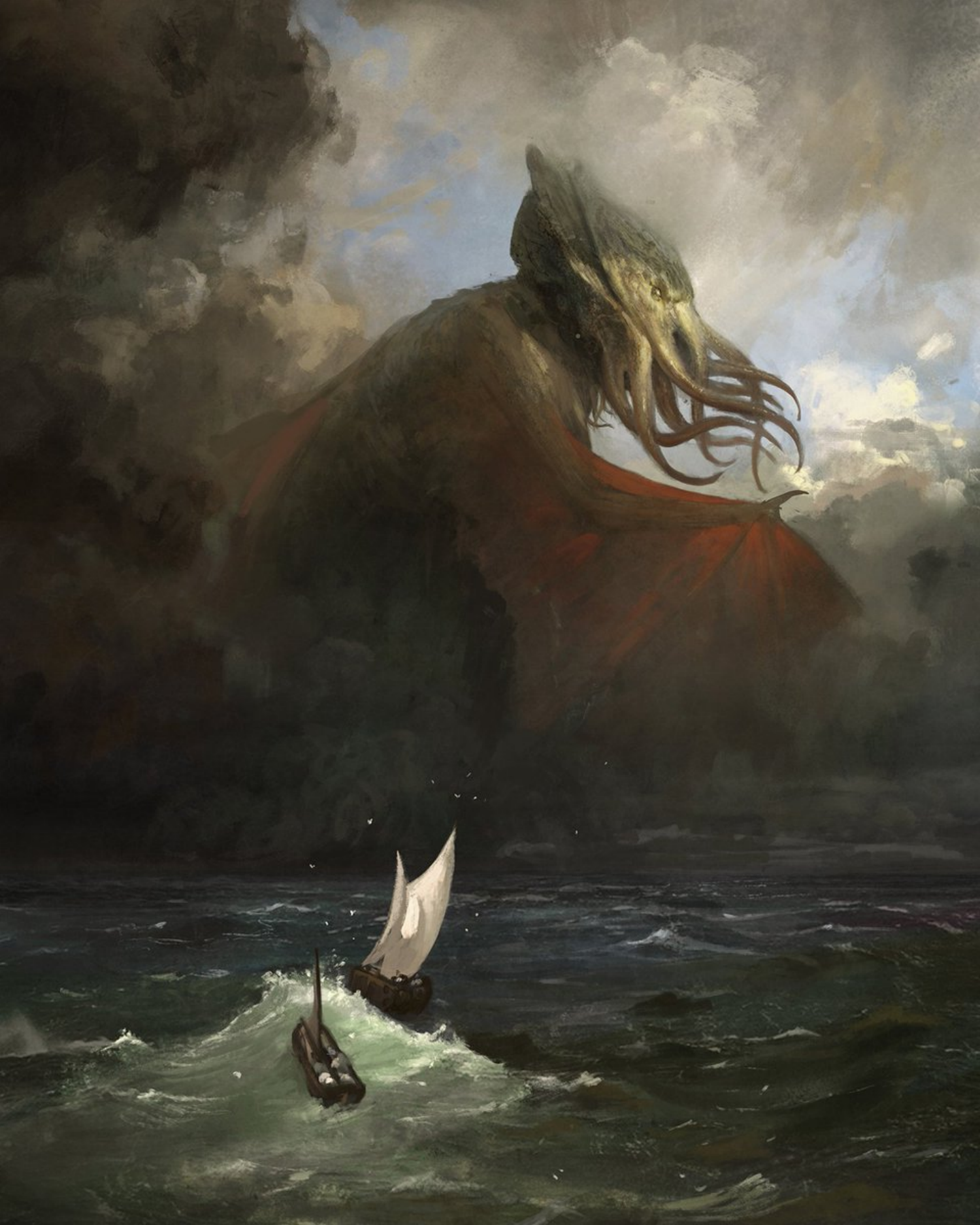}\hfill
    \includegraphics[height=2cm, width=3cm]{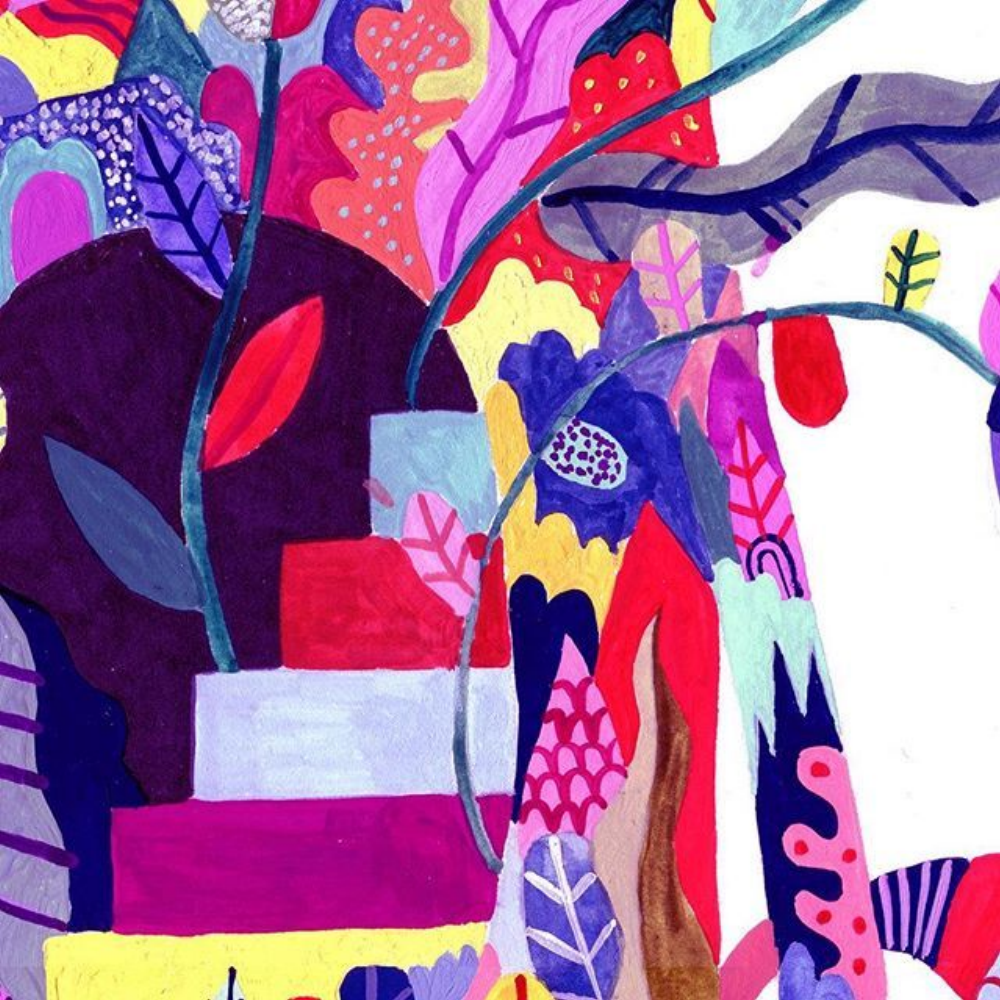}
\caption{\textit{ClimateTV} image examples for the super-class \textit{type}'s class \textit{illustration}. }
\label{fig:type_ill_tv}
\vspace{1em}
\end{figure*}

Very different sets of images are found in the super-category \textit{type}. 
Since we keep the \textit{type} of the image constant, the content varies freely.
For visualisation, \autoref{fig:type_ill_ct} and \autoref{fig:type_ill_tv} contain images of the type \textit{illustration}.
Besides the fore- and background varying drastically, the variance within the same image type is also immense.
Illustrations can take many forms and shapes, as the provided example images display~e.g. drawings, cartoons, comics,~etc.
While some images are rather informative, others are humorous, visually pleasing, entertaining, or a combination thereof.
Within this class, text also plays an important role, as it is frequently contained in illustrations.
This poses a challenge for computer vision models, as the text and the illustration are not necessarily aligned.

\subsection{Category Diversity}

\begin{figure}[ht]
    \centering
    \includegraphics[width=0.85\linewidth]{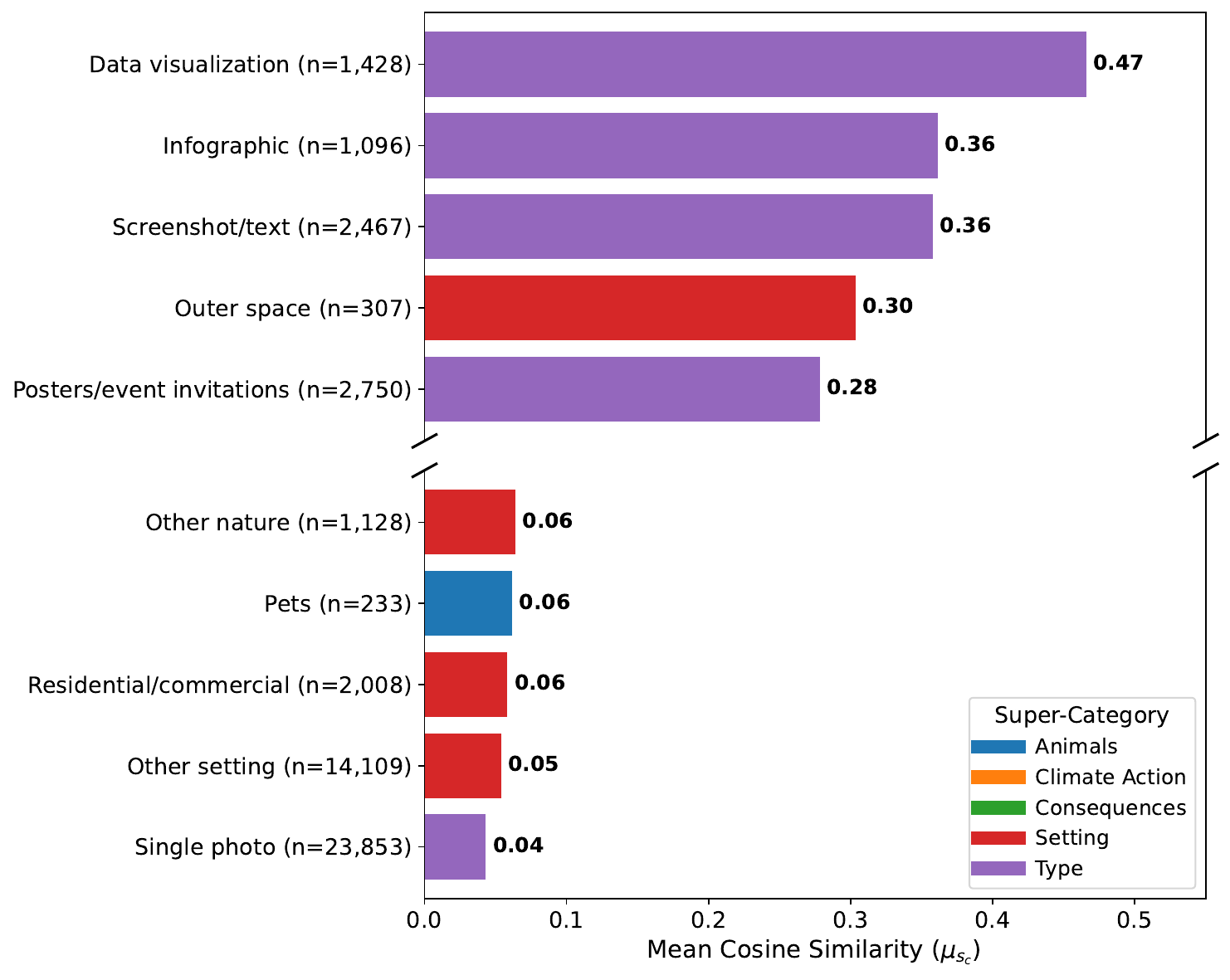}
    \caption{Within category diversity is the highest and lowest within type [$\bar{s}_c$ in DINOv2 embedding space]. We find \textit{type} and \textit{setting} categories to be predominant in both the top and bottom cosine similarities.}
    \label{fig:top5div}
\end{figure}

We investigate the diversity within categories to better understand the dataset analysed.
We compare image diversity using pairwise cosine similarity $ s_c \in [-1,1] $ and total variation in the DINOv2 embedding space~\citep{dinov2}.
\Cref{fig:top5div} shows that the highest similarity can be found between images when comparing the \textit{type} super-category.
We have plotted the mean $s_c$ per category, with higher values indicating higher similarity between image pairs sharing the same label.
This analysis focuses on image labels, images with multiple labels appear multiple times.
\textit{Data visualization} and \textit{infographic} have a high within category similarity.
The detailed analysis in \Cref{tab:combined_diversity_metrics} shows that this is characteristic of these categories, as the mean and median differ by at most 0.01 for both categories. 
For \textit{screenshot/text} and \textit{posters/event invitations}, the gap between median and mean becomes larger (0.04 - 0.05), indicating that the high similarity is driven by several instances of similar visual features.

Images of type \textit{single photo} have the highest diversity with a $\bar{s_c} = 0.04$ and a large range ($s_c \in [-0.28, 0.98]$), similar to the other most common categories, i.e. \textit{no animals}, \textit{no consequences}, \textit{no climate action}, \textit{other setting}.
For the super-category \textit{setting}, the category \textit{other space} is the least visually diverse in this analysis, whereas more abstract labels (\textit{other setting}, \textit{residential/ commercial}, and \textit{other nature}) have much lower $\bar{s_c}$ values. 
This analysis is limited to the expressiveness of the DINOv2 embedding space.

\begin{table}[p] 
    \label{tab:combos}
    \centering
    \scriptsize
    \sisetup{
        retain-explicit-plus = false,
        add-integer-zero = false,
        group-separator = {,}
    }

    \begin{subtable}{\textwidth}
        \centering
        \caption{\textit{animals}}
        \begin{tabularx}{\textwidth}{l @{\extracolsep{\fill}} S[table-format=5.0] S[table-format=-1.2] S[table-format=1.2] S[table-format=1.2] S[table-format=1.2] S[table-format=4.2]}
            \toprule
            \textbf{Category} & {$\mathbf{n}$} & {$\mathbf{\min s_c}$} & {$\mathbf{\max s_c}$} & {$\mathbf{\bar{s}_c}$} & {$\mathbf{\text{med } s_c}$} & {$\mathbf{\text{Tot. Var.}}$} \\ 
            \midrule
            Amphibian or Reptile & 70    & -0.13 & .78 & .10 & .07 & 2074.29 \\
            Birds                & 770   & -0.27 & .96 & .07 & .05 & 2186.98 \\
            Farm animals         & 240   & -0.15 & .82 & .18 & .15 & 1903.18 \\
            Fish                 & 450   & -0.22 & .93 & .15 & .10 & 1994.62 \\
            Insects              & 337   & -0.24 & .99 & .07 & .05 & 2203.34 \\
            Land mammal          & 258   & -0.22 & .87 & .09 & .06 & 2143.93 \\
            Mixed animals        & 44    & -0.13 & .78 & .11 & .07 & 1975.28 \\
            No animals           & 45026 & -0.28 & 1.00 & .08 & .04 & 2085.23 \\
            Pets                 & 233   & -0.21 & .74 & .06 & .04 & 2142.82 \\
            Polar bear           & 195   & -0.14 & .93 & .22 & .19 & 1736.91 \\
            Sea mammal           & 232   & -0.16 & .85 & .14 & .10 & 1974.32 \\ 
            \midrule
            \textbf{Average}     & \textbf{4350} & \textbf{-.20} & \textbf{.88} & \textbf{.11} & \textbf{.08} & \textbf{2038.26} \\
            \bottomrule
        \end{tabularx}
        \label{tab:div_animals_dino}
    \end{subtable}

    \vspace{5pt}

    \begin{subtable}{\textwidth}
        \centering
        \caption{\textit{climate action}}
        \begin{tabularx}{\textwidth}{l @{\extracolsep{\fill}} S[table-format=5.0] S[table-format=-1.2] S[table-format=1.2] S[table-format=1.2] S[table-format=1.2] S[table-format=4.2]}
            \toprule
            \textbf{Category} & {$\mathbf{n}$} & {$\mathbf{\min s_c}$} & {$\mathbf{\max s_c}$} & {$\mathbf{\bar{s}_c}$} & {$\mathbf{\text{med } s_c}$} & {$\mathbf{\text{Tot. Var.}}$} \\ 
            \midrule
            Fossil energy        & 794   & -0.19 & .97 & .12 & .07 & 2008.00 \\
            No climate action    & 35163 & -0.28 & .99 & .07 & .03 & 2124.42 \\
            Other climate action & 2519  & -0.21 & .97 & .13 & .09 & 1945.36 \\
            Politics             & 4868  & -0.21 & .99 & .18 & .14 & 1834.79 \\
            Protests             & 1388  & -0.20 & .97 & .17 & .13 & 1828.04 \\
            Renewable energy     & 1502  & -0.19 & .97 & .13 & .07 & 1991.82 \\ 
            \midrule
            \textbf{Average}     & \textbf{7706} & \textbf{-.22} & \textbf{.98} & \textbf{.13} & \textbf{.09} & \textbf{1955.40} \\
            \bottomrule
        \end{tabularx}
        \label{tab:div_climateaction_dino}
    \end{subtable}

    \vspace{5pt}

    \begin{subtable}{\textwidth}
        \centering
        \caption{\textit{consequences}}
        \begin{tabularx}{\textwidth}{l @{\extracolsep{\fill}} S[table-format=5.0] S[table-format=-1.2] S[table-format=1.2] S[table-format=1.2] S[table-format=1.2] S[table-format=4.2]}
            \toprule
            \textbf{Category} & {$\mathbf{n}$} & {$\mathbf{\min s_c}$} & {$\mathbf{\max s_c}$} & {$\mathbf{\bar{s}_c}$} & {$\mathbf{\text{med } s_c}$} & {$\mathbf{\text{Tot. Var.}}$} \\ 
            \midrule
            Biodiversity loss     & 171   & -0.14 & .87 & .10 & .08 & 2029.71 \\
            Drought               & 424   & -0.17 & .92 & .13 & .10 & 1977.10 \\
            Floods                & 555   & -0.21 & .91 & .12 & .10 & 1916.92 \\
            Melting Ice           & 803   & -0.16 & .96 & .21 & .17 & 1820.23 \\
            No consequences       & 43000 & -0.28 & 1.00 & .08 & .04 & 2095.61 \\
            Rising temperature    & 425   & -0.16 & .96 & .22 & .17 & 1804.81 \\
            Wildfires             & 638   & -0.16 & .97 & .19 & .16 & 1826.37 \\ 
            \midrule
            \textbf{Average}      & \textbf{3614} & \textbf{-.16} & \textbf{.92} & \textbf{.15} & \textbf{.11} & \textbf{1913.56} \\
            \bottomrule
        \end{tabularx}
        \label{tab:div_consequences_dino}
    \end{subtable}

    \vspace{5pt}

    \begin{subtable}{\textwidth}
        \centering
        \caption{\textit{setting}}
        \begin{tabularx}{\textwidth}{l @{\extracolsep{\fill}} S[table-format=5.0] S[table-format=-1.2] S[table-format=1.2] S[table-format=1.2] S[table-format=1.2] S[table-format=4.2]}
            \toprule
            \textbf{Category} & {$\mathbf{n}$} & {$\mathbf{\min s_c}$} & {$\mathbf{\max s_c}$} & {$\mathbf{\bar{s}_c}$} & {$\mathbf{\text{med } s_c}$} & {$\mathbf{\text{Tot. Var.}}$} \\ 
            \midrule
            Agricultural/rural    & 1005  & -0.23 & .92 & .09 & .07 & 2041.05 \\
            Indoor space          & 3417  & -0.24 & .96 & .17 & .13 & 1833.92 \\
            No setting            & 9299  & -0.23 & .99 & .24 & .20 & 1768.11 \\
            Ocean, coastal        & 772   & -0.18 & .90 & .10 & .08 & 2059.81 \\
            Other setting         & 14109 & -0.27 & .99 & .05 & .03 & 2122.84 \\
            Residential/comm.     & 2008  & -0.25 & .95 & .06 & .04 & 2082.54 \\ 
            \midrule
            \textbf{Average}      & \textbf{2805} & \textbf{-.20} & \textbf{.93} & \textbf{.13} & \textbf{.10} & \textbf{1961.18} \\
            \bottomrule
        \end{tabularx}
        \label{tab:div_setting_dino}
    \end{subtable}

    \vspace{5pt}

    \begin{subtable}{\textwidth}
        \centering
        \caption{\textit{type}}
        \begin{tabularx}{\textwidth}{l @{\extracolsep{\fill}} S[table-format=5.0] S[table-format=-1.2] S[table-format=1.2] S[table-format=1.2] S[table-format=1.2] S[table-format=4.2]}
            \toprule
            \textbf{Category} & {$\mathbf{n}$} & {$\mathbf{\min s_c}$} & {$\mathbf{\max s_c}$} & {$\mathbf{\bar{s}_c}$} & {$\mathbf{\text{med } s_c}$} & {$\mathbf{\text{Tot. Var.}}$} \\ 
            \midrule
            Data visualization    & 1428  & -0.13 & .97 & .47 & .47 & 1267.53 \\
            Illustration          & 2668  & -0.18 & .97 & .19 & .16 & 1819.12 \\
            Infographic           & 1096  & -0.12 & .96 & .36 & .35 & 1495.57 \\
            Posters/invites       & 2750  & -0.16 & .97 & .28 & .24 & 1650.72 \\
            Screenshot/text       & 2467  & -0.16 & .99 & .36 & .31 & 1499.95 \\
            Single photo          & 23853 & -0.28 & .98 & .04 & .03 & 2145.23 \\ 
            \midrule
            \textbf{Average}      & \textbf{3986} & \textbf{-.17} & \textbf{.96} & \textbf{.24} & \textbf{.21} & \textbf{1717.01} \\
            \bottomrule
        \end{tabularx}
        \label{tab:div_type_dino}
    \end{subtable}

    \caption{[DINOv2] Diversity analysis across super-categories in the \textit{ClimateTV} dataset.}
    \label{tab:combined_diversity_metrics}
\end{table}

\subsection{Label Combinations}
\label{app:label_combos}

Given the five-dimensional labelling scheme, we can further investigate the label combinations across super-categories.
Out of all possible combinations (92,664), we have 2,637 (2.8\%) in the confirmed \textit{ClimateTV} subset.
We have at least one confirmed label for each category in the codebook, confirmed by the manual validations.
However, for some images, not all labels could be confirmed; hence, we have 1-5 confirmed labels per image, which we use for this visualisation.

\begin{figure}[ht]
    \centering
    \includegraphics[width=0.7\linewidth]{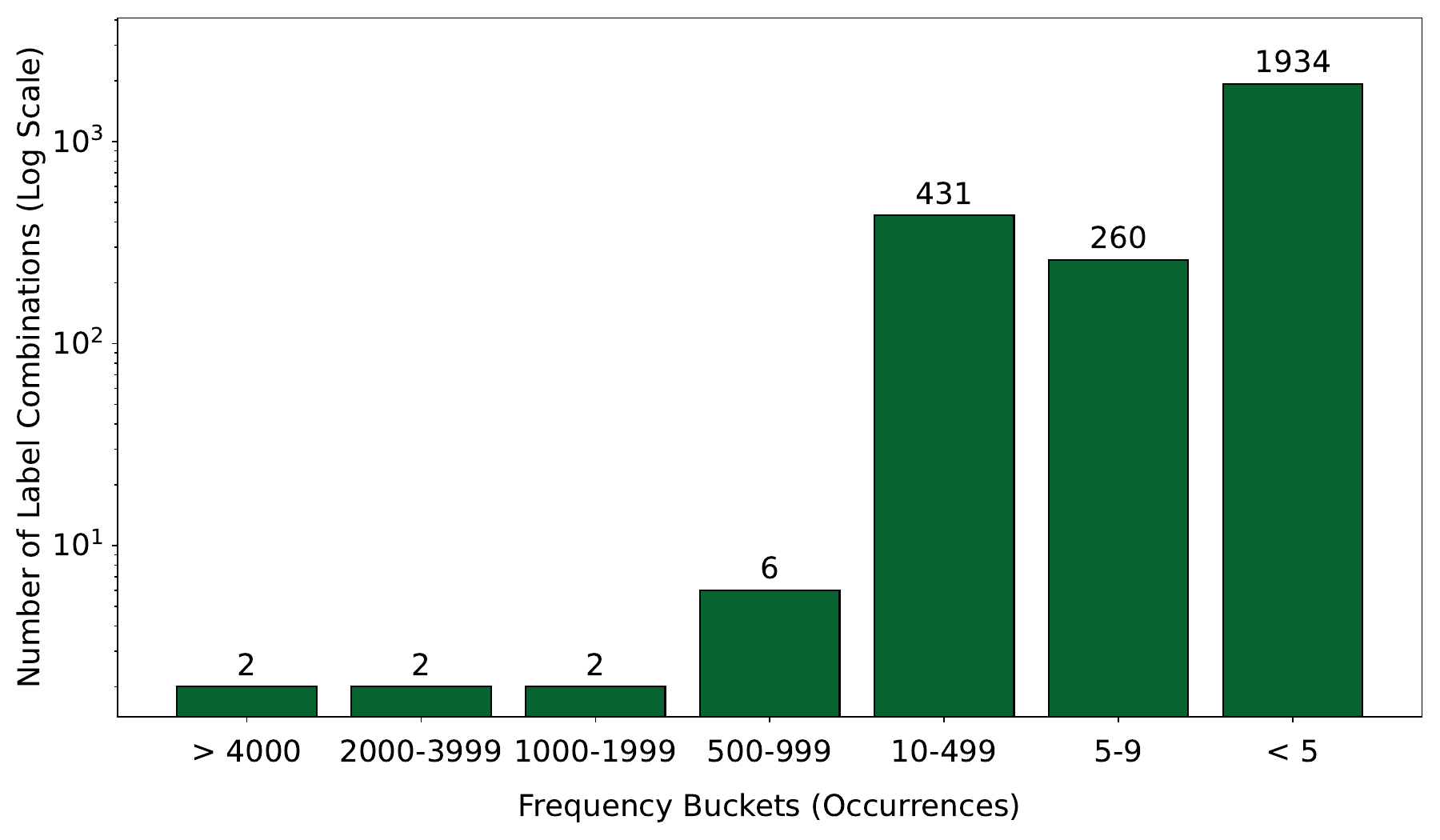}
    \caption{Category combinations across super-categories reveal highly diverse visual content across the confirmed \textit{ClimateTV} images. While the top 12 most common combinations have no direct relation to climate change (no consequences \& no climate action), the majority of other images are part of diverse label combinations.}
    \label{fig:combos}
\end{figure}

\Cref{fig:combos} shows that some combinations have many instances, while many combinations have few instances.
The images with frequent label combinations show little relation to climate change as coded by our codebook.
All top 12 most frequent combinations have \textit{no animals} and \textit{no consequences}.
Most of these combinations include \textit{no climate action} and \textit{single photo} with varying settings, as shown in \Cref{tab:label_combs}.
The descriptive statistics show that, given our codebook, the images in our dataset show a large degree of diversity, with 2,625 label combinations appearing less than 500 times in our sample of 50,000 images.
Moreover, this shows that there are many combinations that are not present in this subset. 

\medskip

\begin{table}[ht]
\scriptsize
\centering
\begin{tabular}{lllllc}
\toprule
\textbf{Animals} & \textbf{Climate action} & \textbf{Consequences} & \textbf{Setting} & \textbf{Type} & \textbf{$n$} \\ \midrule
No animals & No climate action & No consequences & Other setting & Single photo & 4810 \\
No animals & No climate action & No consequences & na & Single photo & 4429 \\
No animals & No climate action & No consequences & na & na & 2475 \\
No animals & No climate action & No consequences & Other setting & na & 2027 \\
No animals & No climate action & No consequences & No setting & na & 1838 \\
No animals & Politics & No consequences & Indoor space & Single photo & 1050 \\
No animals & No climate action & No consequences & Indoor space & Single photo & 847 \\
No animals & No climate action & No consequences & Residential/commercial & Single photo & 772 \\
No animals & No climate action & No consequences & No setting & Single photo & 740 \\
No animals & No climate action & No consequences & No setting & Screenshot & 633 \\ 
No animals & Politics & No consequences & na & Single photo & 595 \\
No animals & No climate action & No consequences & No setting & Illustration & 554 \\ \bottomrule
\end{tabular}
\caption{Top 12 most frequent label combinations in the manually confirmed \textit{ClimateTV}.}
\label{tab:label_combs}
\end{table}

\newpage
\newpage
\newpage

\section{Image Caption Generation}
\label{app:gemini}
This query was used to generate image captions using Gemini-pro-vision~\citep{gemini_team_gemini_2023}.
For some images, the first generation of gemini was unable to generate image captions, for these images, gemini-2.0-flash-lite was used.
For both models, the same prompt was used.

\begin{quote}
    Describe this image in much detail. Do not interpret anything. Whenever there are things in the image, also include their count. If there is text in the image, include the text. Also describe the background of the image.
    Also describe the \textbf{setting} of the background. Examples can be indoor, ocean, desert. outer space, residential area, etc. Also describe the \textbf{type} of the image. Here are examples of image types: photo, illustration, diagram, cartoon, etc.
\end{quote}

\section{Manual annotation}
We have created automated annotations for five super-categories using the Gemini captions and mapped them to category labels. 
50k images out of \textit{ClimateTV} have been manually validated (True / False / Can't Solve) by three independent annotators. 
Moreover, \textit{ClimateCT} consists of manual labels from another work \citep{bravo2025viral} extended manually to contain at least 21 instances per category. 
The added images have been annotated according to the same categories as the first share of images.

\subsection{ClimateTV}
\label{app:tv_stats}

We use QualityMatch workers to manually validate the image labels.
To this end, we use the QualityMatch interface and show short task descriptions (see \Cref{tab:qm_task}).
The annotators were shown the image separately for each super-category with the automatically assigned label overlayed in the top left corner. 

\begin{table}[ht]
\scriptsize
    \centering
    \begin{tabular}{lp{12cm}}
    \toprule
    Super-category & Annotation task and \textit{task description} \\
    \midrule
    Animals & Check for Animals:  Does the image or any text in it depict or refer to the label? (yellow top left corner, if the is no animal, the category "no animal" is correct) \textit{Count small, partial, symbolic, or artistic depictions if they match the label. The label appears in the top left corner in yellow.}\\
    Climate action & Check for Climate change consequences:  Does the image or any text in it depict or refer to the label (in yellow top left corner)? \textit{Count small, partial, symbolic, or artistic depictions if they match the label. The label appears in the top left corner.}\\
    Consequences & Check for Climate change consequences:  Does the image or any text in it depict or refer to the label (in yellow top left corner)? \textit{Count small, partial, symbolic, or artistic depictions if they match the label. The label appears in the top left corner.}\\
    Setting & Check the Background: Does the label match the image’s background setting? (in yellow top left corner, read description below!) \textit{Select "No" if the setting is unclear. If there is no background (e.g., a poster or solid color), the correct label is "No setting". The label appears in the top-left corner.} \\
    Type & Check the Image type: Does the image label describe its type? (in yellow top left corner, read description below!) \textit{The categories are: 'Illustration', 'Infographic', 'Meme', 'Posters/event invitations', 'Single photo', 'Other type', 'Photo collage', 'Data visualization', 'Screenshot'. The label appears in the top left corner. }\\
    \bottomrule
    \end{tabular}
    \caption{Annotation task as shown to QualityMatch workers.}
    \label{tab:qm_task}
\end{table}

We evaluate the quality of annotations using agreement.
Besides validating our assigned labels, this also indicates which categories are more often confused by human annotators.
The annotation statistics for \textit{ClimateTV} validation reveal a clear disparity in clarity and agreement across different super-categories, as shown in \Cref{tab:qm_stats}. 
The \textit{animals} super-category emerges as the most robust, achieving a high acceptance rate of $95.71\%$ and a strong consensus ($78.29\%$), suggesting its labels are highly objective.
In contrast, the \textit{setting} super-category appears to be the most problematic, with the lowest acceptance rate ($67.33\%$) and a significant drop in unanimous agreement to just $32.30\%$.
While \textit{climate action} and \textit{consequences} maintain high general acceptance levels above $92\%$, they exhibit a consensus gap of roughly $30\%$, indicating that these categories likely contain more subjective content that leads to minor disagreements among annotators.
Overall, the negligible rates of ``Multiple 'Can't Solve" responses across all super-categories - all under $0.2\%$ - suggest that while some labels are ambiguous, the data is rarely completely uninterpretable and the category labels and task descriptions are sufficiently clear.

\begin{table}[ht]
\scriptsize
    \centering
    \begin{tabular}{lcccc}
    \toprule
    Super-category & Accept &  All accept & 1 ``Can't Solve'' & Multiple ``Can't Solve'' \\
    \midrule
    \textit{Animals} & 95.71\% & 78.29\% & 0.64\% & 0.01\% \\
    \textit{Climate action} & 92.47\% & 63.81\% & 3.28\% & 0.06\% \\
    \textit{Consequences} & 93.97\% & 60.59\% & 2.34\% & 0.04\% \\
    \textit{Setting} & 67.33\% & 32.30\% & 6.85\% & 0.19\% \\
    \textit{Type} & 71.74\% & 58.29\% & 2.01\% & 0.02\% \\
    \bottomrule
    \end{tabular}
    \caption{Annotation statistics for QualityMatch label validation on \textit{ClimateTV}.}
    \label{tab:qm_stats}
\end{table}

\subsection{ClimateCT}
\label{app:ct_stats}

We have taken the labels from~\citep{bravo2025viral} for the most popular tweets in the duration of the sample.
For the additional annotations, the images were annotated separately for each super-category, thus each image was seen five times by the annotators. 
Two domain experts assigned a label to each image and conflicting labels were resolved during discussion. 
Before annotation, all labels were explained based on the codebook.

\begin{table}[ht]
\caption{We achieve satisfactory annotation metrics for all super-categories. \textit{Consequences} and \textit{Climate action} cause more disagreement between reviewers than the other three super-categories.}
\scriptsize
  \begin{center}{
\begin{tabular}{lccc}
\toprule
Super-category & Kappa & Krippendorff & Agreement \\
\midrule
Animals & 88.7 & 88.7 & 92.3 \\
Climate action & 85.8 & 85.8 & 97.2 \\
Consequences & 84.1 & 84.0 & 89.2 \\
Setting & 87.5 & 87.5 & 88.9 \\
Type & 88.7 & 88.7 & 92.3 \\
\bottomrule
\end{tabular}
}
\end{center}
\label{table:annotationIK}
\vspace{-0.5cm}
\end{table}

In contrast to~\citep{bravo2025viral}, we have combined the super-categories \textit{human activity} and \textit{adaptation and mitigation} in the super-category \textit{climate action}. 
In comparison to the previous annotation metrics, we achieve a similar performance in both Kappa (-0.002) and Krippendorff (+0.023).

\section{Models employed}
\label{app:models}

We compared 6 promptable VLM models (\Cref{tab:vlm_models}) and 15 CLIP-like models (\Cref{tab:zeroshot_models}). Gemini-3.1-flash-lite-preview [5] was the promptable VLM model in our analysis and MetaCLIP ViT-L/14 [6] was the best CLIP-like zero-shot model.

\begin{table}[ht]
    \scriptsize
    \centering
    \begin{minipage}[t]{0.45\textwidth}
        \centering
        \begin{tabular}{@{}llll@{}}
        \toprule
        \# & Model name & Params & Authors \\
        \midrule
        1 & qwen3-VL-8B-Instruct & 8B & Qwen Team \\
        2 & qwen3-VL-30B-A3B-Instruct & 30B & Qwen Team \\
        3 & moondream3-preview & 9B* & Moondream \\
        4 & gemma-3-4-it & 4B & Google \\
        5 & gemini-3.1-flash-lite & n/a & Google \\
        6 & gpt-5.4-mini & n/a & OpenAI \\
        \bottomrule
        \end{tabular}
        \caption{Promptable VLMs analysed.}
        \label{tab:vlm_models}
    \end{minipage}
    \hfill 
    \begin{minipage}[t]{0.45\textwidth}
        \centering
        \begin{tabular}{@{}lll@{}}
        \toprule
        \# & Backbone & Pre-training dataset \\
        \midrule
        1 & CLIP-RN50 & cc12m \\
        2 & CLIP-RN50-qgelu & cc12m \\
        3 & CLIP-RN101 & yfcc15m \\
        4 & CLIP-ViT-L/14 & openai \\
        5 & CLIP-ViT-L/14 & commonpool\_xl \\
        6 & MetaCLIP-ViT-L/14 & fullcc \\
        7 & ViT-L/14-qgelu & metaclip\_400m \\
        8 & SigLIP-ViT-L/16 & webli \\
        9 & EVO02-B/16 & merged2b\_s8b \\
        10 & EVA02-L/14-336 & merged2b\_s6b \\
        11 & EVA02-L/14 & merged2b\_s4b \\ 
        12 & coca\_ViT-L/14 & laion2b\_s13b \\
        13 & coca\_ViT-L/14 & mscoco\_ft \\
        14 & convnext\_xxlarge & laion2b\_s34b \\
        15 & convnext\_large\_d & laion2b\_s29b \\
        \bottomrule
        \end{tabular}
        \caption{Zero-shot VLMs analysed, all from openclip \citep{openclip}.}
        \label{tab:zeroshot_models}
    \end{minipage}
\end{table}

For gpt-5.4-mini and Gemini-3.1-flash-lite, we submit our requests to the batch API.
We resize the images to have a maximum of 512*512 pixels while maintaining their aspect ratio. 
All model scripts can be found at \url{https://github.com/KathPra/Codebooks2VLMs.git}.
For Gemini-3.1-flash-lite, we use the default of \textit{minimal} thinking and for gpt-5.4-mini we use the default of \textit{none}.
All VLM experiments were run in February and March 2026.

\section{Benchmarking}
\begin{figure}[ht]
    \centering
    \begin{subfigure}[b]{0.48\textwidth}
        \centering
        \includegraphics[width=0.7\linewidth]{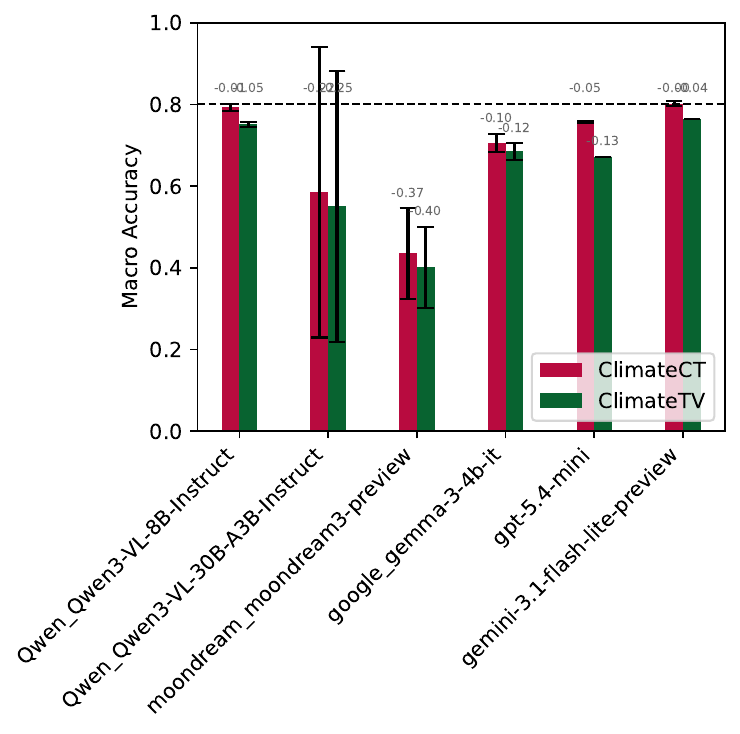}
        \caption{\textit{animals}}
        \label{fig:vlm_bench_animals}
    \end{subfigure}
    \hfill
    \begin{subfigure}[b]{0.48\textwidth}
        \centering
        \includegraphics[width=0.7\linewidth]{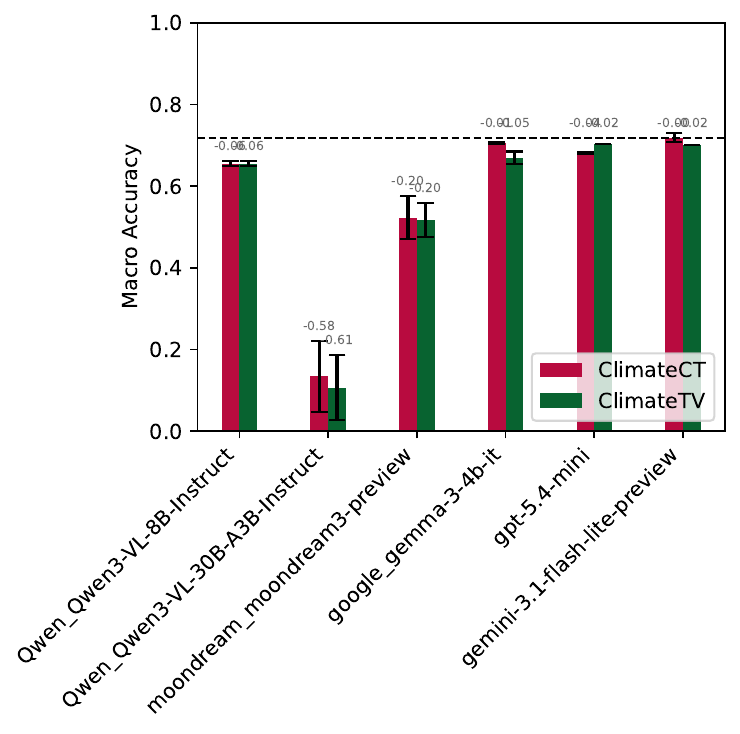}
        \caption{\textit{climate action}}
        \label{fig:vlm_bench_climateaction}
    \end{subfigure}

    \vspace{10pt} 

    \begin{subfigure}[b]{0.48\textwidth}
        \centering
        \includegraphics[width=0.7\linewidth]{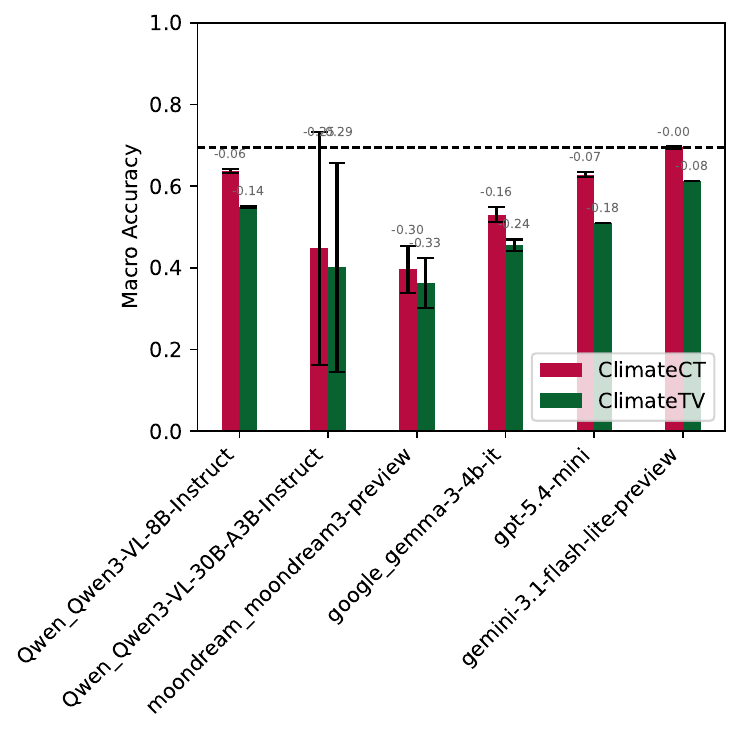}
        \caption{\textit{setting}}
        \label{fig:vlm_bench_setting}
    \end{subfigure}
    \hfill
    \begin{subfigure}[b]{0.48\textwidth}
        \centering
        \includegraphics[width=0.7\linewidth]{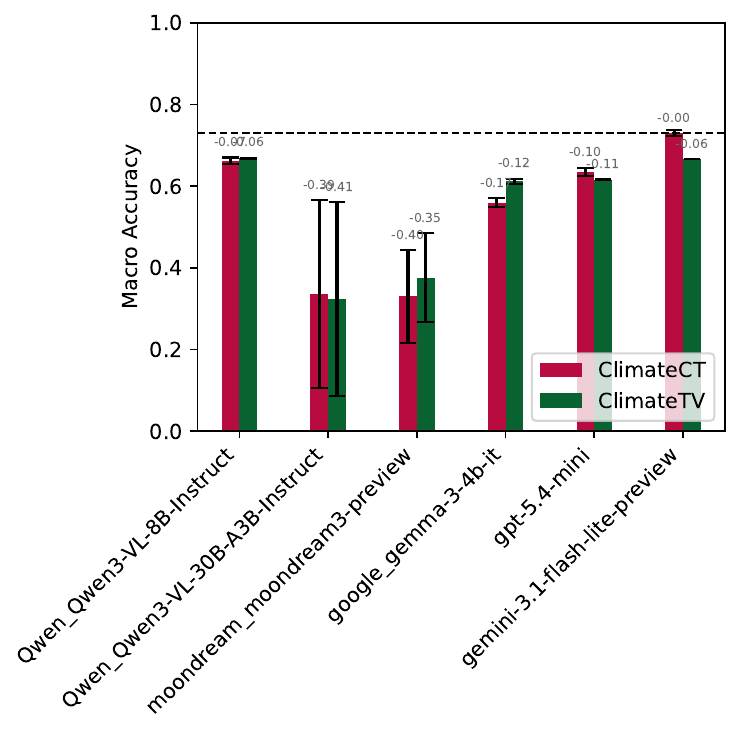}
        \caption{\textit{type}}
        \label{fig:vlm_bench_type}
    \end{subfigure}

    \caption{VLM Benchmarking across super-categories on \textit{ClimateTV} and \textit{ClimateCT}.}
    \label{fig:vlm_bench_combined}
\end{figure}

\Cref{fig:vlm_bench_combined} contains the benchmarking results for the remaining super-categories.
Across almost all super-categories and models, performance on \textit{ClimateCT} consistently exceeds that of \textit{ClimateTV}. 
This suggests that the \textit{ClimateTV} dataset presents a higher level of complexity, as we expected, given that it contains all images posted over the course of 4 years, whereas \textit{ClimateCT} contains the most popular ones.
Qwen-30B shows the highest variance of all assessed models.
This is due to highly variable performance given different versions of the same prompt.
Within the super-category \textit{climate action}, the performance across the three prompt paraphrases is equally low; thus, the variance is lower compared to all other super-categories.
The other model's performance is stable across super-categories.
For moondream, we notice higher variance for \textit{animals} and \textit{type} in comparison to \textit{climate action} and \textit{setting}. 

\medskip
\medskip
\medskip

\section{Gemini-3.1-flash-lite - additional results}
\label{app:gemini}

We provide additional results for the best performing promptable VLM model, Gemini-3.1-flash-lite-preview, to complement the analysis in the main paper. 
\Cref{tab:gemini_metrics_final} contains category-wise metrics for Gemini-3.1-flash-lite-preview and \Cref{tab:combined_confusion_metrics} contains the top five category confusions per super-category.
The provided performance metrics reveal a classification system that is heavily influenced by significant category imbalance, where exceptionally high weighted averages (such as the $0.97$ in \textit{animals}) mask underlying difficulties in identifying minority classes. 
While the model excels at identifying common or ``null" categories - most notably \textit{no animals} and \textit{single photo} it struggles with nuanced or rare labels, evidenced by the dismal $0.05$ F1-score for \textit{other/mixed animals} and the $0.06$ for \textit{biodiversity loss}. 
The \textit{setting} super-category appears the most problematic, with a weighted F1 of only $0.33$, largely due to a near-total failure to correctly recall the \textit{other setting} category. 
Overall, the system is reliable for broad structural classification (\textit{type}) but lacks the discriminative power for the overflow categories.

\begin{table}[ht]
    \scriptsize
    \centering
    \caption{Classification performance metrics for all super-categories reporting accuracy (Acc.), precision (P), recall (R), and support (Supp.).}
    \label{tab:gemini_metrics_final}
    \addtolength{\tabcolsep}{-3pt} 

    \begin{subtable}[t]{0.48\textwidth}
        \centering
        \caption{\textit{animals}}
        \begin{tabular}{lrrrrr}
            \toprule
            \textbf{Category} & \textbf{Acc.} & \textbf{P} & \textbf{R} & \textbf{F1} & \textbf{Supp.} \\
            \midrule
            Amphibian/reptile & $.86$ & $.55$ & $.86$ & $.67$ & 70 \\
            Birds & $.72$ & $.89$ & $.72$ & $.80$ & 770 \\
            Farm animals & $.76$ & $.69$ & $.76$ & $.73$ & 240 \\
            Fish & $.52$ & $.85$ & $.52$ & $.64$ & 450 \\
            Invertebrates & $.79$ & $.56$ & $.79$ & $.66$ & 337 \\
            Land mammal & $.68$ & $.60$ & $.68$ & $.64$ & 258 \\
            Mixed animals & $.09$ & $.04$ & $.09$ & $.05$ & 44 \\
            No animals & $.99$ & $.99$ & $.99$ & $.99$ & 45,026 \\
            Pets & $.66$ & $.82$ & $.66$ & $.73$ & 233 \\
            Polar bear & $.74$ & $.80$ & $.74$ & $.77$ & 195 \\
            Sea mammal & $.59$ & $.80$ & $.59$ & $.68$ & 232 \\
            \midrule
            \textbf{Macro Avg} & $.67$ & $.69$ & $.67$ & $.67$ & 47,855 \\
            \textbf{Weighted Avg} & $.97$ & $.97$ & $.97$ & $.97$ & 47,855 \\
            \bottomrule
        \end{tabular}
    \end{subtable}
    \hfill 
    \begin{subtable}[t]{0.48\textwidth}
        \centering
        \caption{\textit{climate action}}
        \begin{tabular}{lrrrrr}
            \toprule
            \textbf{Category} & \textbf{Acc.} & \textbf{P} & \textbf{R} & \textbf{F1} & \textbf{Supp.} \\
            \midrule
            Fossil energy & $.62$ & $.39$ & $.62$ & $.48$ & 794 \\
            No climate action & $.57$ & $.96$ & $.57$ & $.72$ & 35,163 \\
            Other climate & $.79$ & $.17$ & $.79$ & $.28$ & 2,519 \\
            Politics & $.71$ & $.40$ & $.71$ & $.52$ & 4,868 \\
            Protests & $.78$ & $.56$ & $.78$ & $.65$ & 1,388 \\
            Sustainable energy & $.74$ & $.60$ & $.74$ & $.66$ & 1,502 \\
            \midrule
            \textbf{Macro Avg} & $.70$ & $.51$ & $.70$ & $.55$ & 46,234 \\
            \textbf{Weighted Avg} & $.61$ & $.83$ & $.61$ & $.66$ & 46,234 \\
            \bottomrule
        \end{tabular}
    \end{subtable}
    
    \vspace{15pt} 
    
    \begin{subtable}[t]{0.48\textwidth}
    \centering
    \caption{\textit{consequences}}
    \begin{tabular}{lrrrrr}
            \toprule
            \textbf{Category} & \textbf{Acc.} & \textbf{P} & \textbf{R} & \textbf{F1} & \textbf{Supp.} \\
            \midrule
            Biodiversity loss & $.61$ & $.03$ & $.61$ & $.06$ & 171 \\
            Drought & $.64$ & $.27$ & $.64$ & $.38$ & 424 \\
            Floods & $.80$ & $.40$ & $.80$ & $.53$ & 555 \\
            Melting Ice & $.84$ & $.45$ & $.84$ & $.59$ & 803 \\
            No consequences & $.31$ & $.99$ & $.31$ & $.47$ & 43,000 \\
            Rising temp. & $.79$ & $.14$ & $.79$ & $.23$ & 425 \\
            Wildfires & $.81$ & $.65$ & $.81$ & $.72$ & 638 \\
            \midrule
            \textbf{Macro Avg} & $.62$ & $.25$ & $.62$ & $.27$ & 46,988 \\
            \textbf{Weighted Avg} & $.35$ & $.94$ & $.35$ & $.47$ & 46,988 \\
            \bottomrule
        \end{tabular}
    \end{subtable}
    \hfill
    \begin{subtable}[t]{0.48\textwidth}
        \centering
        \caption{\textit{setting}}
        \begin{tabular}{lrrrrr}
            \toprule
            \textbf{Category} & \textbf{Acc.} & \textbf{P} & \textbf{R} & \textbf{F1} & \textbf{Supp.} \\
            \midrule
            Agricultural/rural & $.61$ & $.29$ & $.61$ & $.40$ & 1,005 \\
            Arctic/Antarctica & $.64$ & $.38$ & $.64$ & $.48$ & 367 \\
            Forest, jungle & $.73$ & $.23$ & $.73$ & $.35$ & 408 \\
            Indoor space & $.92$ & $.34$ & $.92$ & $.50$ & 3,417 \\
            Nature & $.41$ & $.17$ & $.41$ & $.24$ & 1,128 \\
            No setting & $.60$ & $.70$ & $.60$ & $.65$ & 9,299 \\
            Other setting & $.02$ & $.40$ & $.02$ & $.04$ & 14,109 \\
            Residential & $.70$ & $.27$ & $.70$ & $.39$ & 2,008 \\
            \midrule
            \textbf{Macro Avg} & $.61$ & $.38$ & $.61$ & $.42$ & 34,064 \\
            \textbf{Weighted Avg} & $.40$ & $.46$ & $.40$ & $.33$ & 34,064 \\
            \bottomrule
        \end{tabular}
    \end{subtable}

        \vspace{15pt} 
        
    \begin{subtable}[t]{\textwidth}
        \centering
        \caption{\textit{type}}
        \begin{tabular}{lrrrrr}
            \toprule
            \textbf{Category} & \textbf{Acc.} & \textbf{P} & \textbf{R} & \textbf{F1} & \textbf{Supp.} \\
            \midrule
            Data viz. & $.82$ & $.70$ & $.82$ & $.75$ & 1,428 \\
            Illustration & $.53$ & $.79$ & $.53$ & $.64$ & 2,668 \\
            Infographic & $.61$ & $.43$ & $.61$ & $.50$ & 1,096 \\
            Meme & $.62$ & $.50$ & $.62$ & $.55$ & 559 \\
            Posters/Invite & $.93$ & $.57$ & $.93$ & $.71$ & 2,750 \\
            Screenshot & $.88$ & $.71$ & $.88$ & $.79$ & 2,467 \\
            Single photo & $.89$ & $.99$ & $.89$ & $.94$ & 23,853 \\
            \midrule
            \textbf{Macro Avg} & $.67$ & $.60$ & $.67$ & $.62$ & 35,870 \\
            \textbf{Weighted Avg} & $.84$ & $.87$ & $.84$ & $.84$ & 35,870 \\
            \bottomrule
        \end{tabular}
    \end{subtable}
\end{table}

\clearpage

\begin{table}[htbp]
    \scriptsize
    \centering
    \caption{Error analysis: top 5 confused pairs across super-categories in \textit{ClimateTV}.}
    \label{tab:combined_confusion_metrics}

    \begin{subtable}{0.48\textwidth}
        \centering
        \caption{\textit{animals}}
        \begin{tabularx}{\textwidth}{l X X S}
        \toprule
        \textbf{\#} & \textbf{True Class} & \textbf{Pred.} & {\textbf{Conf.($\mu$)}} \\ 
        \midrule
        1  & Mixed animals & No animals & 0.523 \\
        2  & Fish & Mixed & 0.176 \\
        3  & Fish & Invert. & 0.171 \\
        4  & Sea mammal & Amphib. & 0.116 \\
        5  & Birds & No animals & 0.112 \\
        \bottomrule
        \end{tabularx}
        \label{tab:confusion_animals_tv}
    \end{subtable}
    \hfill
    \begin{subtable}{0.48\textwidth}
        \centering
        \caption{\textit{climate action}}
        \begin{tabularx}{\textwidth}{l X X S}
        \toprule
        \textbf{\#} & \textbf{True Class} & \textbf{Pred.} & {\textbf{Conf.($\mu$)}} \\ 
        \midrule
        1  & No climate & Other & 0.335 \\
        2  & Politics & Other & 0.317 \\
        3  & Sust. energy & Other & 0.208 \\
        4  & No climate & Politics & 0.135 \\
        5  & Fossil energy & Other & 0.125 \\
        \bottomrule
        \end{tabularx}
        \label{tab:confusion_action_tv}
    \end{subtable}

    \vspace{20pt} 

    \begin{subtable}{0.48\textwidth}
        \centering
        \caption{\textit{setting}}
        \begin{tabularx}{\textwidth}{l X X S}
        \toprule
        \textbf{\#} & \textbf{True Class} & \textbf{Pred.} & {\textbf{Conf.($\mu$)}} \\ 
        \midrule
        1  & Other setting & Indoor & 0.284 \\
        2  & Other setting & Res./comm. & 0.215 \\
        3  & Forest, jungle & Nature & 0.184 \\
        4  & No setting & Indoor & 0.174 \\
        5  & Nature & Forest & 0.168 \\
        \bottomrule
        \end{tabularx}
        \label{tab:confusion_setting_tv}
    \end{subtable}
    \hfill
    \begin{subtable}{0.48\textwidth}
        \centering
        \caption{\textit{type}}
        \begin{tabularx}{\textwidth}{l X X S}
        \toprule
        \textbf{\#} & \textbf{True Class} & \textbf{Pred.} & {\textbf{Conf.($\mu$)}} \\ 
        \midrule
        1  & Other type & Posters & 0.330 \\
        2  & Infographic & Data viz & 0.287 \\
        3  & Meme & Illustr. & 0.224 \\
        4  & Illustration & Posters & 0.214 \\
        5  & Other type & Infogr. & 0.142 \\
        \bottomrule
        \end{tabularx}
        \label{tab:confusion_type_tv}
    \end{subtable}
\end{table}

\section{MetaCLIP-ViT-L-14-quickgelu - additional results}

The best performing CLIP-like model is MetaCLIP-ViT-L/14-quickgelu pretrained on the fullcc dataset, as reported in \Cref{tab:metaclip}.

\begin{table}[ht!]
\centering
\scriptsize
\caption{Classification performance metrics for \textit{consequences}.}
\label{tab:consequences_metrics_large}
\begin{tabular}{lccccc}
\toprule
\textbf{Class} & \textbf{Accuracy} & \textbf{Precision} & \textbf{Recall} & \textbf{F1} & \textbf{Support} \\
\midrule
Biodiversity loss            & $.70$ & $.01$ & $.70$ & $.02$ & 171 \\
Covid \& general health      & $.66$ & $.01$ & $.66$ & $.01$ & 58 \\
Drought                      & $.65$ & $.17$ & $.65$ & $.27$ & 424 \\
Economic consequences        & $.56$ & $.01$ & $.56$ & $.02$ & 115 \\
Floods                       & $.72$ & $.30$ & $.72$ & $.43$ & 555 \\
Human rights                 & $.39$ & $.01$ & $.39$ & $.01$ & 88 \\
Melting Ice                  & $.68$ & $.57$ & $.68$ & $.62$ & 803 \\
No consequences              & $.09$ & $.99$ & $.09$ & $.17$ & 43,000 \\
Other consequences           & $.17$ & $.00$ & $.17$ & $.00$ & 36 \\
Other extreme weather events & $.65$ & $.10$ & $.65$ & $.18$ & 393 \\
Rising temperature           & $.77$ & $.08$ & $.77$ & $.14$ & 425 \\
Sea level rise               & $.42$ & $.08$ & $.42$ & $.13$ & 278 \\
Wildfires                    & $.76$ & $.55$ & $.76$ & $.64$ & 638 \\
\midrule
\textbf{Macro Avg}           & $.55$ & $.22$ & $.55$ & $.20$ & 46,938 \\
\textbf{Weighted Avg}        & $.15$ & $.92$ & $.15$ & $.20$ & 46,938 \\
\bottomrule
\end{tabular}
\label{tab:metaclip}
\end{table}

\newpage
\bibliography{sample}

\end{document}